\documentclass[runningheads]{llncs}
\usepackage{graphicx}
\usepackage{amsmath,amssymb} 
\usepackage{color}

\usepackage{times}
\usepackage{epsfig}
\usepackage{graphicx}
\usepackage{float}
\usepackage{wrapfig}

\usepackage{amsmath,amssymb}

\usepackage{bm,xspace}
\usepackage{comment}
\usepackage{verbatim}
\usepackage{multirow}
\usepackage{balance}
\usepackage{url}
\usepackage{booktabs}
\usepackage{etoolbox,siunitx}
\usepackage{calc}
\usepackage{pifont,hologo}
\usepackage{color}

\newcommand{\ra}[1]{\renewcommand{\arraystretch}{#1}}

\usepackage[super]{nth}
\usepackage{nicefrac}
\def\vV{\mathcal{V}}

\DeclareMathOperator*{\argmin}{arg\,min}

\DeclareMathSymbol{@}{\mathord}{letters}{"3B}

\newcommand\abs[1]{\left\lvert#1\right\rvert}
\newcommand\norm[1]{\left\lVert#1\right\rVert}
\newcommand\tuple[1]{\left\langle#1\right\rangle}

\definecolor{alexey}{rgb}{0.8,0.,0.8}

\newcommand\mypara[1]{\vspace{1mm}\noindent\textbf{#1}}

\def\latex/{\LaTeX}
\def\bibtex/{\hologo{BibTeX}}

\newcommand{\act}{\mathbf{a}}

\newcommand{\obs}{\mathbf{o}}

\newcommand{\cmd}{\mathbf{c}}

\newcommand{\params}{\boldsymbol{\theta}}

\newcommand{\net}{F}

\newcommand{\cmdleft}{{\texttt{left}}}
\newcommand{\cmdstraight}{{\texttt{straight}}}
\newcommand{\cmdright}{{\texttt{right}}}
\newcommand{\cmdcontinue}{{\texttt{continue}}}

\newcommand{\agt}{\act}
\newcommand{\apred}{\widehat{\act}}

\definecolor{dred}{rgb}{0.5,0.,0.}
\newcommand{\dred}[1]{{\color{dred}#1}}
\definecolor{dgreen}{rgb}{0.,0.5,0.}
\newcommand{\dgreen}[1]{{\color{dgreen}#1}}

\graphicspath{{./figures/}}

\begin{document}
\title{On Offline Evaluation of Vision-based Driving Models}

\titlerunning{On Offline Evaluation of Vision-based Driving Models}
%
\author{Felipe Codevilla\inst{1} \and
Antonio M. L\'{o}pez\inst{1} \and
Vladlen Koltun\inst{2} \and
Alexey Dosovitskiy\inst{2}}
%
\authorrunning{F. Codevilla, A. M. L\'{o}pez, V. Koltun, and A.Dosovitskiy}
%

\institute{Computer Vision Center, Universitat Aut\`{o}noma de Barcelona \and
Intel Labs}
\maketitle              
\begin{abstract}
Autonomous driving models should ideally be evaluated by deploying them on a fleet of physical vehicles in the real world. Unfortunately, this approach is not practical for the vast majority of researchers. An attractive alternative is to evaluate models offline, on a pre-collected validation dataset with ground truth annotation. In this paper, we investigate the relation between various online and offline metrics for evaluation of autonomous driving models. We find that offline prediction error is not necessarily correlated with driving quality, and two models with identical prediction error can differ dramatically in their driving performance. We show that the correlation of offline evaluation with driving quality can be significantly improved by selecting an appropriate validation dataset and suitable offline metrics.

\end{abstract}

\section{Introduction}
\label{sec:introduction}
\begin{figure}[t]
	\includegraphics[width=\linewidth]{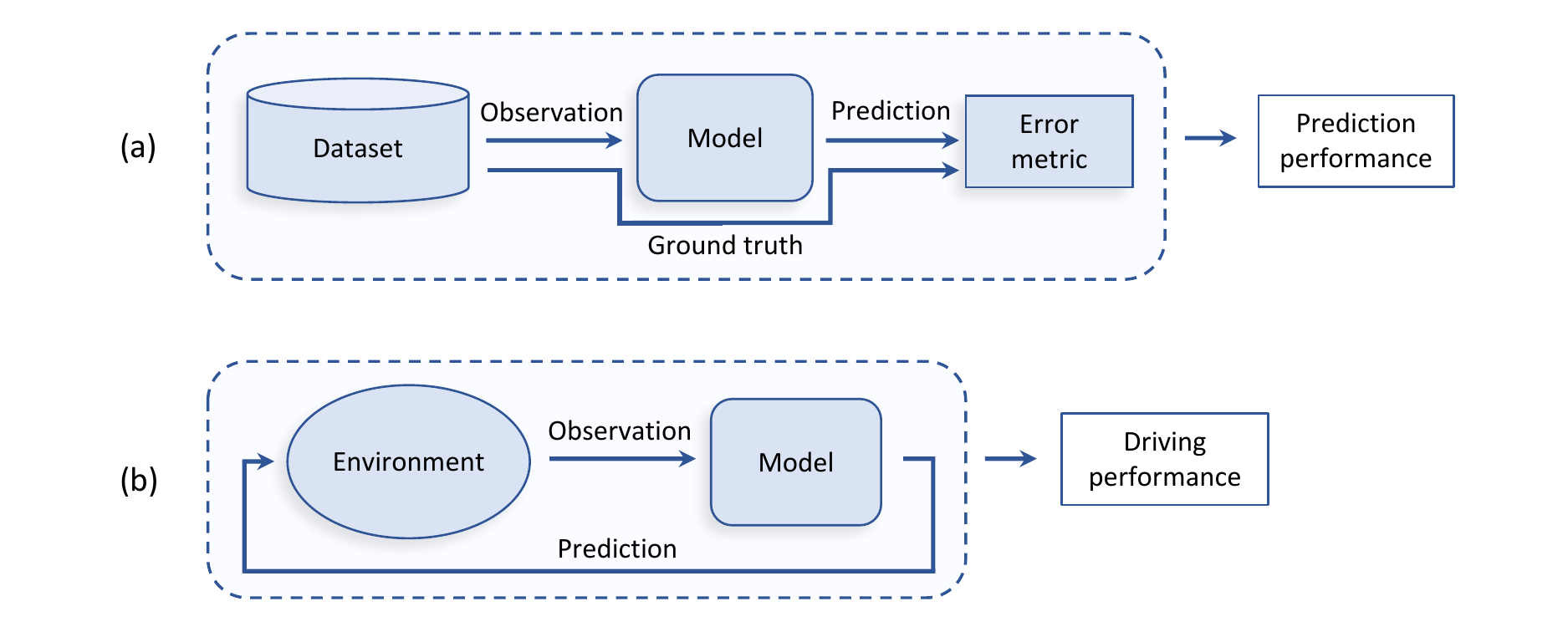}
	\caption{Two approaches to evaluation of a sensorimotor control model. Top: offline (passive) evaluation on a fixed dataset with ground-truth annotation. Bottom: online (active) evaluation with an environment in the loop.}
\label{fig:overview}
\end{figure}

Camera-based autonomous driving can be viewed as a computer vision problem. It requires analyzing the input video stream and estimating certain high-level quantities, such as the desired future trajectory of the vehicle or the raw control signal to be executed.
Standard methodology in computer vision is to evaluate an algorithm by collecting a dataset with ground-truth annotation and evaluating the results produced by the algorithm against this ground truth (Figure~\ref{fig:overview}(a)).
However, driving, in contrast with most computer vision tasks, is inherently active. That is, it involves interaction with the world and other agents.
The end goal is to drive well: safely, comfortably, and in accordance with traffic rules.
An ultimate evaluation would involve deploying a fleet of vehicles in the real world and executing the model on these (Figure~\ref{fig:overview}(b)).
The logistical difficulties associated with such an evaluation lead to the question: Is it possible to evaluate a driving  model without actually letting it drive, but rather following the offline dataset-centric methodology?

One successful approach to evaluation of driving systems is via decomposition.
It stems from the modular approach to driving where separate subsystems deal with subproblems, such as environment perception, mapping, and vehicle control.
The perception stack provides high-level understanding of the scene in terms of semantics, 3D layout, and motion.
These lead to standard computer vision tasks, such as object detection, semantic segmentation, depth estimation, 3D reconstruction, or optical flow estimation, which can be evaluated offline on benchmark datasets~\cite{Geiger2012,Cordts2016,Richter2017}.
This approach has been extremely fruitful, but it only applies to modular driving systems.

Recent deep learning approaches~\cite{Bojarski2016nvidiadriving,Xu2017} aim to replace modular pipelines by end-to-end learning from images to control commands.
The decomposed evaluation does not apply to models of this type.
End-to-end methods are commonly evaluated by collecting a large dataset of expert driving~\cite{Xu2017} and measuring the average prediction error of the model on the dataset.
This offline evaluation is convenient and is consistent with standard practice in computer vision, but how much information does it provide about the actual driving performance of the models?

In this paper, we empirically investigate the relation between (offline) prediction accuracy and (online) driving quality.
We train a diverse set of models for urban driving in realistic simulation~\cite{Dosovitskiy2017} and correlate their driving performance with various metrics of offline prediction accuracy.
By doing so, we aim to find offline evaluation procedures that can be executed on a static dataset, but at the same time correlate well with driving quality.
We empirically discover best practices both in terms of selection of a validation dataset and the design of an error metric.
Additionally, we investigate the performance of several models on the real-world Berkeley DeepDrive Video (BDDV) urban driving dataset~\cite{Xu2017}.

Our key finding is that offline prediction accuracy and actual driving quality are surprisingly weakly correlated.
This correlation is especially low when prediction is evaluated on data collected by a single forward-facing camera on expert driving trajectories~-- the setup used in most existing works.
A network with very low prediction error can be catastrophically bad at actual driving. Conversely, a model with relatively high prediction error may drive well.

We found two general approaches to increasing this poor correlation between prediction and driving.
The first is to use more suitable validation data.
We found that prediction error measured in lateral cameras (sometimes mounted to collect additional images for imitation learning) better correlates with driving performance than prediction in the forward-facing camera alone.
The second approach is to design offline metrics that depart from simple mean squared error (MSE).
We propose offline metrics that correlate with driving performance more than 60\% better than MSE.



\section{Related Work}
\label{sec:relatedwork}
Vision-based autonomous driving tasks have traditionally been evaluated on dedicated annotated real-world datasets. For instance, KITTI~\cite{Geiger2012} is a comprehensive benchmarking suite with annotations for stereo depth estimation, odometry, optical flow estimation, object detection, semantic segmentation, instance segmentation, 3D bounding box prediction, etc.
The Cityscapes dataset~\cite{Cordts2016} provides annotations for semantic and instance segmentation.
The BDDV dataset~\cite{Xu2017} includes semantic segmentation annotation.
For some tasks, ground truth data acquisition is challenging or nearly impossible in the physical world (for instance, for optical flow estimation).
This motivates the use of simulated data for training and evaluating vision models, as in the SYNTHIA~\cite{Ros:2016}, Virtual KITTI~\cite{Gaidon:2016}, and GTA5 datasets~\cite{Richter:2016}, and the VIPER benchmark~\cite{Richter2017}.
These datasets and benchmarks are valuable for assessing the performance of different components of a vision pipeline, but they do not allow evaluation of a full driving system.

Recently, increased interest in end-to-end learning for driving has led to the emergence of datasets and benchmarks for the task of direct control signal prediction from observations (typically images).
To collect such a dataset, a vehicle is equipped with one or several cameras and additional sensors recording the coordinates, velocity, sometimes the control signal being executed, etc.
The Udacity dataset~\cite{UdacityDataset} contains recordings of lane following in highway and urban scenarios.
The CommaAI dataset~\cite{Santana:2016} includes $7$ hours of highway driving.
The Oxford RobotCar Dataset~\cite{Maddern2017} includes over $1000$ km of driving recoded under varying weather, lighting, and traffic conditions.
The BDDV dataset~\cite{Xu2017} is the largest publicly available urban driving dataset to date, with $10,\!000$ hours of driving recorded from forward-facing cameras together with smartphone sensor data such as GPS, IMU, gyroscope, and magnetometer readings.
These datasets provide useful training data for end-to-end driving systems.
However, due to their static nature (passive pre-recorded data rather than a living environment), they do not support evaluation of the actual driving performance of the learned models.

Online evaluation of driving models is technically challenging.
In the physical world, tests are typically restricted to controlled simple environments~\cite{Kahn:2017,Codevilla2018} and qualitative results~\cite{Pomerleau1988,Bojarski2016nvidiadriving}.
Large-scale real-world evaluations are impractical for the vast majority of researchers.
One alternative is simulation.
Due of its logistical feasibility, simulation have been commonly employed for driving research, especially in the context of machine learning.
The TORCS simulator~\cite{Wymann2014torcs} focuses on racing, and has been applied to evaluating road following~\cite{Chen:2015}.
Rich active environments provided by computer games have been used for training and evaluation of driving models~\cite{Ebrahimi17}; however, the available information and the controllability of the environment are typically limited in commercial games.
The recent CARLA driving simulator~\cite{Dosovitskiy2017} allows evaluating driving policies in living towns, populated with vehicles and pedestrians, under different weather and illumination conditions.
In this work we use CARLA to perform an extensive study of offline performance metrics for driving.

Although the analysis we perform is applicable to any vision-based driving pipeline (including ones that comprise separate perception~\cite{Ros:2015,Schenider:2016,Zhu:2016,Bresson:2017,Jin:2017} and control modules~\cite{Paden2016}), in this paper we focus on end-to-end trained models.
This line of work dates back to the ALVINN model of Pomerleau~\cite{Pomerleau1988}, capable of road following in simple environments.
More recently, LeCun et al.~\cite{LeCun2005driving} demonstrated collision avoidance with an end-to-end trained deep network.
Chen et al.~\cite{Chen:2015} learn road following in the TORCS simulator, by introducing an intermediate representation of ``affordances'' rather than going directly from pixels to actions.
Bojarski et al.~\cite{Bojarski2016nvidiadriving} train deep convolutional networks for lane following on a large real-world dataset and deploy the system on a physical vehicle.
Fernando et al.~\cite{Fernando:2017} use neural memory networks combining visual inputs and steering wheel trajectories to perform long-term planning, and use the CommaAI dataset to validate the method.
Hubschneider et al. \cite{Hubschneider:2017} incorporate turning signals as additional inputs to their DriveNet.
Codevilla et al.~\cite{Codevilla2018} propose conditional imitation learning, which allows imitation learning to scale to complex environments such as urban driving by conditioning action prediction on high-level navigation commands.
The growing interest in end-to-end learning for driving motivates our investigation of the associated evaluation metrics.

\section{Methodology}
\label{sec:method}
We aim to analyze the relation between offline prediction performance and online driving quality.
To this end, we train models using conditional imitation learning~\cite{Codevilla2018} in a simulated urban environment~\cite{Dosovitskiy2017}.
We then evaluate the driving quality on goal-directed navigation and correlate the results with multiple offline prediction-based metrics.
We now describe the methods used to train and evaluate the models.

\subsection{Conditional Imitation Learning}
For training the models we use conditional imitation learning~-- a variant of imitation learning that allows providing high-level commands to a model.
When coupled with a high-level topological planner, the method can scale to complex navigation tasks such as driving in an urban environment.
We briefly review the approach here and refer the reader to Codevilla et al.~\cite{Codevilla2018} for further details.

We start by collecting a training dataset of tuples $\{\tuple{\obs_i, \cmd_i, \act_i}\}$, each including an observation $\obs_i$, a command $\cmd_i$, and an action $\act_i$.
The observation $\obs_i$ is an image recorded by a camera mounted on a vehicle.
The command $\cmd_i$ is a high-level navigation instruction, such as ``turn left at the next intersection''.
We use four commands~-- \cmdcontinue, \cmdstraight, \cmdleft, and \cmdright~-- encoded as one-hot vectors.
Finally, $\act_i$ is a vector representing the action executed by the driver.
It can be raw control signal~-- steering angle, throttle, and brake~-- or a more abstract action representation, such as a waypoint representing the intended trajectory of the vehicle.
We focus on predicting the steering angle in this work.

Given the dataset, we train a convolutional network $\net$ with learnable parameters $\params$ to perform command-conditional action prediction, by minimizing the average prediction loss:
\begin{equation}
\params^{*} = \argmin_{\params} \sum_i \ell(\net(\obs_i,\cmd_i,\params),\, \act_i),
\end{equation}
where $\ell$ is a per-sample loss.
We experiment with several architectures of the network $\net$, all based on the branched model of Codevilla et al.~\cite{Codevilla2018}.
Training techniques and network architectures are reviewed in more detail in section~\ref{sec:model_training}.
Further details of training are provided in the supplement.


\subsection{Training} \label{sec:model_training}
\mypara{Data collection.}
We collect a training dataset by executing an automated navigation expert in the simulated environment.
The expert makes use of privileged information about the environment, including the exact map of the environment and the exact positions of the ego-car, all other vehicles, and pedestrians.
The expert keeps a constant speed of 35 km/h when driving straight and reduces the speed when making turns.
We record the images from three cameras: a forward-facing one and two lateral cameras facing $30$ degrees left and right.
In $10\%$ of the data we inject noise in the driving policy to generate examples of recovery from perturbations.
In total we record $80$ hours of driving data.

\mypara{Action representation.}
The most straightforward approach to end-to-end learning for driving is to output the raw control command, such as the steering angle, directly~\cite{Bojarski2016nvidiadriving,Codevilla2018}.
We use this representation in most of our experiments.
The action is then a vector $\act \in \mathbb{R}^3$, consisting of the steering angle, the throttle value, and the brake value.
To simplify the analysis and preserve compatibility with prior work~\cite{Bojarski2016nvidiadriving,Xu2017}, we only predict the steering angle with a deep network.
We use throttle and brake values provided by the expert policy described above.


\mypara{Loss function.}
In most of our experiments we follow standard practice~\cite{Bojarski2016nvidiadriving,Codevilla2018} and use mean squared error (MSE) as a per-sample loss:
\begin{equation}
 \ell(\net(\obs_i,\cmd_i,\params),\, \act_i)  = \norm{\net(\obs_i,\cmd_i,\params) - \act_i}^2.
\end{equation}
We have also experimented with the L1 loss.
In most experiments we balance the data during training.
We do this by dividing the data into $8$ bins based on the ground-truth steering angle and sampling an equal number of datapoints from each bin in every mini-batch.
As a result, the loss being optimized is not the average MSE over the dataset, but its weighted version with higher weight given to large steering angles.

\mypara{Regularization.}
Even when evaluating in the environment used for collecting the training data, a driving policy needs to generalize to previously unseen views of this environment.
Generalization is therefore crucial for a successful driving policy.
We use dropout and data augmentation as regularization measures when training the networks.

Dropout ratio is $0.2$ in convolutional layers and $0.5$ in fully-connected layers.
For each image to be presented to the network, we apply a random subset of a set of transformations with randomly sampled magnitudes.
Transformations include contrast change, brightness, and tone, as well as the addition of Gaussian blur, Gaussian noise, salt-and-pepper noise, and region dropout (masking out a random set of rectangles in the image, each rectangle taking roughly 1\% of image area).
In order to ensure good convergence, we found it helpful to gradually increase the data augmentation magnitude in proportion to the training step.
Further details are provided in the supplement.

%
\mypara{Model architecture.}
We experiment with a feedforward convolutional network, which takes as input the current observation as well as an additional vector of measurements (in our experiments the only measurement is the current speed of the vehicle).
This network implements a purely reactive driving policy, since by construction it cannot make use of temporal context.
We experiment with three variants of this model.
The architecture used by Codevilla et al.~\cite{Codevilla2018}, with $8$ convolutional layers, is denoted as ``standard''.
We also experiment with a deeper architecture with $12$ convolutional layers and a shallower architecture with $4$ convolutional layers.



\subsection{Performance metrics} \label{sec:methods_offline_metrics}
\mypara{Offline error metrics.}
Assume we are given a validation set $\vV$ of tuples $\tuple{\obs_i,\cmd_i,\act_i, v_i}$, indexed by $i \in V$.
Each tuple includes an observation, an input command, a ground-truth action vector, and the speed of the vehicle.
We assume the validation set consists of one or more temporally ordered driving sequences.
(For simplicity in what follows we assume it is a single sequence, but generalization to multiple sequences is trivial.)
Denote the action predicted by the model by $\apred_i = \net(\obs_i,\cmd_i,\params)$.
In our experiments, $\agt_i$ and $\apred_i$ will be scalars, representing the steering angle.
Speed is also a scalar (in m/s).

\begin{table}[]
\centering
\caption{Offline metrics used in the evaluation. $\delta$ is the Kronecker delta function, $\theta$ is the Heaviside step function, $Q$ is a quantization function (see text for details),  $\abs{V}$ is the number of samples in the validation dataset. }
{\small
\ra{1.2}
\resizebox{0.95\linewidth}{!}{
\setlength{\tabcolsep}{6pt}
 \begin{tabular}{lcl}
\toprule
  Metric name                               & Parameters & Metric definition \\ \midrule
  Squared error                             & --          & $\frac{1}{|V|}\sum\limits_{i \in V} \norm{\agt_i - \apred_i}^2$ \\
  Absolute error                            & --          & $\frac{1}{|V|}\sum\limits_{i \in V} \norm{\agt_i - \apred_i}_1$ \\
  Speed-weighted absolute error             & --          & $\frac{1}{|V|}\sum\limits_{i \in V} \norm{\agt_i - \apred_i}_1 v_i$ \\
  Cumulative speed-weighted absolute error  & $T$        & $\frac{1}{|V|}\sum\limits_{i \in V} \norm{\sum\limits_{t=0}^T (\agt_{i+t} - \apred_{i+t}) v_{i+t} }_1$ \\
  Quantized classification error            & $\sigma$   & $\frac{1}{|V|}\sum\limits_{i \in V} \left( 1- \delta \left(Q(\agt_i, \sigma), Q(\apred_i, \sigma) \right)\right)$ \\
  Thresholded relative error                & $\alpha$   & $\frac{1}{|V|}\sum\limits_{i \in V} \theta\left(\norm{\apred_i - \agt_i} - \alpha \norm{\agt_i} \right)$ \\
\bottomrule
 \end{tabular}
}
}
\label{tbl:offline_metrics}
\end{table}

Table~\ref{tbl:offline_metrics} lists offline metrics we evaluate in this paper.
The first two metrics are standard: mean squared error (which is typically the training loss) and absolute error.
Absolute error gives relatively less weight to large mistakes than MSE.

The higher the speed of the car, the larger the impact a control mistake can have.
To quantify this intuition, we evaluate speed-weighted absolute error.
This metric approximately measures how quickly the vehicle is diverging from the ground-truth trajectory, that is, the projection of the velocity vector onto the direction orthogonal to the heading direction.

We derive the next metric by accumulating speed-weighted errors over time.
The intuition is that the average prediction error may not be characteristic of the driving quality, since it does not take into account the temporal correlations in the errors.
Temporally uncorrelated noise may lead to slight oscillations around the expert trajectory, but can still result in successful driving.
In contrast, a consistent bias in one direction for a prolonged period of time inevitably leads to a crash.
We therefore accumulate the speed-weighted difference between the ground-truth action and the prediction over $T$ time steps.
This measure is a rough approximation of the divergence of the vehicle from the desired trajectory over $T$ time steps.

Another intuition is that small noise may be irrelevant for the driving performance, and what matters is getting the general direction right.
Similar to Xu et al.~\cite{Xu2017}, we quantize the predicted actions and evaluate the classification error.
For quantization, we explicitly make use of the fact that the actions are scalars (although a similar strategy can be applied to higher-dimensional actions).
Given a threshold value $\sigma$, the quantization function $Q(x,\sigma)$ returns $-1$ if $x < -\sigma$, $0$ if $-\sigma \leq x < \sigma$, and $1$ if $x \geq \sigma$.
For steering angle, these values correspond to going left, going straight, and going right.
Given the quantized predictions and the ground truth, we compute the classification error.

Finally, the last metric is based on quantization and relative errors.
Instead of quantizing with a fixed threshold as in the previous metric, here the threshold is adaptive, proportional to the ground truth steering signal.
The idea is that for large action values, small discrepancies with the ground truth are not as important as for small action values.
Therefore, we count the fraction of samples for which $\norm{\apred_i - \agt_i} \geq \alpha \norm{\agt_i}$.

\mypara{Online performance metrics.}
We measure the driving quality using three metrics.
The first one is the success rate, or simply the fraction of successfully completed navigation trials.
The second is the average fraction of distance traveled towards the goal per episode (this value can be negative is the agent moves away form the goal).
The third metric measures the average number of kilometers traveled between two infractions. (Examples of infractions are collisions, driving on the sidewalk, or driving on the opposite lane.)

\section{Experiments}
\label{sec:experiments}
We perform an extensive study of the relation between online and offline performance of driving models.
Since conducting such experiments in the real world would be impractical, the bulk of the experiments are performed in the CARLA simulator~\cite{Dosovitskiy2017}.
We start by training a diverse set of driving models with varying architecture, training data, regularization, and other parameters.
We then correlate online driving quality metrics with offline prediction-based metrics, aiming to find offline metrics that are most predictive of online driving performance.
Finally, we perform an additional analysis on the real-world BDDV dataset.
Supplementary materials can be found on the project page:~\url{https://sites.google.com/view/evaluatedrivingmodels}.

\subsection{Experimental setup}
\mypara{Simulation.}
We use the CARLA simulator to evaluate the performance of driving models in an urban environment.
We follow the testing protocol of Codevilla et al.~\cite{Codevilla2018} and Dosovitskiy et al.~\cite{Dosovitskiy2017}.
We evaluate goal-directed navigation with dynamic obstacles.
One evaluation includes $25$ goal-directed navigation trials.

CARLA provides two towns (Town 1 and Town 2) and configurable weather and lighting conditions.
We make use of this capability to evaluate generalization of driving methods.
We use Town 1 in 4 weathers (Clear Noon, Heavy Rain Noon, Clear Sunset and Clear After Rain) for training data collection, and we use two test conditions: Town 1 in clear noon weather and Town 2 in Soft Rain Sunset weather.
The first condition is present in the training data; yet, note that the specific images observed when evaluating the policies have almost certainly not been seen during training.
Therefore even this condition requires generalization.
The other condition~-- Town 2 and soft rain sunset weather~-- is completely new and requires strong generalization.

For validation we use $2$ hours of driving data with action noise and $2$ hours of data without action noise, in each of the conditions.
With three cameras and a frame rate of $10$ frames per second, one hour of data amounts to $108@000$ validation images.

\mypara{Real-world data.}
For real-world tests we use the validation set of the BDDV dataset~\cite{Xu2017},
containing $1816$ dashboard camera videos.
We computed the offline metrics over the entire dataset using the pre-trained models and the data filtering procedures provided by Xu et al.~\cite{Xu2017}.

\mypara{Network training and evaluation.}
All models were trained using the Adam optimizer~\cite{Kingma2015adam} with minibatches of $120$ samples and an initial learning rate of $10^{-4}$.
We reduce the learning rate by a factor of $2$ every $50K$ iterations.
All models were trained up to $500K$ iterations.
In order to track the evolution of models during the course of training, for each model we perform both online and offline evaluation after the following numbers of training mini-batches: 2K, 4K, 8K, 16K, 32K, 64K, 100K, 200K, 300K, 400K, and 500K.

\subsection{Evaluated models}
We train a total of $45$ models.
The parameters we vary can be broadly separated into three categories: properties of the training data, of the model architecture, and of the training procedure.
We vary the amount and the distribution of the training data.
The amount varies between $0.2$ hours and $80$ hours of driving.
The distribution is one of the following four: all data collected from three cameras and with noise added to the control, only data from the central camera, only data without noise, and data from the central camera without noise.
The model architecture variations amount to varying the depth between $4$ and $12$ layers.
The variations in the training process are the use of data balancing, the loss function, and the regularization applied (dropout and the level of data augmentation).
A complete list of parameters varied during the evaluation is provided in the supplement.

\subsection{Correlation between offline and online metrics}
We start by studying the correlation between online and offline performance metrics on the whole set of evaluated models.
We represent the results by scatter plots and correlation coefficients.
To generate a scatter plot, we select two metrics and plot each evaluated model as a circle, with the coordinates of the center of the circle equal to the values of these two metrics, and the radius of the circle proportional to the training iteration the model was evaluated at.
To quantify the correlations, we use the standard sample Pearson correlation coefficient, computed over all points in the plot.
In the figures below, we plot results in generalization conditions (Town 2, unseen weather).
We focus our analysis on the well-performing models, by discarding the 50\% worst models according to the offline metric.
Results in training conditions, as well as scatter plots with all models, are shown in the supplement.


\mypara{The effect of validation data.}
We first plot the (offline) average steering MSE versus the (online) success rate in goal-directed navigation, for different offline validation datasets.
We vary the number of cameras used for validation (just a forward-facing camera or three cameras including two lateral ones) and the presence of action noise in the validation set.
This experiment is inspired by the fact that
the 3-camera setup and the addition of noise have been advocated for training end-to-end driving models~\cite{Xu2017,Bojarski2016nvidiadriving,Dosovitskiy2017,Codevilla2018}.

\begin{figure}
  \centering
  { \fontsize{7pt}{9pt}\selectfont
  \begin{tabular}{ccc}
    \quad Central camera, no noise & \quad Central camera, with noise & \quad Three cameras, no noise \\
    \includegraphics[height=0.32\linewidth]{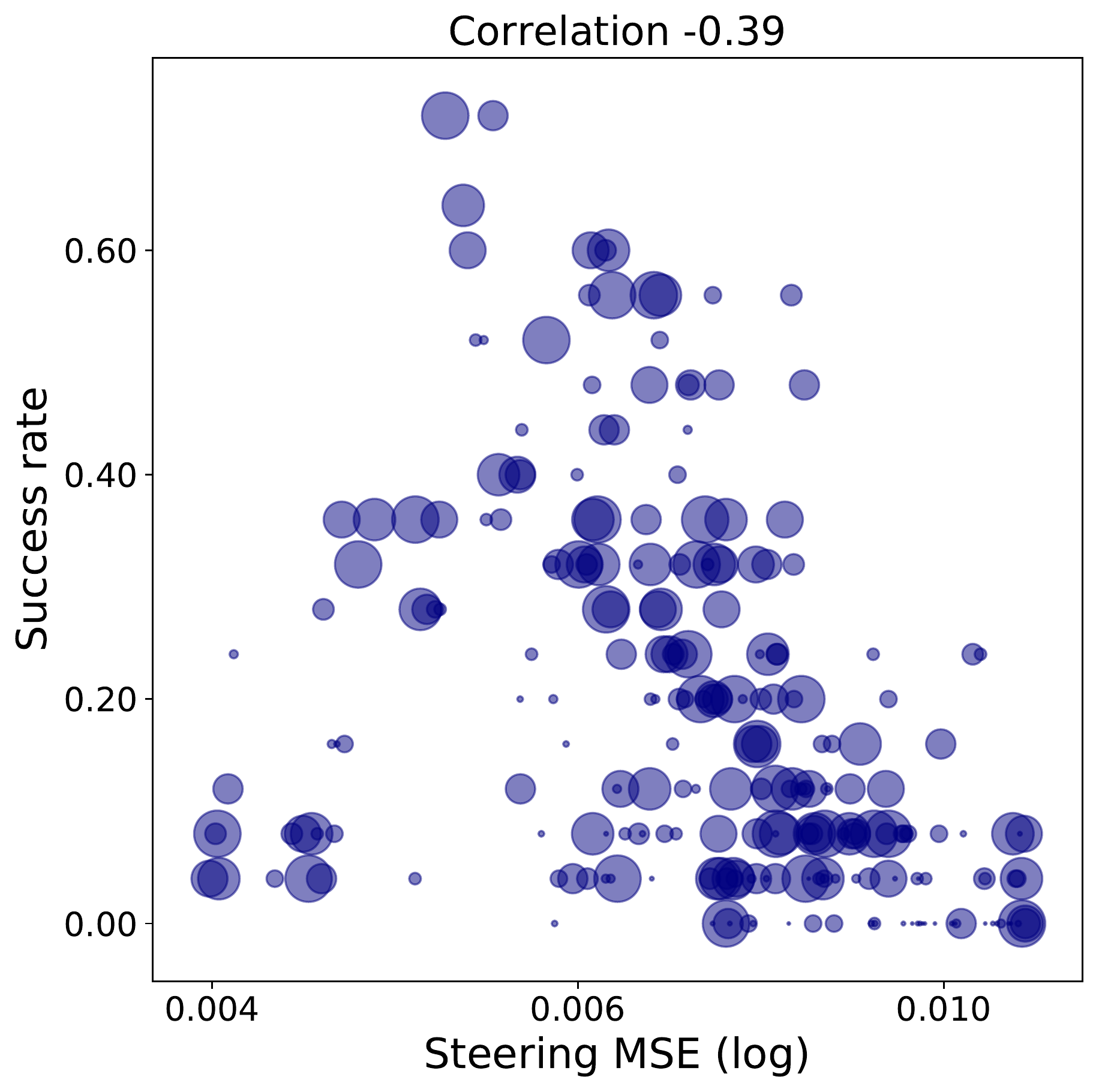} &
    \includegraphics[height=0.32\linewidth]{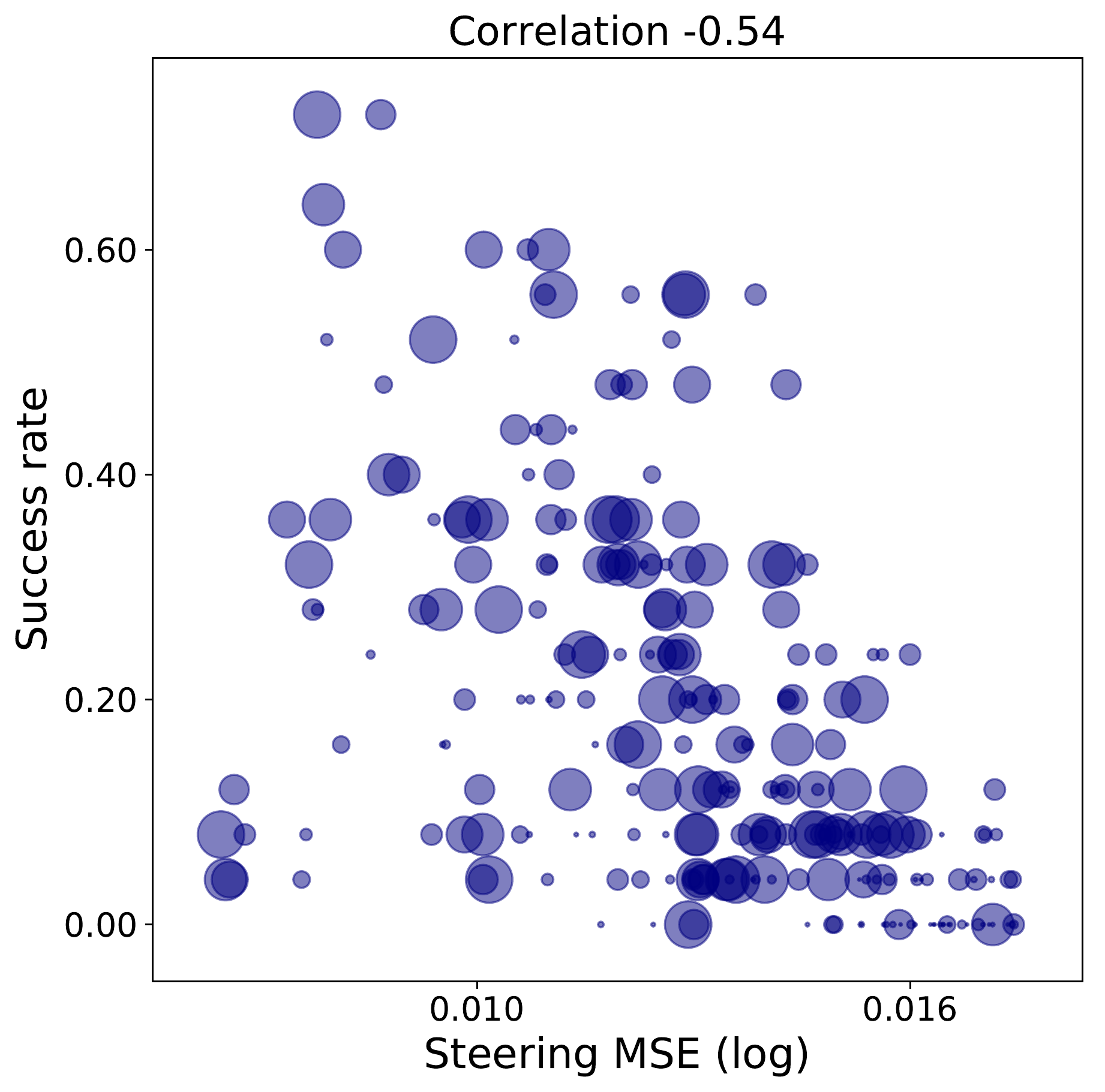} &
    \includegraphics[height=0.32\linewidth]{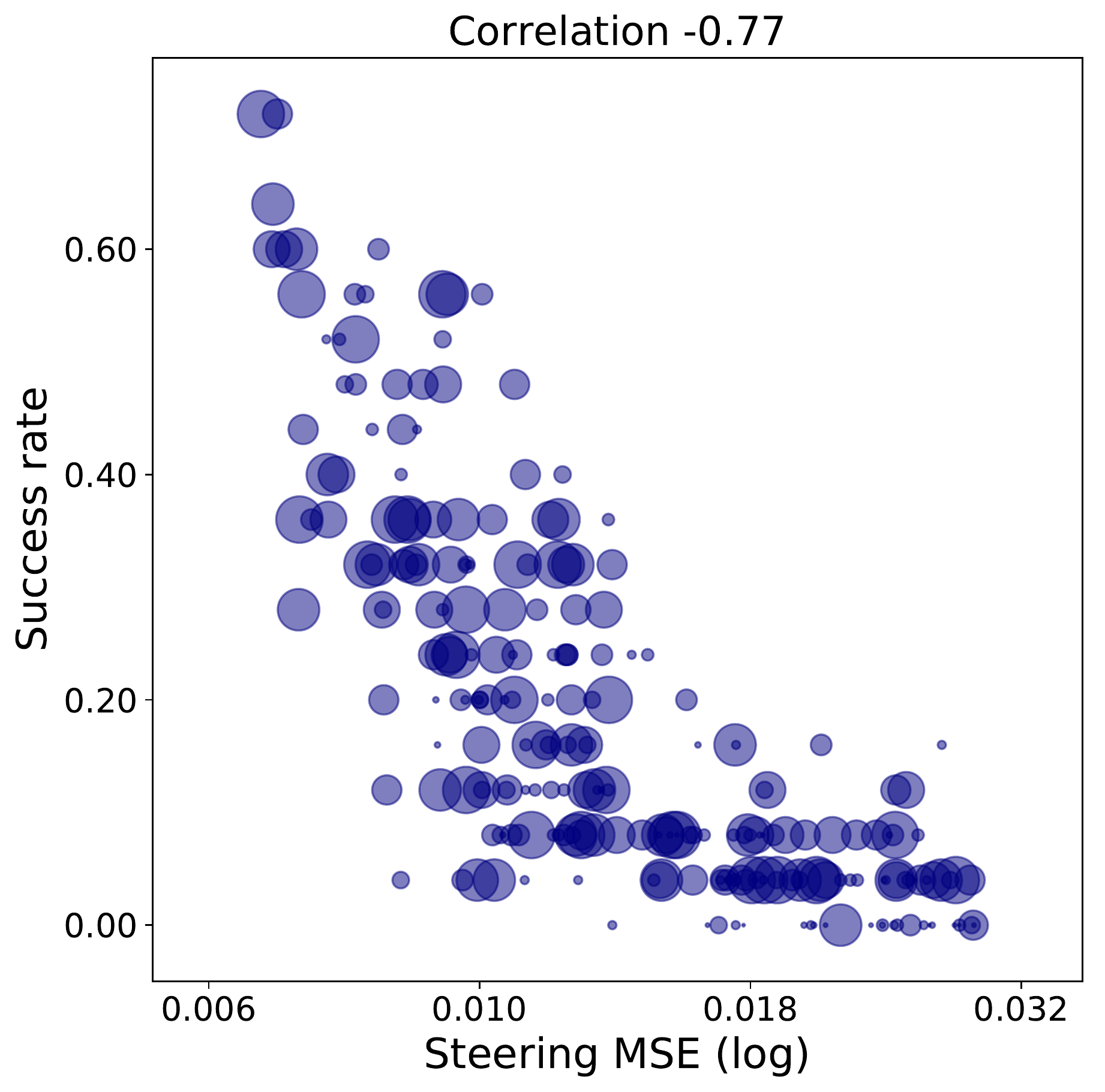}
  \end{tabular}
  }
  \caption{Scatter plots of goal-directed navigation success rate vs.\ steering MSE when evaluated on data from different distributions. We evaluate the models in the generalization condition (Town 2) and we plot the 50\% best-performing models according to the offline metric. Sizes of the circles denote the training iterations at which the models were evaluated. We additionally show the sample Pearson correlation coefficient for each plot. Note how the error on the basic dataset (single camera, no action noise) is the least informative of the driving performance.}
  \label{fig:scatter_steer_vary_data}
\end{figure}

The results are shown in Figure~\ref{fig:scatter_steer_vary_data}.
The most striking observation is that the correlation between offline prediction and online performance is weak.
For the basic setup~-- central camera and no action noise~-- the absolute value of the correlation coefficient is only $0.39$.
The addition of action noise improves the correlation to $0.54$.
Evaluating on data from three cameras brings the correlation up to $0.77$.
This shows that a successful policy must not only predict the actions of an expert on the expert's trajectories, but also for observations away from the expert's trajectories.
Proper validation data should therefore include examples of recovery from perturbations.


\mypara{Offline metrics.}
Offline validation data from three cameras or with action noise may not always be available.
Therefore, we now aim to find offline metrics that are predictive of driving quality even when evaluated in the basic setup with a single forward-facing camera and no action noise.

\begin{figure}
  \centering
  { \fontsize{7pt}{9pt}\selectfont
  \begin{tabular}{ccc}
    \quad Steering MSE & \quad Steering absolute error & \quad Speed-weighted error \\
    \includegraphics[height=0.32\linewidth]{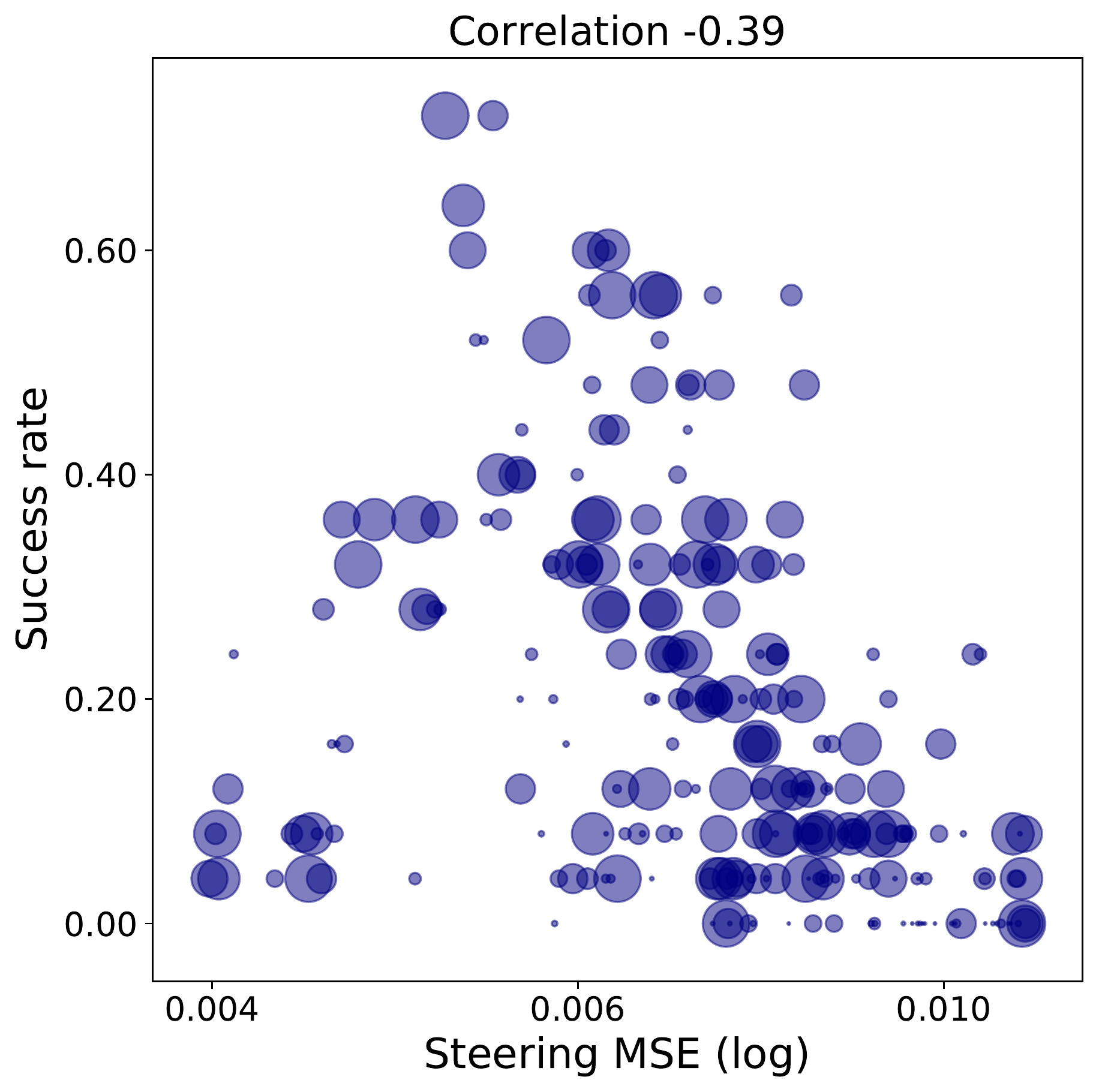} &
    \includegraphics[height=0.32\linewidth]{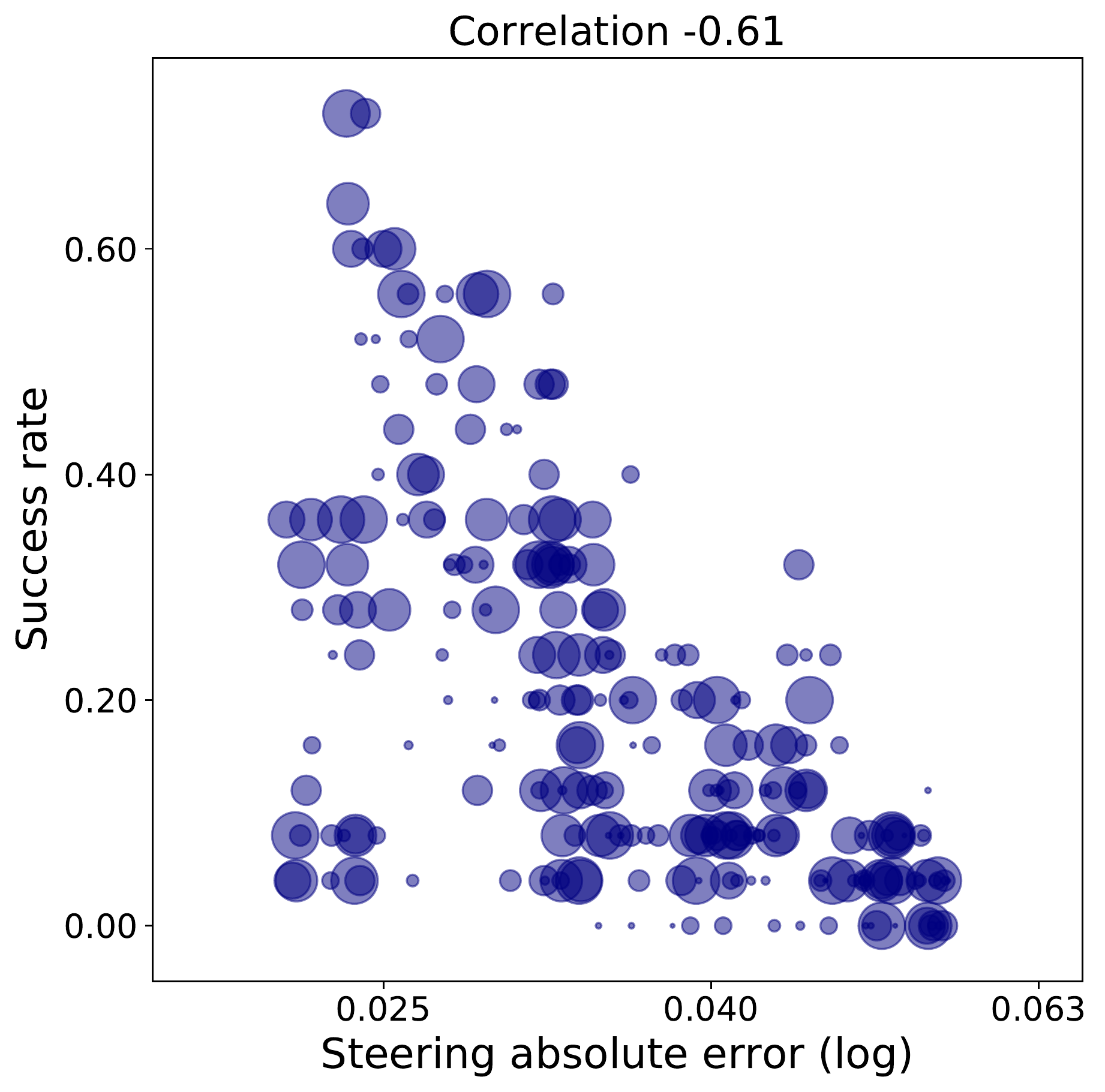} &
    \includegraphics[height=0.32\linewidth]{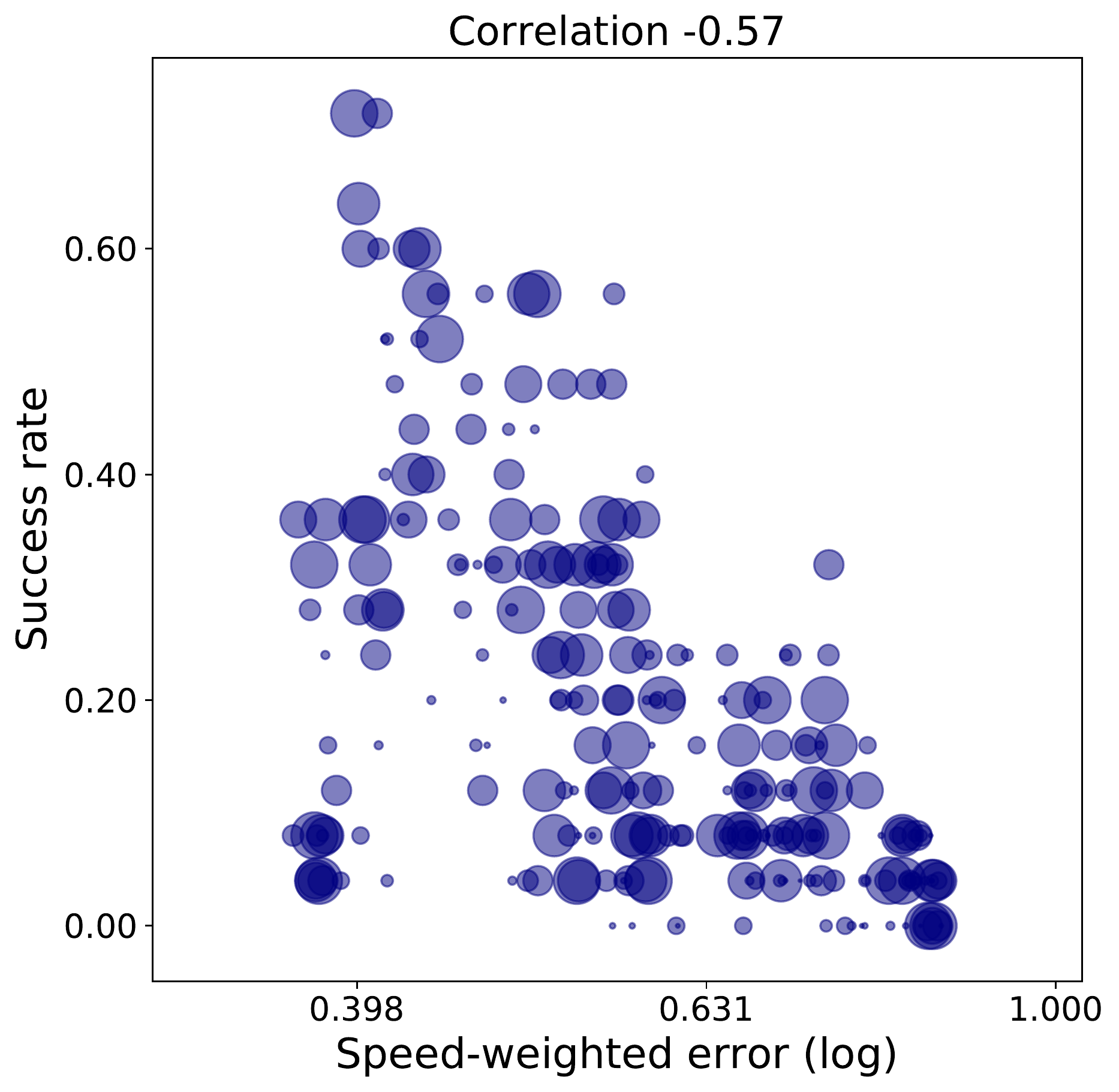} \\
    \quad Cumulative error & \quad Quantized classification & \quad Thresholded relative error \\
    \includegraphics[height=0.32\linewidth]{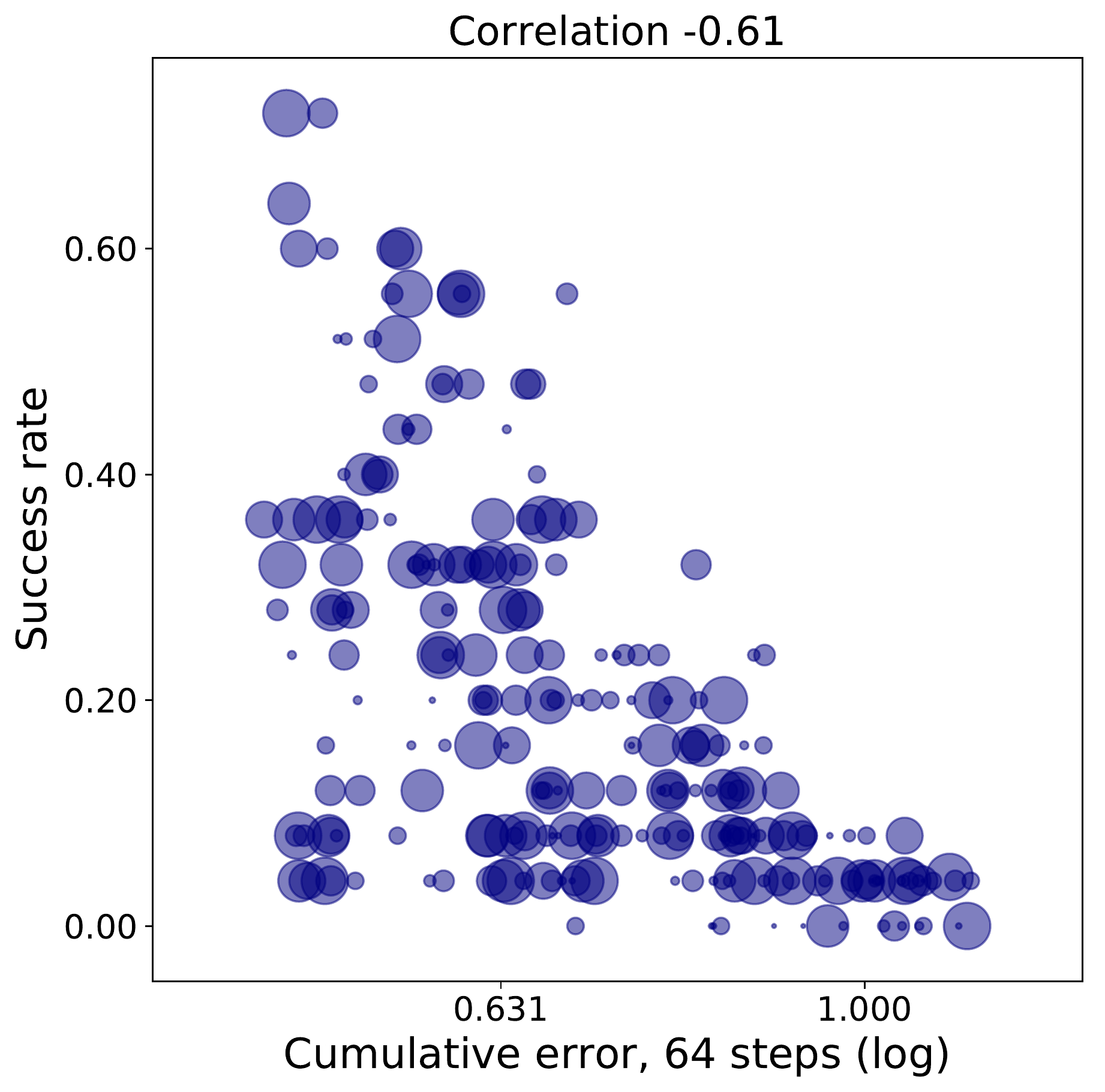} &
    \includegraphics[height=0.32\linewidth]{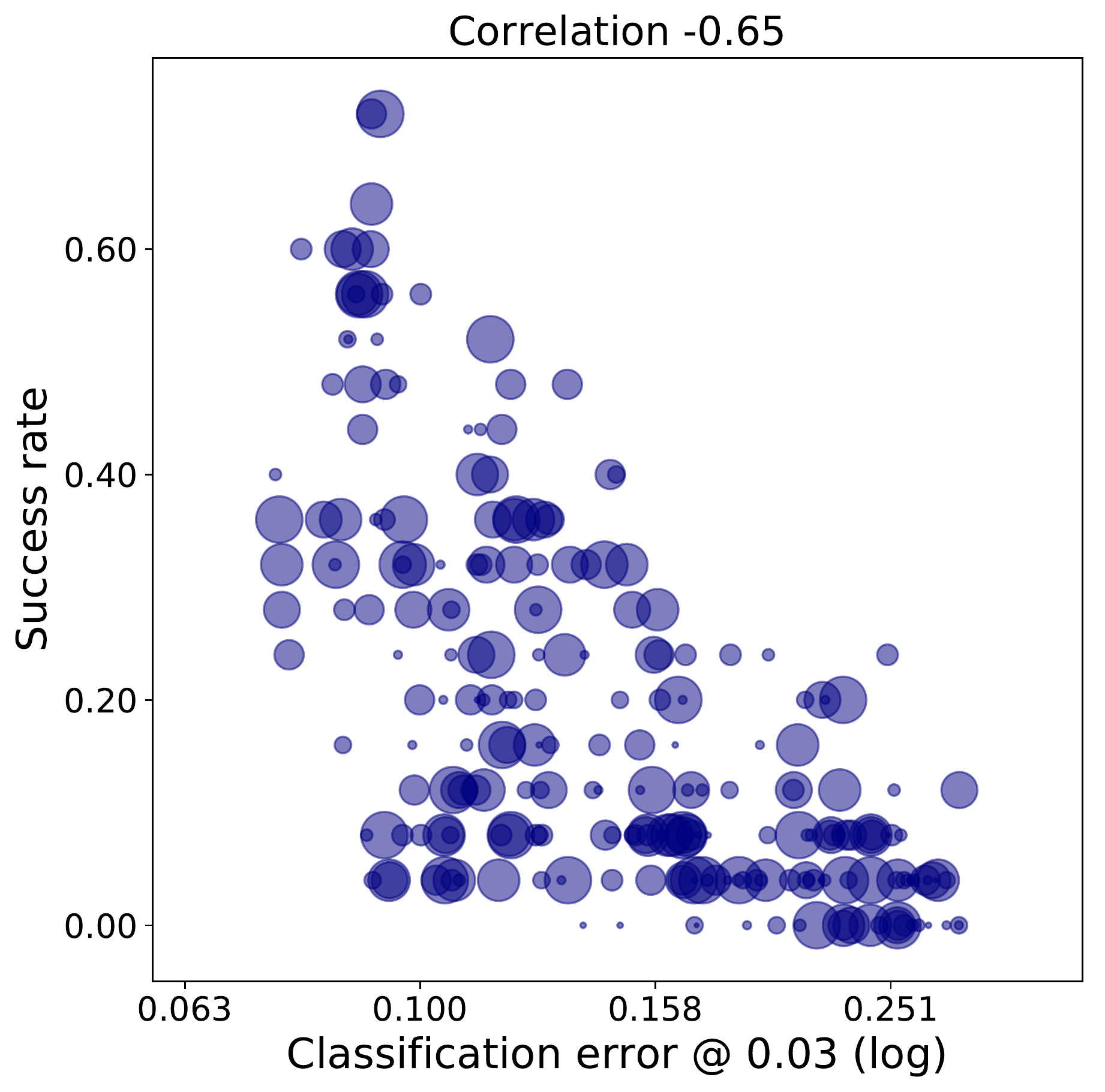} &
    \includegraphics[height=0.32\linewidth]{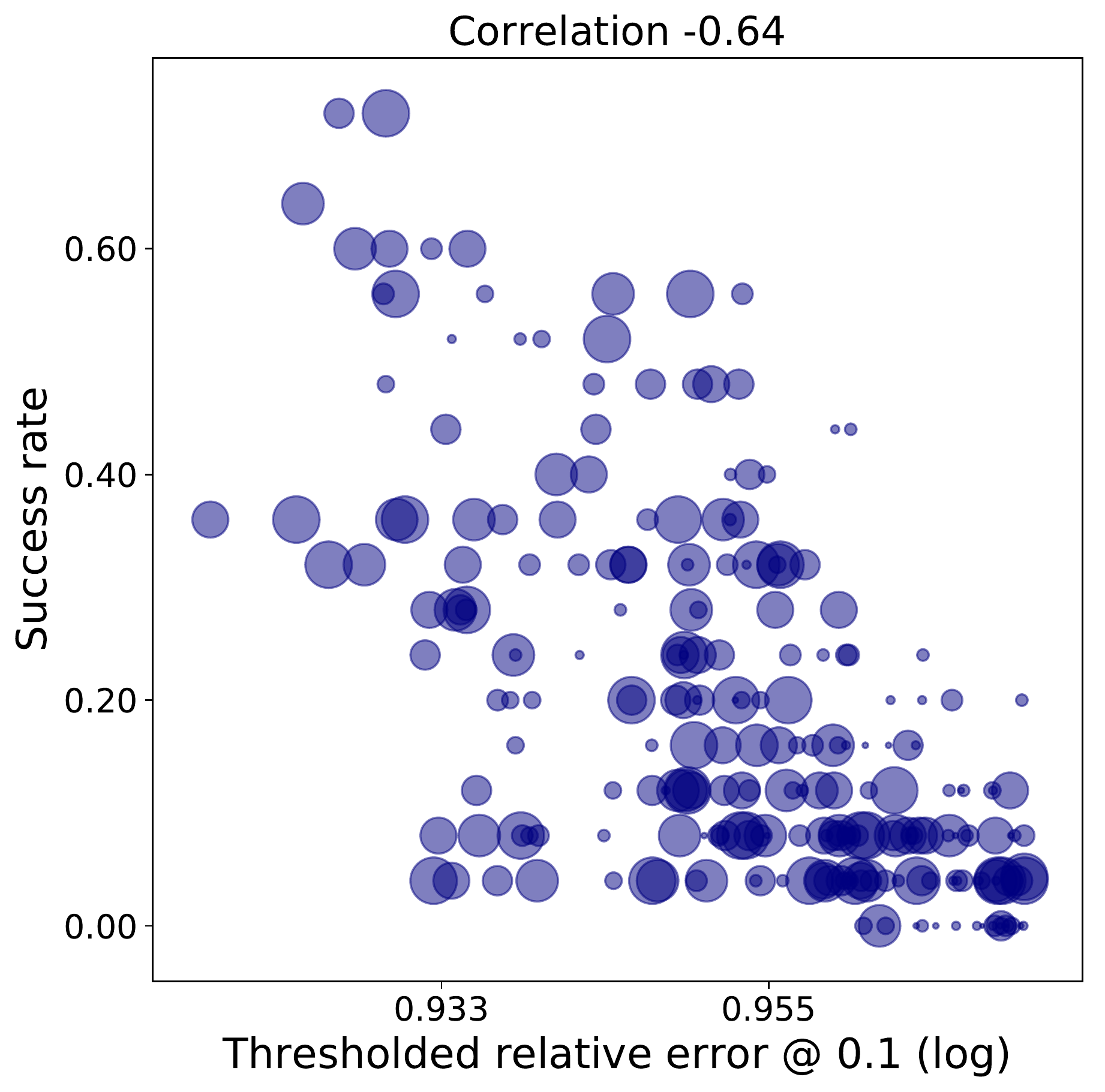} \\
  \end{tabular}
  }
  \caption{Scatter plots of goal-directed navigation success rate vs.\ different offline metrics. We evaluate the models in the generalization condition (Town 2) and we plot the 50\% best-performing models according to the offline metric. Note how correlation is generally weak, especially for mean squred error (MSE).}
  \label{fig:scatter_offline_metrics}
\end{figure}

Figure~\ref{fig:scatter_offline_metrics} shows scatter plots of offline metrics described in Section~\ref{sec:methods_offline_metrics}, versus the navigation success rate.
MSE is the least correlated with the driving success rate: the absolute value of the correlation coefficient is only $0.39$.
Absolute steering error is more strongly correlated, at $0.61$.
Surprisingly, weighting the error by speed or accumulating the error over multiple subsequent steps does not improve the correlation.
Finally, quantized classification error and thresholded relative error are also more strongly correlated, with the absolute value of the correlation coefficient equal to $0.65$ and $0.64$, respectively.

\mypara{Online metrics.}
So far we have looked at the relation between offline metrics and a single online metric~-- success rate.
Is success rate fully representative of actual driving quality?
Here we compare the success rate with two other online metrics: average fraction of distance traveled towards the goal and average number of kilometers traveled between two infractions.

\begin{figure}
  { \fontsize{7pt}{9pt}\selectfont
  \centering
  \begin{tabular}{ccc}
    \quad Success rate vs Avg. completion & \quad Km per infraction vs Success rate & \quad Km per infraction vs Avg. completion \\
    \includegraphics[height=0.32\linewidth]{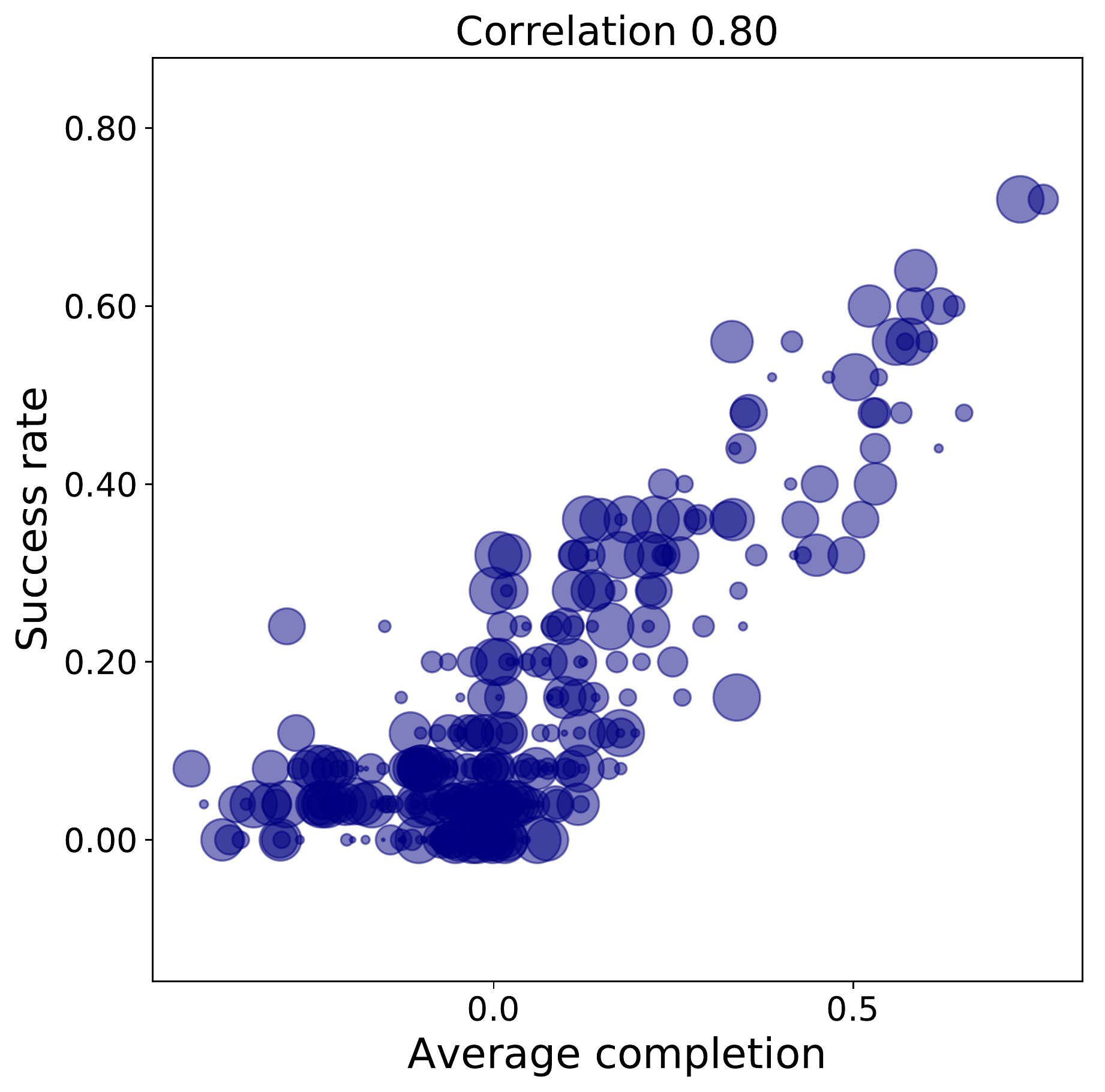} &
    \includegraphics[height=0.32\linewidth]{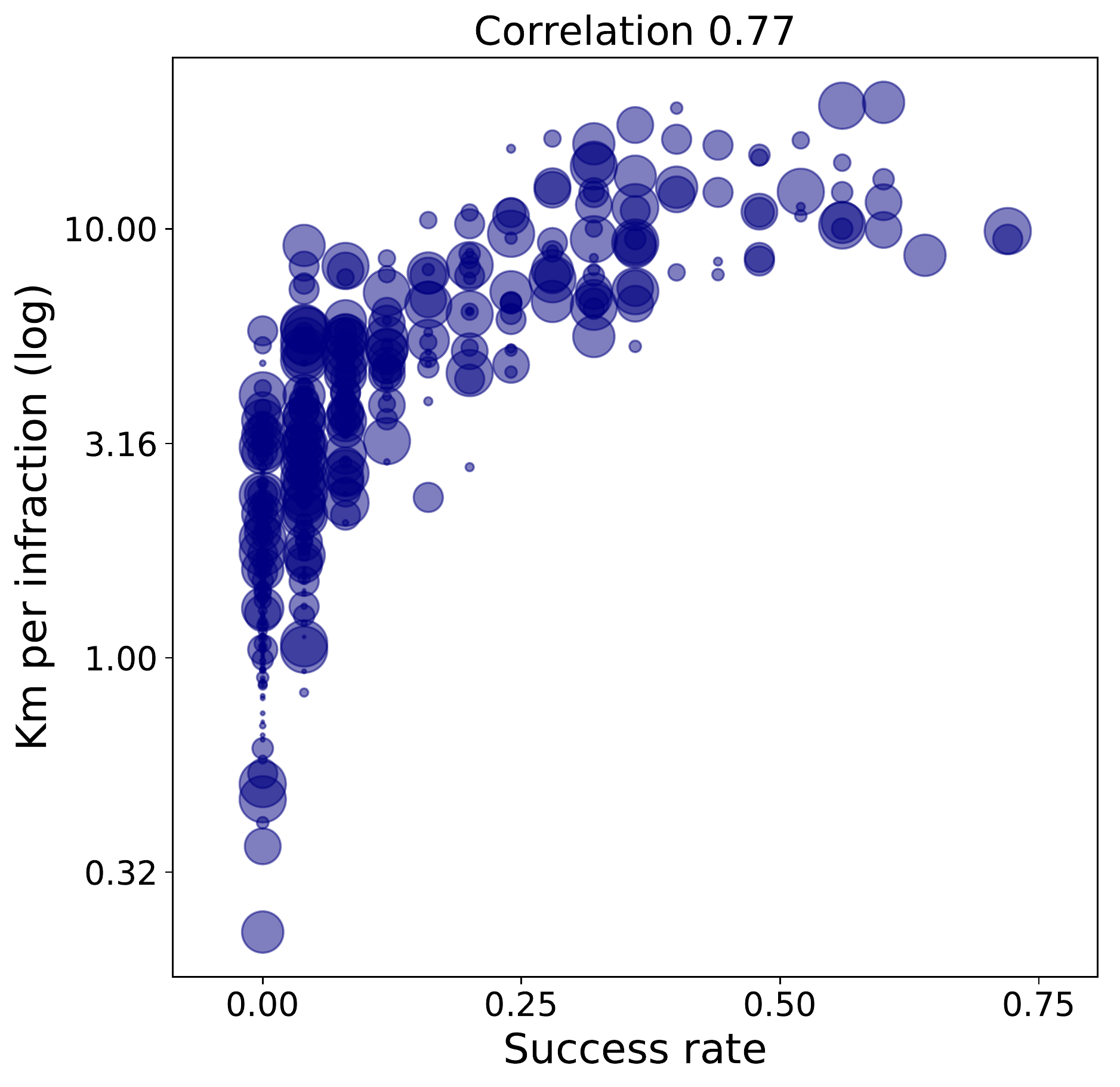} &
    \includegraphics[height=0.32\linewidth]{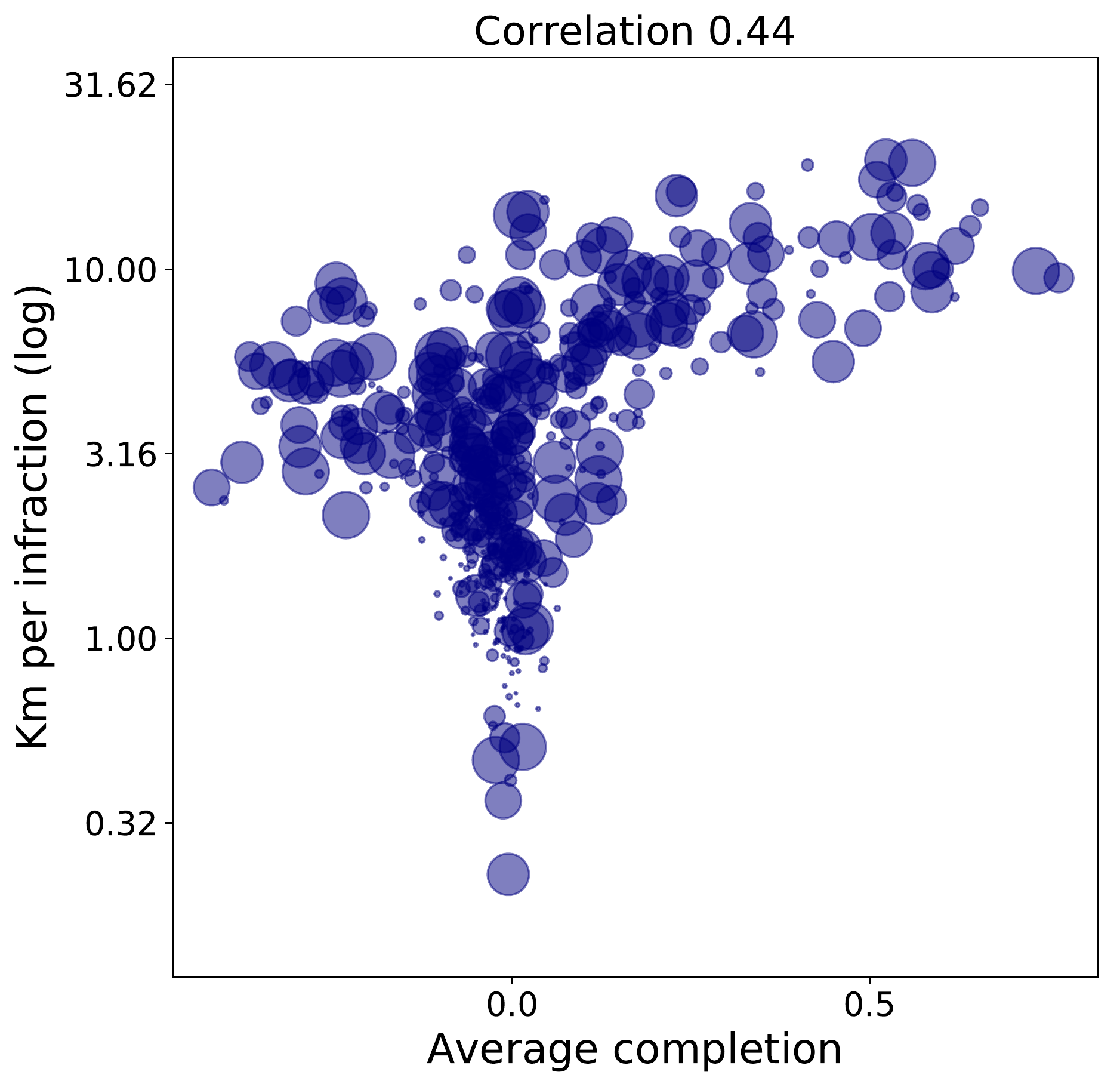}
  \end{tabular}
  }
  \caption{Scatter plots of online driving quality metrics versus each other. The metrics are: success rate, average fraction of distance to the goal covered (average completion), and average distance (in km) driven between two infractions. Success rate is strongly correlated with the other two metrics, which justifies its use as the main online metric in our analysis.}
  \label{fig:scatter_online_metrics}
\end{figure}

Figure~\ref{fig:scatter_online_metrics} shows pairwise scatter plots of these three online metrics.
Success rate and average completion are strongly correlated, with a correlation coefficient of $0.8$.
The number of kilometers traveled between two infractions is similarly correlated with the success rate ($0.77$), but much less correlated with the average completion ($0.44$).
We conclude that online metrics are not perfectly correlated and it is therefore advisable to measure several online metrics when evaluating driving models.
Success rate is well correlated with the other two metrics, which justifies its use as the main online metric in our analysis.

\mypara{Case study.} We have seen that even the best-correlated offline and online metrics have a correlation coefficient of only $0.65$.
Aiming to understand the reason for this remaining discrepancy, here we take a closer look at two models which achieve similar prediction accuracy, but drastically different driving quality.
The first model was trained with the MSE loss and forward-facing camera only.
The second model used the L1 loss and three cameras.
We refer to these models as Model 1 and Model 2, respectively.

\begin{figure}
  \centering
  { \fontsize{8pt}{10pt}\selectfont
  \setlength{\tabcolsep}{4pt}
  \begin{tabular}{cccc}
    \multicolumn{3}{c}{Steering angle prediction vs Time} & Zoom-in of one turn \\
    \multicolumn{3}{c}{\includegraphics[height=0.25\linewidth]{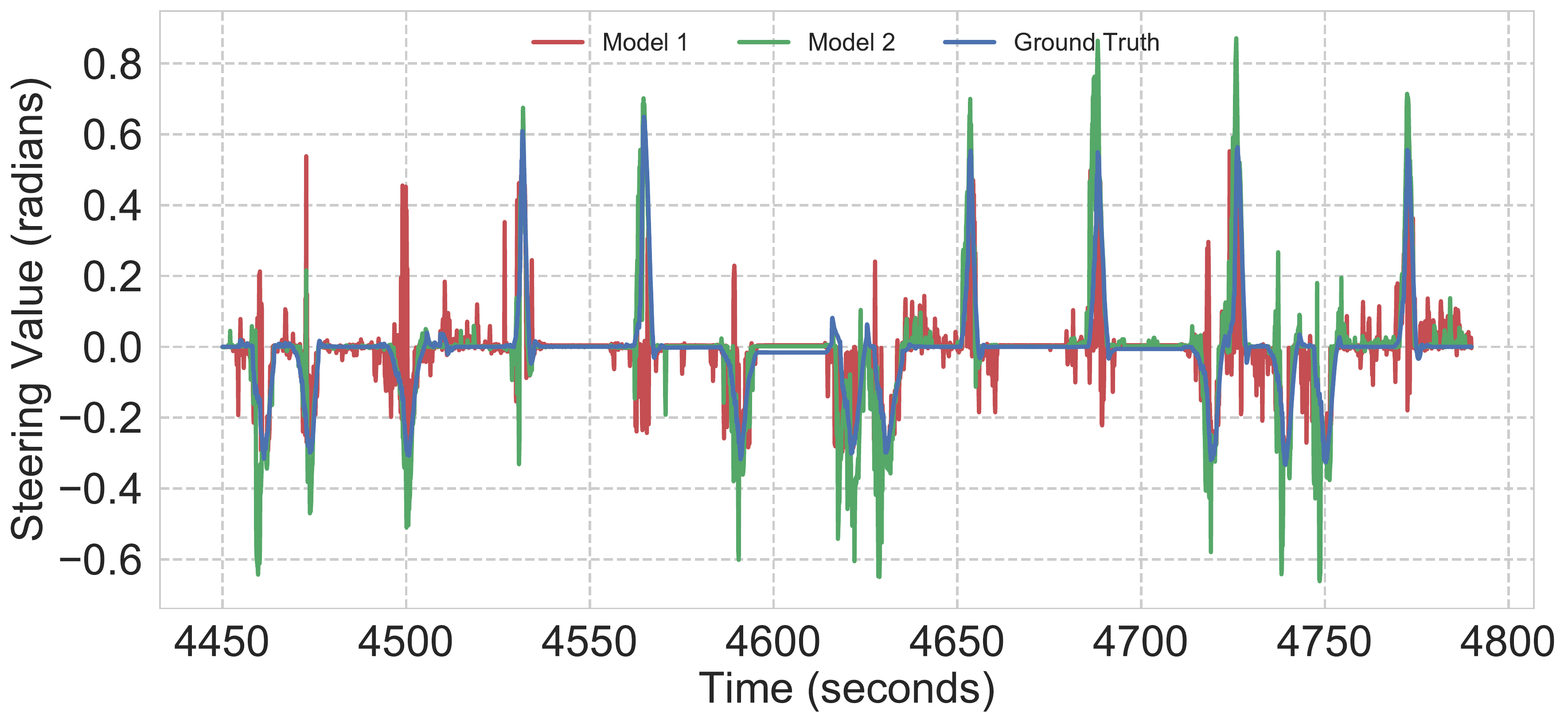}}\hspace{-5mm}&
    \includegraphics[height=0.25\linewidth]{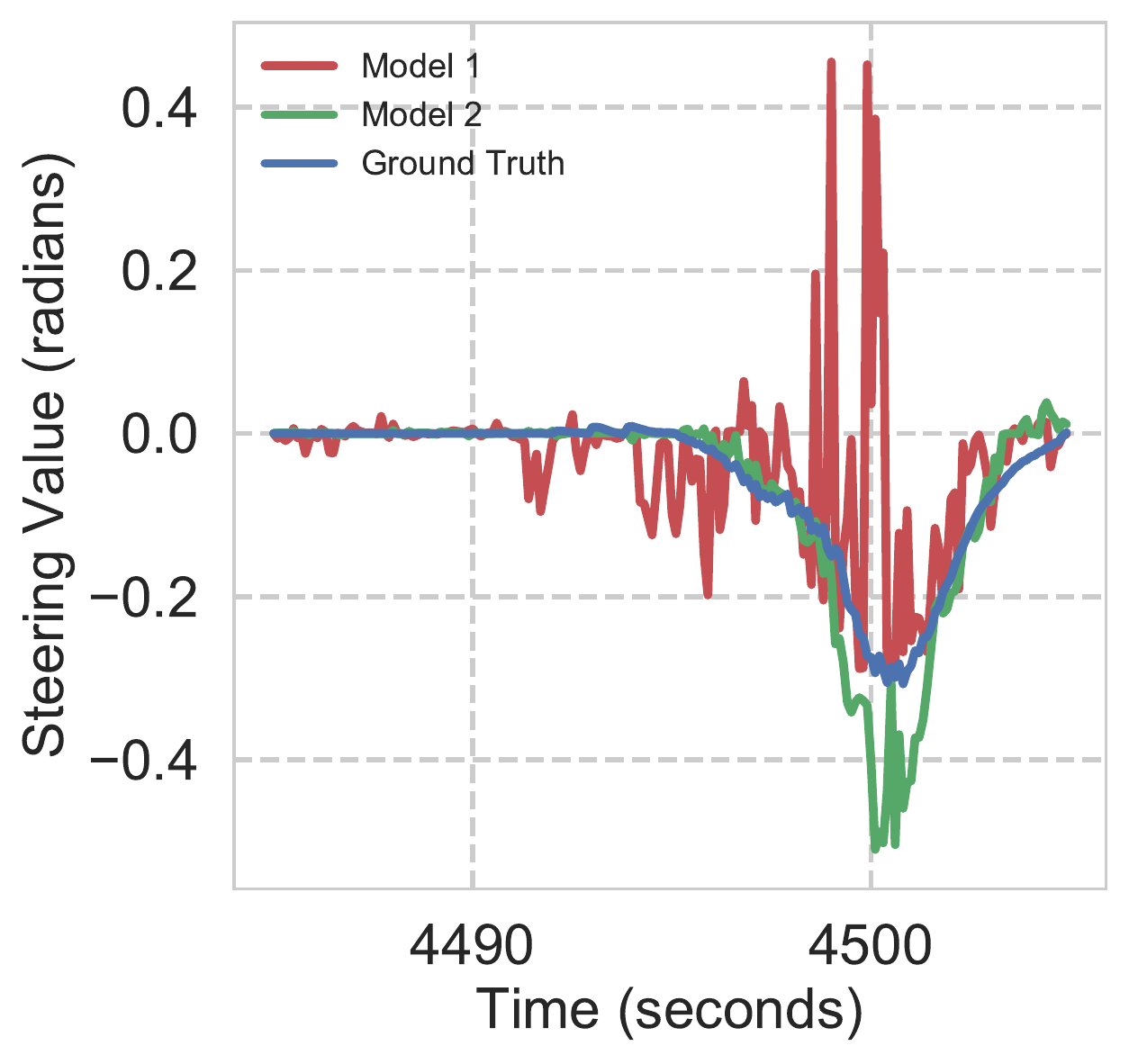} \\
    \multicolumn{2}{c}{Driving trajectories of Model 1} & \multicolumn{2}{c}{Driving trajectories of Model 2} \\
    \multicolumn{2}{c}{\includegraphics[height=0.35\linewidth,width=0.45\linewidth]{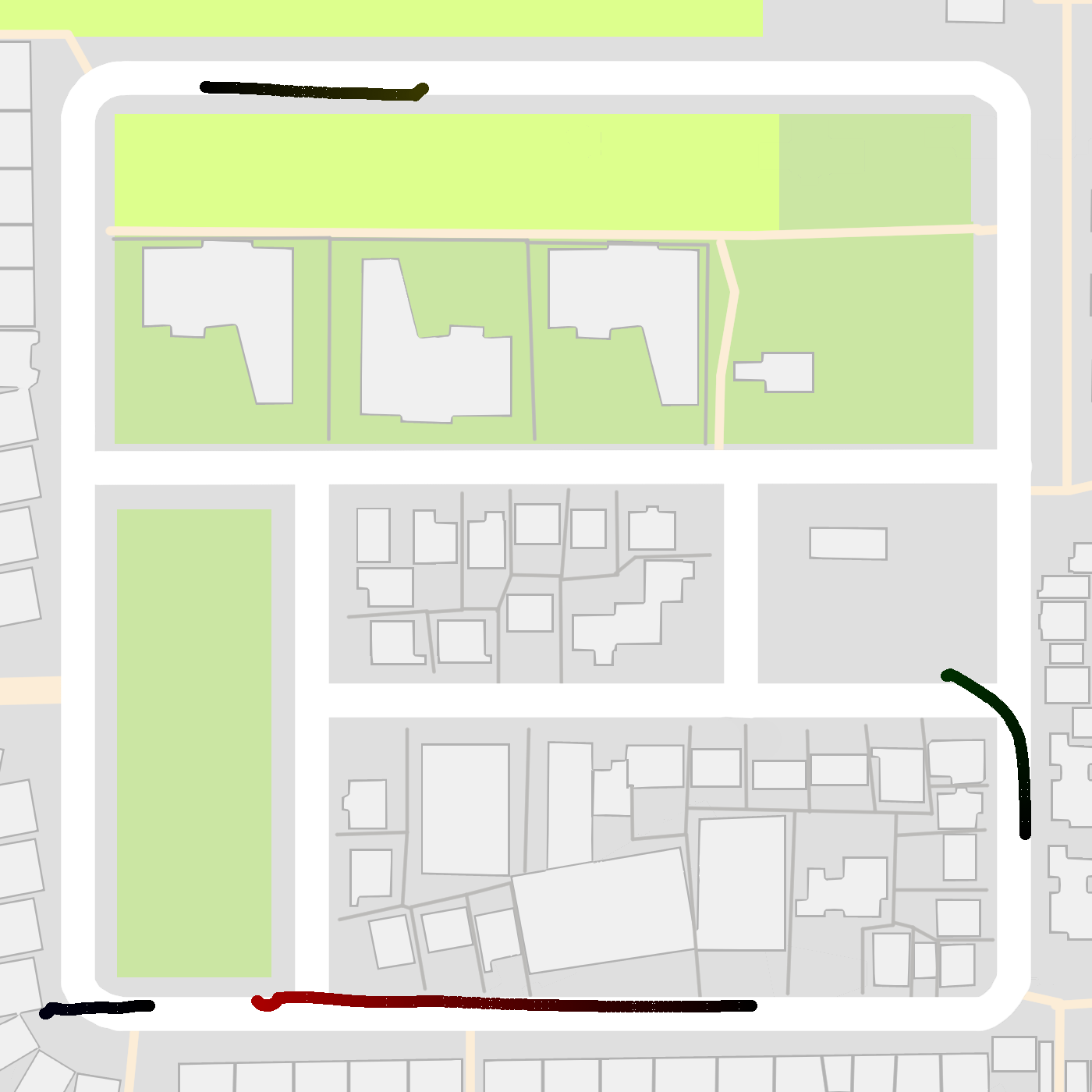}} &
    \multicolumn{2}{c}{\includegraphics[height=0.35\linewidth,width=0.45\linewidth]{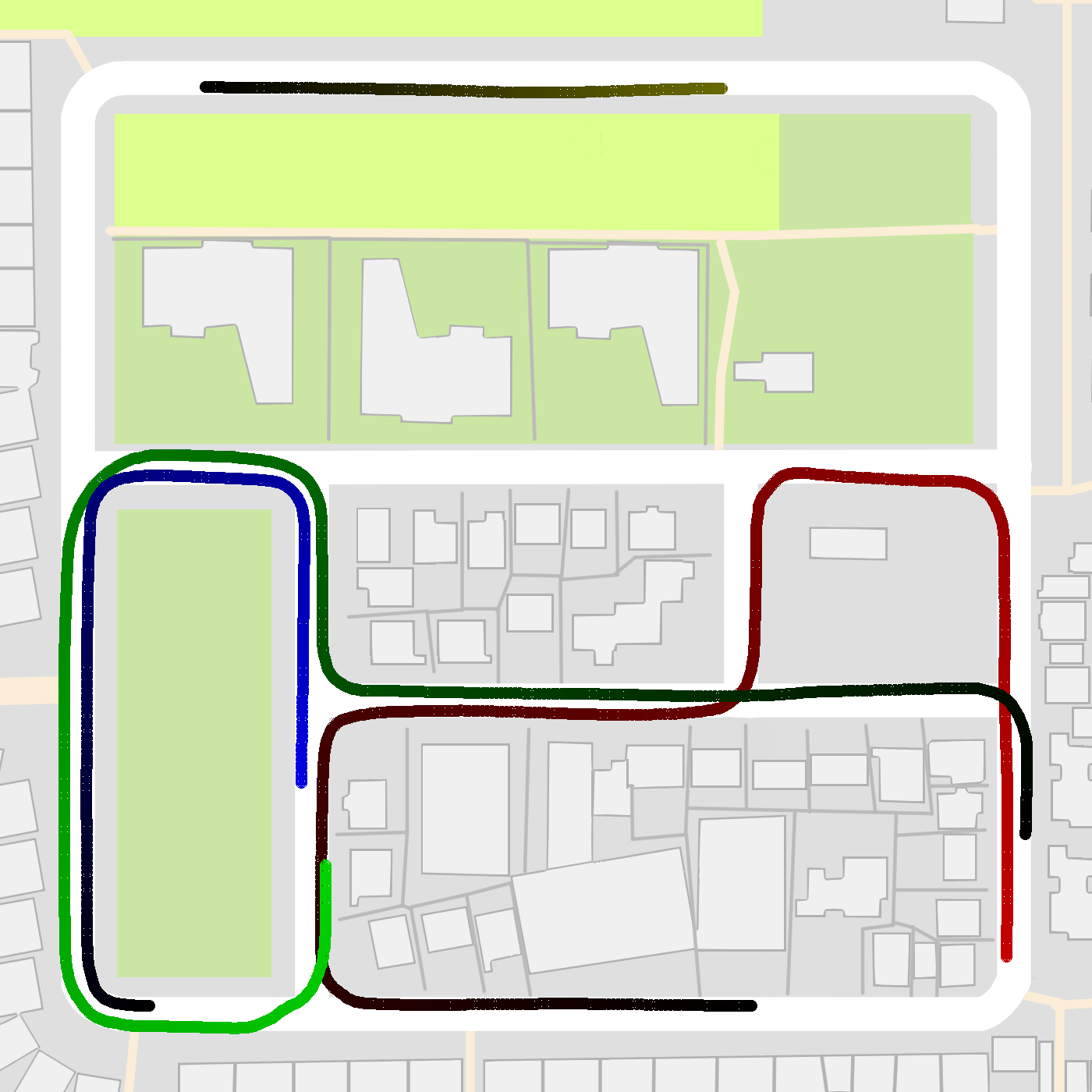}}
  \end{tabular}
  }
  \caption{Detailed evaluation of two driving models with similar offline prediction quality, but very different driving behavior. Top left: Ground-truth steering signal (blue) and predictions of two models (red and green) over time. Top right: a zoomed fragment of the steering time series, showing a large mistake made by Model 1 (red). Bottom: Several trajectories driven by the models in Town 1. Same scenarios indicated with the same color in both plots. Note how the driving performance of the models is dramatically different: Model 1 crashes in every trial, while Model 2 can drive successfully.}
  \label{fig:case_study}
\end{figure}

Figure~\ref{fig:case_study} (top left) shows the ground truth steering signal over time (blue), as well as the predictions of the models (red and green, respectively).
There is no obvious qualitative difference in the predictions of the models: both often deviate from the ground truth.
One difference is a large error in the steering signal predicted by Model 1 in a turn, as shown in Figure \ref{fig:case_study} (top right).
Such a short-term discrepancy can lead to a crash, and it is difficult to detect based on average prediction error.
The advanced offline metrics evaluated above are designed to be better at capturing such mistakes.

Figure~\ref{fig:case_study} (bottom) shows several trajectories driven by both models.
Model 1 is able to drive straight for some period of time, but eventually crashes in every single trial, typically because of wrong timing or direction of a turn.
In contrast, Model 2 drives well and successfully completes most trials.
This example illustrates the difficulty of using offline metrics for predicting online driving behavior.

\begin{table}
\centering
\caption{Detailed accuracy evaluations on the BDDV dataset. We report the $4$-way classification accuracy (in \%) for various data subsets and varying speed.}

\resizebox{0.7\linewidth}{!}{
\setlength{\tabcolsep}{4pt}
\footnotesize
 \begin{tabular}{ccccccccc}
\toprule
                     && \multicolumn{4}{c}{Average accuracy} && \multicolumn{2}{c}{Weighted with speed} \\
    Model            && All data & Straight & Stop & Turns   && All data & Turns \\
    \midrule
    Feedforward      &&  78.0    &  90.0    & 72.0 & 32.4    && 80.7 & 27.7 \\
    CNN + LSTM       &&  81.8    &  90.2    & 78.1 & 49.3    && 83.0 & 43.2 \\
    FCN + LSTM       &&  83.3    &  90.4    & 80.7 & 50.7    && 83.6 & 44.4 \\
\bottomrule
 \end{tabular}
 }
\label{tbl:bdd_models}
\end{table}

\subsection{Real-world data}
Evaluation of real-world urban driving is logistically complicated, therefore we restrict the experiments on real-world data to an offline evaluation.
We use the BDDV dataset and the trained models provided by~\cite{Xu2017}.
The models are trained to perform 4-way classification (accelerate, brake, left, right), and we measure their classification accuracy.
We evaluate on the validation set of BDDV.

The offline metrics we presented above are designed for continuous values and cannot be directly applied to classification-based models.
Yet, some of them can be adapted to this discrete setting.
Table \ref{tbl:bdd_models} shows the average accuracy, as well as several additional metrics.
First, we provide a breakdown of classification accuracy by subsets of the data corresponding to different ground truth labels.
The prediction error in the turns is most informative, yielding the largest separation between the best and the worst models.
Second, we try weighting the errors with the ground-truth speed.
We measure the resulting metric for the full validation dataset, as well as for turns only.
These metrics reduce the gap between the feedforward and the LSTM models.


\subsection{Detailed evaluation of models}

Scatter plots presented in the previous sections indicate general tendencies, but not the performance of specific models.
Here we provide a more detailed evaluation of several driving models, with a focus on several parameters: the amount of training data, its distribution, the regularization being used, the network architecture, and the loss function.
We evaluate two offline metrics~-- MSE and the thresholded relative error (TRE)~-- as well as the goal-directed navigation success rate.
For TRE we use the parameter $\alpha=0.1$.

The results are shown in Table~\ref{tbl:parameter_analysis}.
In each section of the table all parameters are fixed, except for the parameter of interest.
(Parameters may vary between sections.)
Driving performance is sensitive to all the variations.
Larger amount of training data generally leads to better driving.
Training with one or three cameras has a surprisingly minor effect.
Data balancing helps in both towns.
Regularization helps generalization to the previously unseen town and weather.
Deeper networks generally perform better.
Finally, the L1 loss leads to better driving than the usual MSE loss.
This last result is in agreement with Figure~\ref{fig:scatter_offline_metrics}, which shows that absolute error is better correlated with the driving quality than MSE.

Next, for each of the $6$ parameters and each of the $2$ towns we check if the best model chosen based on the offline metrics is also the best in terms of the driving quality.
This simulates a realistic parameter tuning scenario a practitioner might face.
We find that TRE is more predictive of the driving performance than MSE, correctly identifying the best-driving model in 10 cases out of 12, compared to 6 out of 12 for MSE.
This demonstrates that TRE, although far from being perfectly correlated with the online driving quality, is much more indicative of well-driving models than MSE.

\begin{table}
\centering
\caption{Detailed evaluation of models in CARLA. ``TRE'' stands for thresholded relative error, ``Success rate'' for the driving success rate. For MSE and TRE lower is better, for the success rate higher is better. We mark with bold the best result in each section. We highlight in green the cases where the best model according to an offline metric is also the best at driving, separately for each section and each town. Both MSE and TRE are not necessarily correlated with driving performance, but generally TRE is more predictive of driving quality, correctly identifying 10 best-driving models out of 12, compared to 6 out of 12 for MSE.}
\resizebox{0.97\linewidth}{!}{
\setlength{\tabcolsep}{3pt}
\footnotesize
 \begin{tabular}{lllccccccccc}
\toprule
                            &&                      && \multicolumn{2}{c}{MSE}     && \multicolumn{2}{c}{TRE @ 0.1} && \multicolumn{2}{c}{Success rate} \\
    Parameter               && Value                && Town 1 & Town 2 && Town 1 & Town 2 && Town 1 & Town 2 \\
    \midrule
    \textbf{Amount of training data } && 0.2 hours                                &  & 0.0086 & 0.0481 &  & 0.970 & 0.985 &  & 0.44 & 0.00 \\
                                      && 1 hour                                   &  & 0.0025 & 0.0217 &  & 0.945 & 0.972 &  & 0.44 & 0.04 \\
                                      && 5 hours                                  &  & \dred{\textbf{0.0005}} & \dgreen{\textbf{0.0093}} &  & 0.928 & 0.961 &  & 0.60 & \textbf{0.08} \\
                                      && 25 hours                                 &  & 0.0007 & 0.0166 &  & \dgreen{\textbf{0.926}} & \dred{\textbf{0.958}} &  & \textbf{0.76} & 0.04 \\
    \midrule
    \textbf{Type of training data   } && 1 cam., no noise                         &  & 0.0007 & \dred{\textbf{0.0066}} &  & \dgreen{\textbf{0.922}} & 0.947 &  & \textbf{0.84} & 0.04 \\
                                      && 1 cam., noise                            &  & 0.0009 & 0.0077 &  & 0.926 & \dgreen{\textbf{0.946}} &  & 0.80 & \textbf{0.20} \\
                                      && 3 cam., no noise                         &  & \dgreen{\textbf{0.0004}} & 0.0086 &  & 0.928 & 0.953 &  & \textbf{0.84} & 0.08 \\
                                      && 3 cam., noise                            &  & 0.0007 & 0.0166 &  & 0.926 & 0.958 &  & 0.76 & 0.04 \\
    \midrule
    \textbf{Data balancing          } && No balancing                             &  & 0.0012 & \dred{\textbf{0.0065}} &  & 0.907 & \dred{\textbf{0.924}} &  & 0.88 & 0.36 \\
                                      && With balancing                           &  & \dgreen{\textbf{0.0011}} & 0.0066 &  & \dgreen{\textbf{0.891}} & 0.930 &  & \textbf{0.92} & \textbf{0.56} \\
    \midrule
    \textbf{Regularization          } && None                                     &  & 0.0014 & 0.0092 &  & \dgreen{\textbf{0.911}} & 0.953 &  & \textbf{0.92} & 0.08 \\
                                      && Mild dropout                             &  & 0.0010 & 0.0074 &  & 0.921 & 0.953 &  & 0.84 & 0.20 \\
                                      && High dropout                             &  & \dred{\textbf{0.0007}} & 0.0166 &  & 0.926 & 0.958 &  & 0.76 & 0.04 \\
                                      && High drop., data aug.                    &  & 0.0013 & \dgreen{\textbf{0.0051}} &  & 0.919 & \dgreen{\textbf{0.931}} &  & 0.88 & \textbf{0.36} \\
    \midrule
    \textbf{Network architecture    } && Shallow                                  &  & \dred{\textbf{0.0005}} & 0.0111 &  & 0.936 & 0.963 &  & 0.68 & 0.12 \\
                                      && Standard                                 &  & 0.0007 & 0.0166 &  & \dgreen{\textbf{0.926}} & 0.958 &  & \textbf{0.76} & 0.04 \\
                                      && Deep                                     &  & 0.0011 & \dgreen{\textbf{0.0072}} &  & 0.928 & \dgreen{\textbf{0.949}} &  & \textbf{0.76} & \textbf{0.24} \\
    \midrule
    \textbf{Loss function           } && L2                                       &  & \dred{\textbf{0.0010}} & 0.0074 &  & 0.921 & 0.953 &  & 0.84 & 0.20 \\
                                      && L1                                       &  & 0.0012 & \dgreen{\textbf{0.0061}} &  & \dgreen{\textbf{0.891}} & \dgreen{\textbf{0.944}} &  & \textbf{0.96} & \textbf{0.52} \\
\bottomrule
 \end{tabular}
 }
\label{tbl:parameter_analysis}
\end{table}


\section{Conclusion}
\label{sec:conclusion}
We investigated the performance of offline versus online evaluation metrics for autonomous driving.
We have shown that the MSE prediction error of expert actions is not a good metric for evaluating the performance of autonomous driving systems, since it is very weakly correlated with actual driving quality.
We explore two avenues for improving the offline metrics: modifying the validation data and modifying the metrics themselves.
Both paths lead to improved correlation with driving quality.

Our work takes a step towards understanding the evaluation of driving models, but it has several limitations that can be addressed in future work.
First, the evaluation is almost entirely based on simulated data.
We believe that the general conclusion about weak correlation of online and offline metrics is likely to transfer to the real world; however, it is not clear if the details of our correlation analysis will hold in the real world.
Performing a similar study with physical vehicles operating in rich real-world environments would therefore be very valuable.
Second, we focus on the correlation coefficient as the measure of relation between two quantities.
Correlation coefficient estimates the connection between two variables to some degree, but a finer-grained analysis may be needed provide a more complete understanding of the dependencies between online and offline metrics.
Third, even the best offline metric we found is far from being perfectly correlated with actual driving quality.
Designing offline performance metrics that are more strongly correlated with driving performance remains an important challenge.

%

\section*{Acknowledgements}
Antonio M. L\'{o}pez and Felipe Codevilla acknowledge the Spanish project TIN2017-88709-R (Ministerio de Economia, Industria y Competitividad), the Generalitat de Ca\-ta\-lu\-nya CERCA Program and its ACCIO agency. Felipe Codevilla was supported in part by FI grant 2017FI-B1-00162.

{\small
\bibliographystyle{splncs04}
\bibliography{paper}
}


\end{document}


\title{Supplementary material for ``On Offline Evaluation of Vision-based Driving Models''}

%
\author{Felipe Codevilla\inst{1} \and
Antonio M. L\'{o}pez\inst{1} \and
Vladlen Koltun\inst{2} \and
Alexey Dosovitskiy\inst{2}}
%
%

\institute{Computer Vision Center, Universitat Aut\`{o}noma de Barcelona \and
Intel Labs}
\maketitle

\section{Training Details}

\subsection{Network architecture}

Table \ref{tbl:net_arch} provides the details of the standard architecture used in the experiments.
We also experimented with a deeper architecture with $12$ convolutional layers (an additional layer at each feature map resolution) and a shallower network with $4$ convolutional layers (one layer at each resolution instead of two).

		\begin{table}[]
		  \centering
		    \begin{tabular}{|c|cccc|}
		      \hline
		      module                       & input dimension           & channels            & kernel & stride \\ \midrule
		      \multirow{10}{*}{Perception}  & $200 \times 88 \times 3$   & $32$                & $5$    & $2$    \\
										  & $98 \times 48 \times 32$      & $32$                & $3$    & $1$    \\
		                                   & $96 \times 46 \times 32$  & $64$                & $3$    & $2$    \\
										  & $47 \times 22 \times 64$  & $64$                & $3$    & $1$    \\
		                                   & $45 \times 20 \times 64$  & $128$                & $3$    & $2$    \\
										  & $22 \times 9 \times 128$  & $128$                & $3$    & $1$    \\
		                                   & $20 \times 7 \times 128$   & $256$               & $3$    & $2$   \\
		                                   & $9 \times 3 \times 256$   & $256$               & $3$    & $1$   \\
		                                   & $7 \cdot 1 \cdot 256$   & $512$               & $-$    & $-$   \\

		                                   & $512$   & $512$               & $-$    & $-$

		\\ \hline
		      \multirow{3}{*}{Measurement} & $1$                       & $128$               & $-$    & $-$    \\
		                                   & $128$                     & $128$               & $-$    & $-$    \\
		                                   & $128$                     & $128$               & $-$    & $-$    \\  \hline
		      \multirow{1}{*}{Joint input} & $512 + 128$               & $512$       & $-$    & $-$    \\ \hline
		      \multirow{3}{*}{Control}     & $512$                     & $256$       & $-$    & $-$    \\
											                 & $256$                     & $256$         & $-$    & $-$      \\
		                                   & $256$                     & $1$         & $-$    & $-$      \\ \hline
		    \end{tabular}
		    \vspace{0.5cm}
		  \caption{Architecture of the ``standard'' architecture used in the experiments.}
		  \label{tbl:net_arch}
		\end{table}

\subsection{Image Input}
 We record images from the simulator at the resolution of $800 \times 600$ pixels.
 Before feeding the image to the network, we cropped $171$ pixels at the top and $45$ pixels at the bottom, resized the resulting $800 \times 384$ pixels image to $200 \timess 88$ pixels resolution, subtracted the mean, and normalized.

\subsection{Augmentation}
We used the following transformations for data augmentation: Gaussian blur,
additive Gaussian noise, pixel dropout, additive and multiplicative brightness variation, contrast variation, saturation variation and also region pixel dropout.

\subsection{Noise Distribution}
During training data collection, $10\%$ of the time we inject noise into the expert's steering.
Namely, at a random point in time we add a perturbation to the steering angle provided by the driver.
The perturbation is a triangular impulse, please see Codevilla et al.~\cite{Codevilla2018} for details.

\section{Evaluated Model Details}
Table \ref{tbl:parameters} lists the model parameters we varied in our experiments.


\begin{table}[]
\small
\centering
\resizebox{0.95\linewidth}{!}{
 \begin{tabular}{ll}
\toprule
    Training data & \\
    \quad Amount & The amount of training data (in hours of driving footage)\\
    \quad Distribution & Distribution of the data used for training \\
    \quad \quad 3 cam. + noise & Data from three cameras, with action noise in 10\% of data \\
    \quad \quad 3 cam.  & Data from three cameras, without action noise \\
    \quad \quad 1 cam. + noise  & Data only from the central camera, with action noise \\
    \quad \quad 1 cam.   & Data only from the central camera, without action noise \\ \midrule
    Model architecture & \\
    \quad Architecture & The architecture of the network\\
    \quad \quad Shallow & 4-layer convolutional network  \\
    \quad \quad Standard  & 8-layer convolutional network  \\
    \quad \quad Deep     & 12-layer convolutional network  \\ \midrule
    Training procedure& \\
    \quad Regularization  & Regularization techniques applied during training \\
    \quad \quad None  & No regularization \\
    \quad \quad Mild Dropout  &  Dropout:  50\% in the FC layers of the measurements and control modules  \\
    \quad \quad High Dropout  & Dropout:  50\% in all FC layers \\ 
    \quad \quad Drop. + aug.  & Dropout and random transformations applied to input images \\
    \quad Balancing & If data balancing w.r.t. steering angles is applied during training \\
    \quad Loss  & Type of loss function used for training \\
    \quad \quad MSE (L2)  & Regression with MSE loss \\
		\quad \quad L1  & Regression with absolute error (L1) \\
\bottomrule
 \end{tabular}
}
\vspace{2mm}
\caption{Parameters of driving models explored in the evaluation.}
\label{tbl:parameters}
\end{table}

\section{Additional results}
In the main paper we evaluate the models in the generalization condition (Town 2) and we plot 50\% best-performing models according to the offline metric.
Here we show results in the training condition (Town 1) and show plots with all models, not only best-performing ones.

\begin{figure}
  \centering
  { \fontsize{7pt}{9pt}\selectfont
  \begin{tabular}{ccc}
		\multicolumn{3}{c}{\normalsize Town 1 (training conditions), best 50\% of the models.}\\
    \quad Central camera, no noise & \quad Central camera, with noise & \quad Three cameras, no noise \\
    \includegraphics[height=0.32\linewidth]{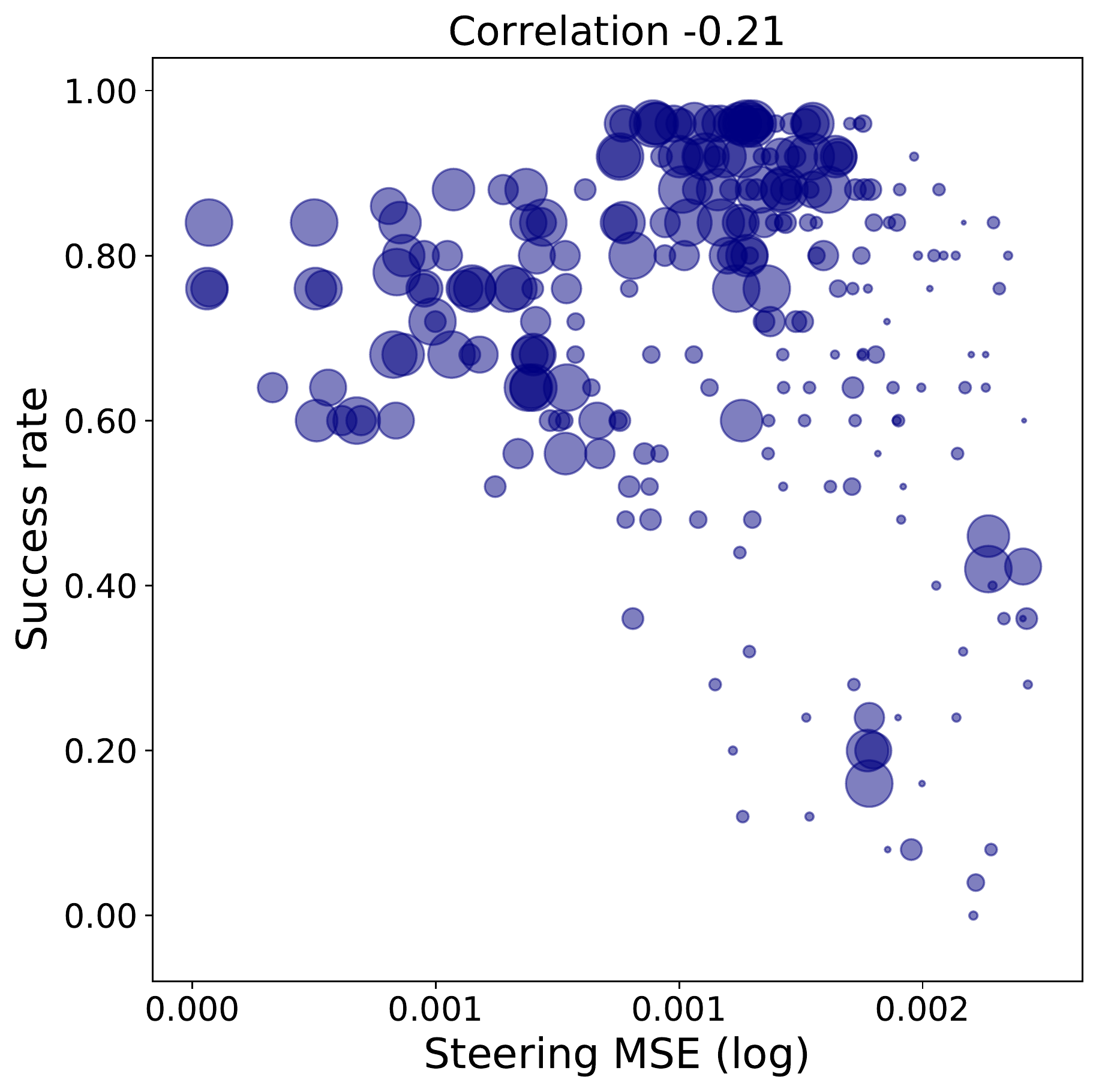} &
    \includegraphics[height=0.32\linewidth]{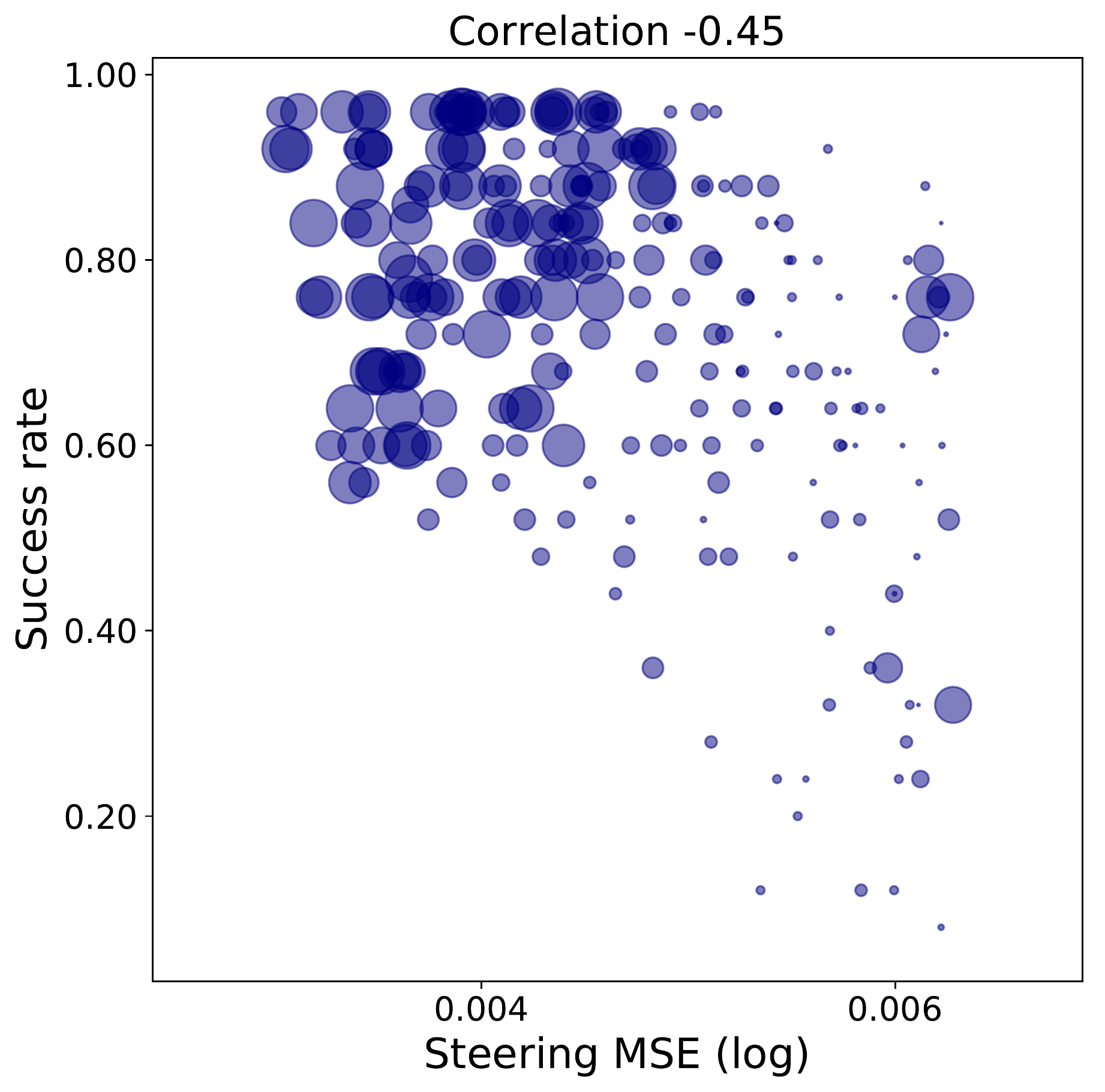} &
    \includegraphics[height=0.32\linewidth]{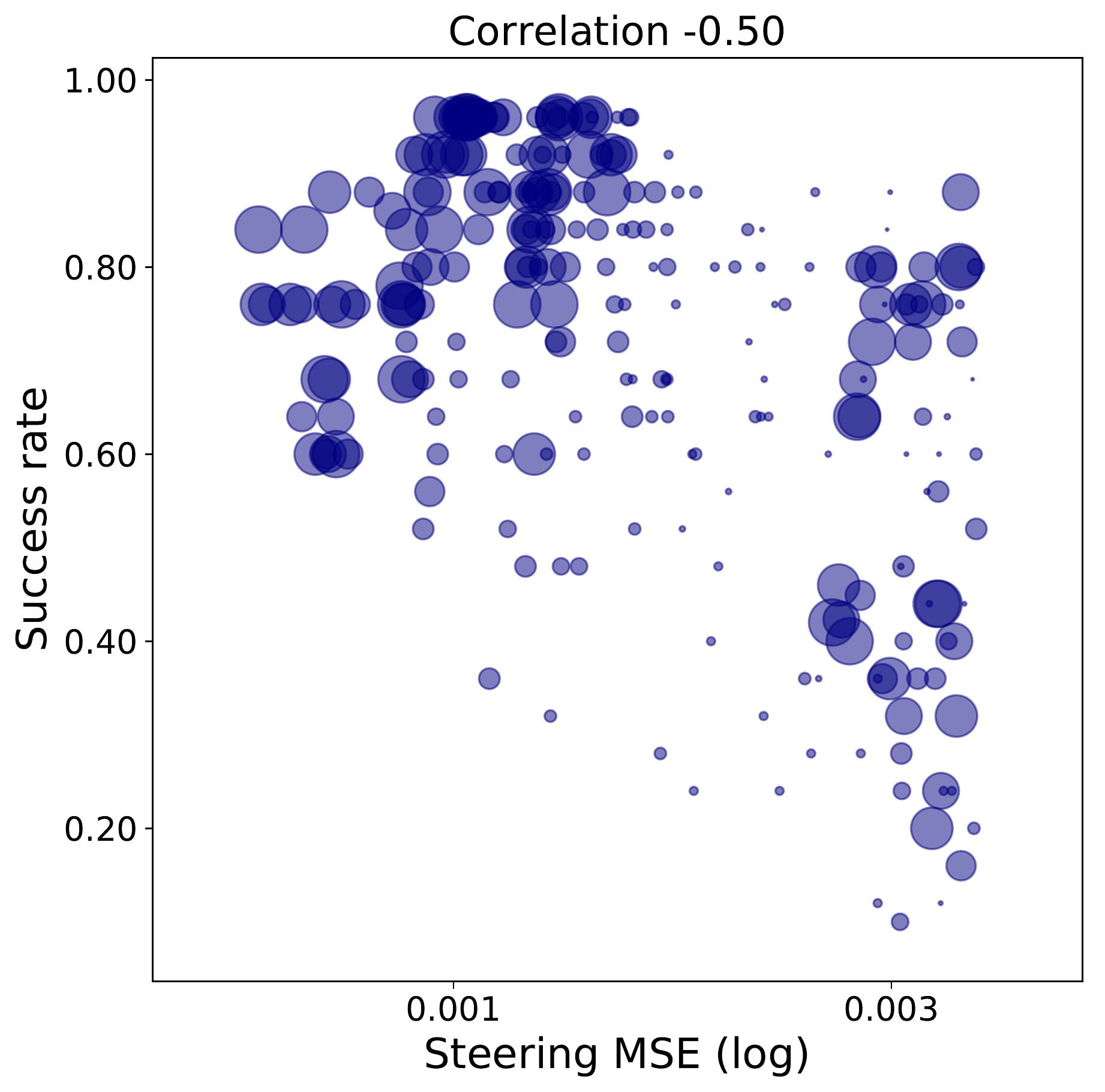}
  \end{tabular}
  }
  \caption{Scatter plots of goal-directed navigation success rate vs steering absolute error when evaluated on data from different distributions. Town 1 (training conditions), best 50\% of the models.}
  \label{fig:scatter_steer_vary_data_town1best}
\end{figure}

\begin{figure}
  \centering
  { \fontsize{7pt}{9pt}\selectfont
  \begin{tabular}{ccc}
		\multicolumn{3}{c}{\normalsize Town 1 (training conditions), best 50\% of the models.}\\
    \quad Steering MSE & \quad Steering absolute error & \quad Speed-weighted error \\
    \includegraphics[height=0.32\linewidth]{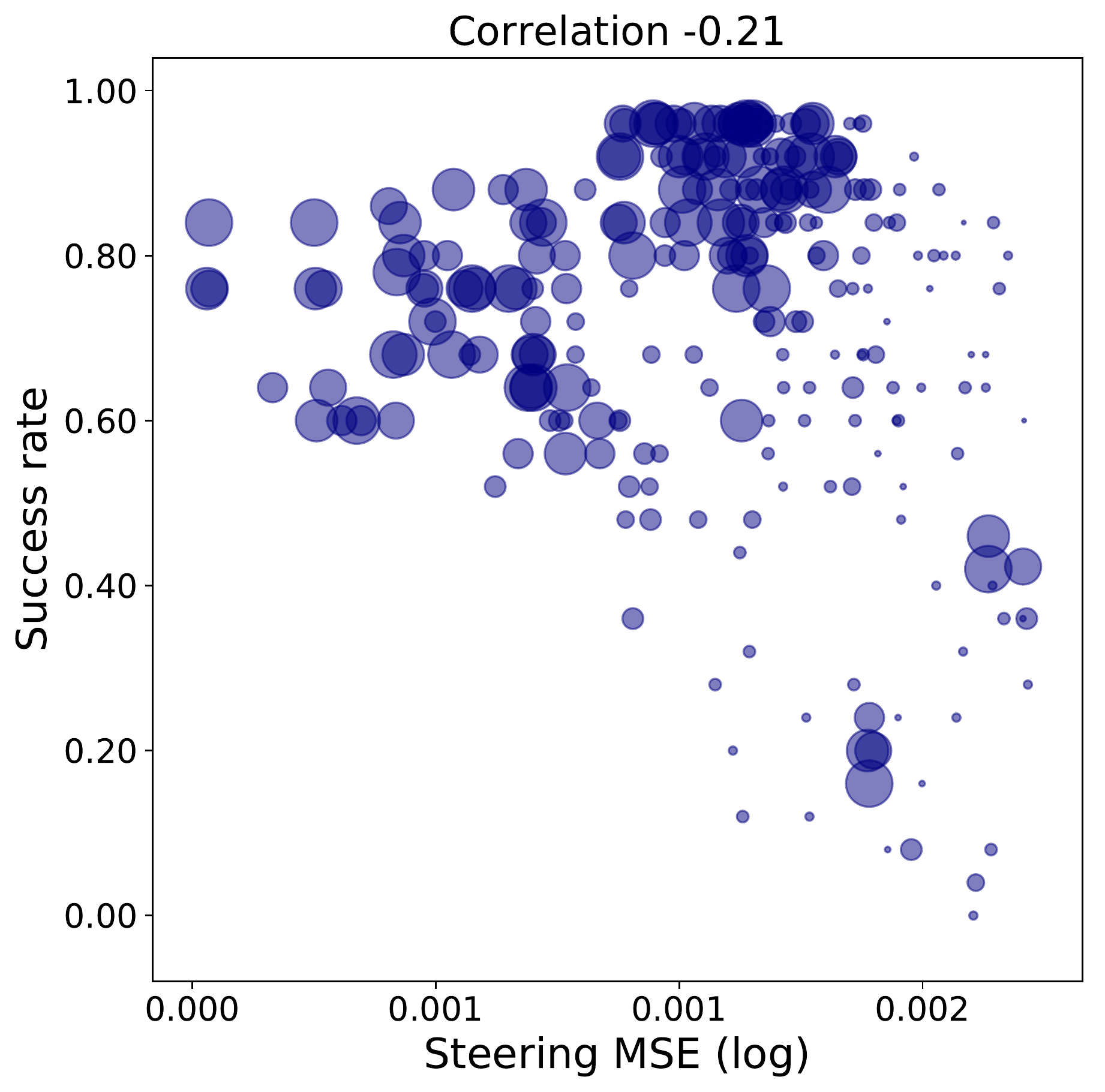} &
    \includegraphics[height=0.32\linewidth]{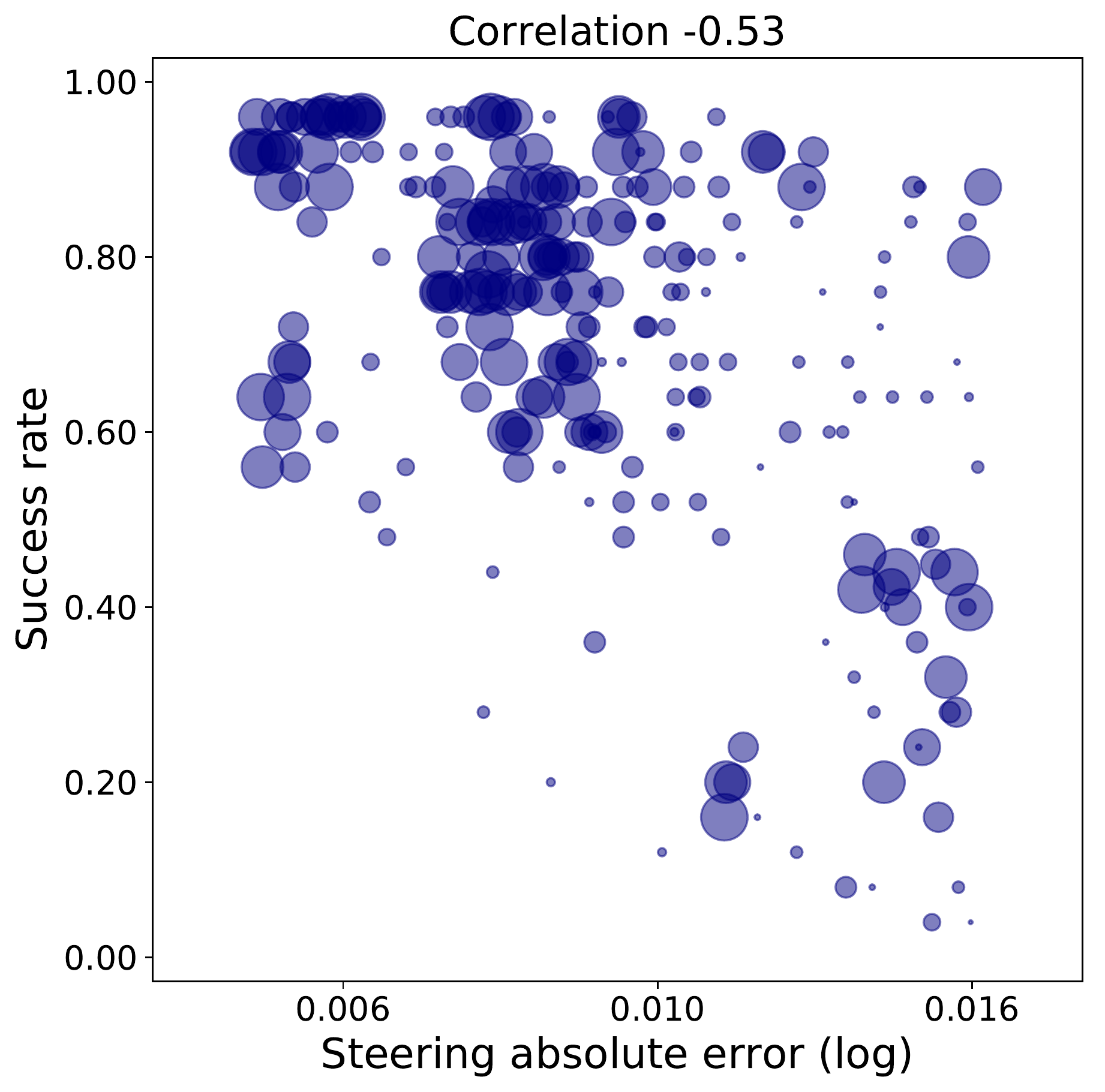} &
    \includegraphics[height=0.32\linewidth]{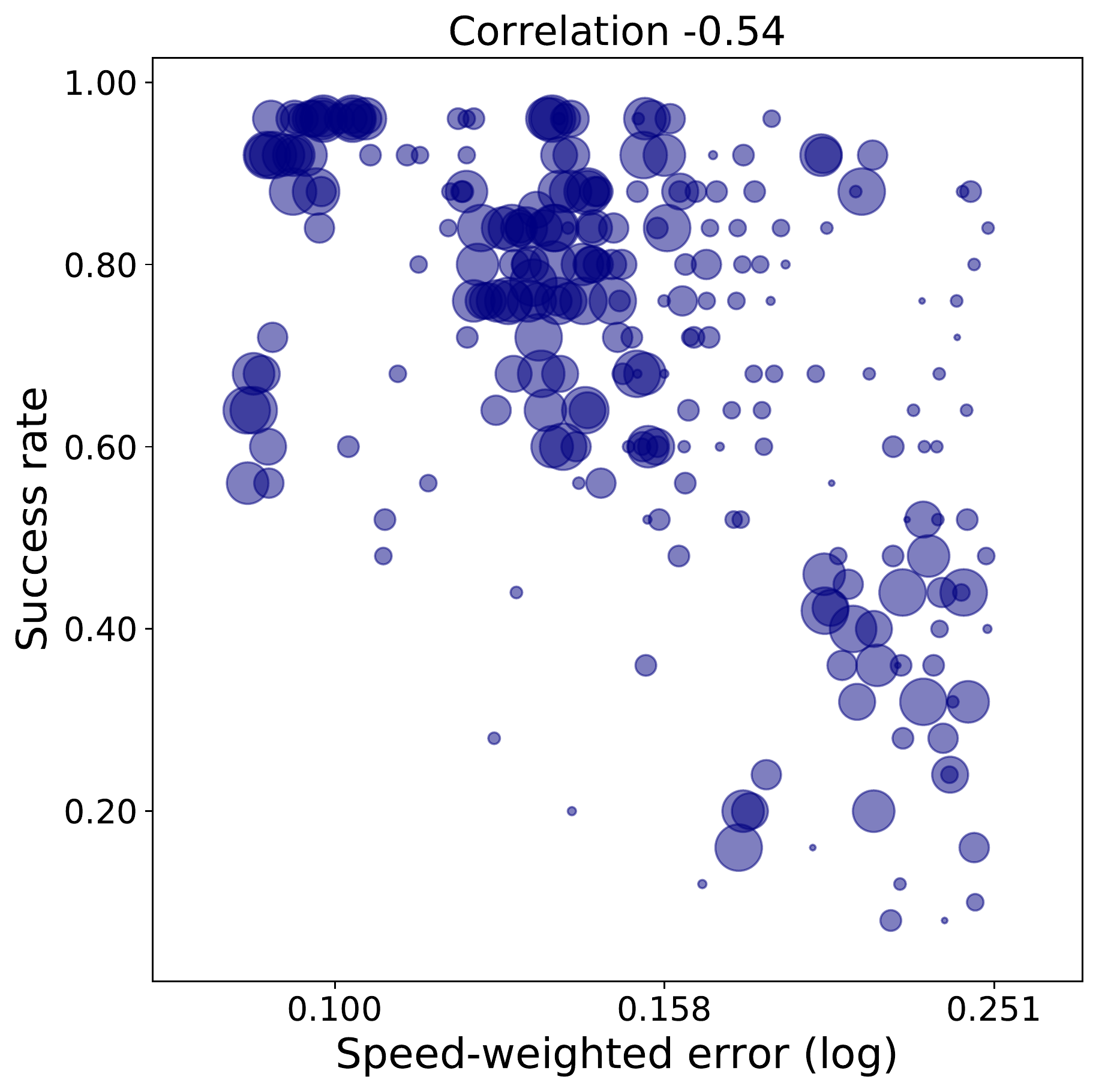} \\
    \quad Cumulative error & \quad Quantized classification & \quad Thresholded relative error \\
    \includegraphics[height=0.32\linewidth]{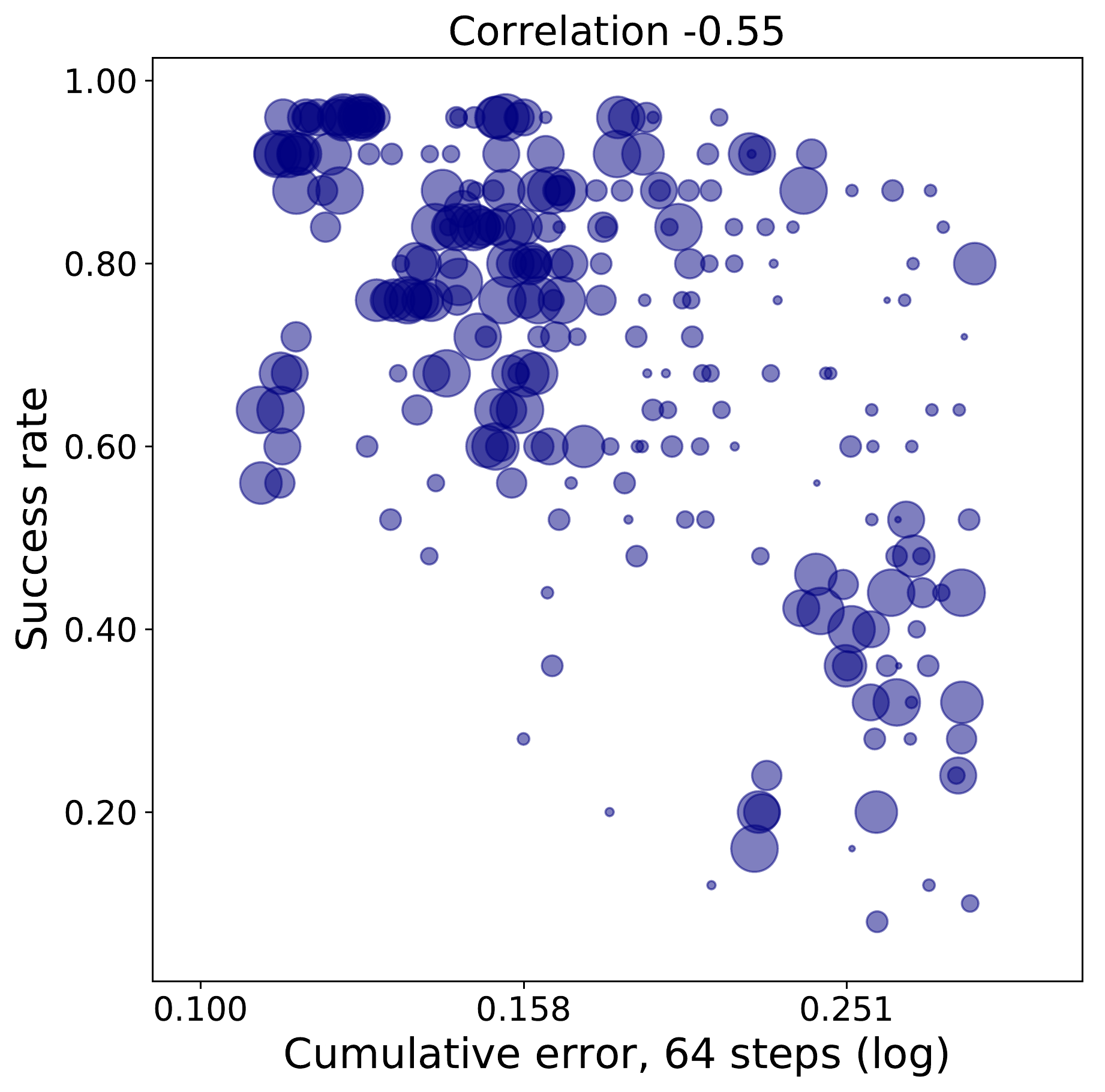} &
    \includegraphics[height=0.32\linewidth]{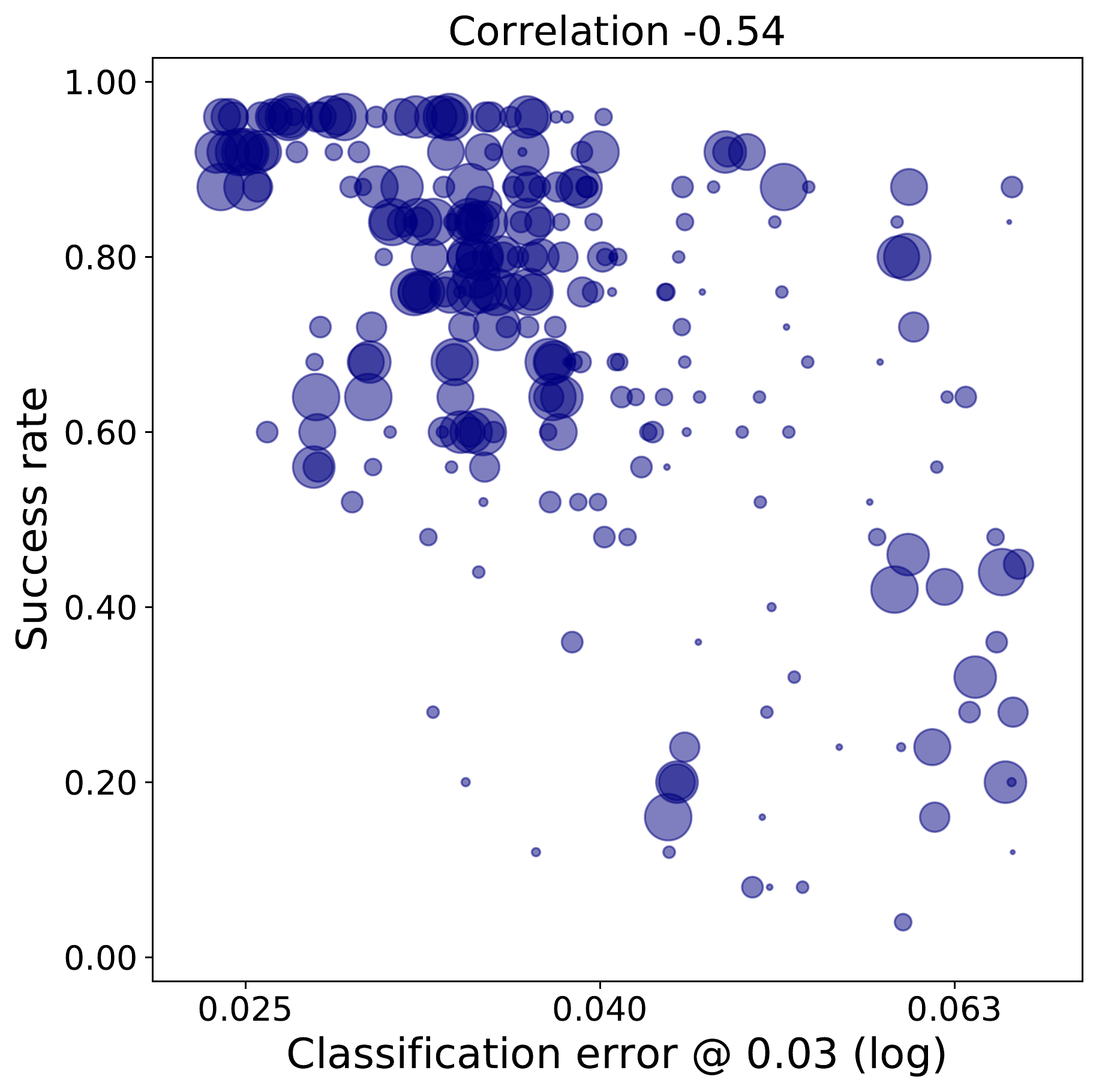} &
    \includegraphics[height=0.32\linewidth]{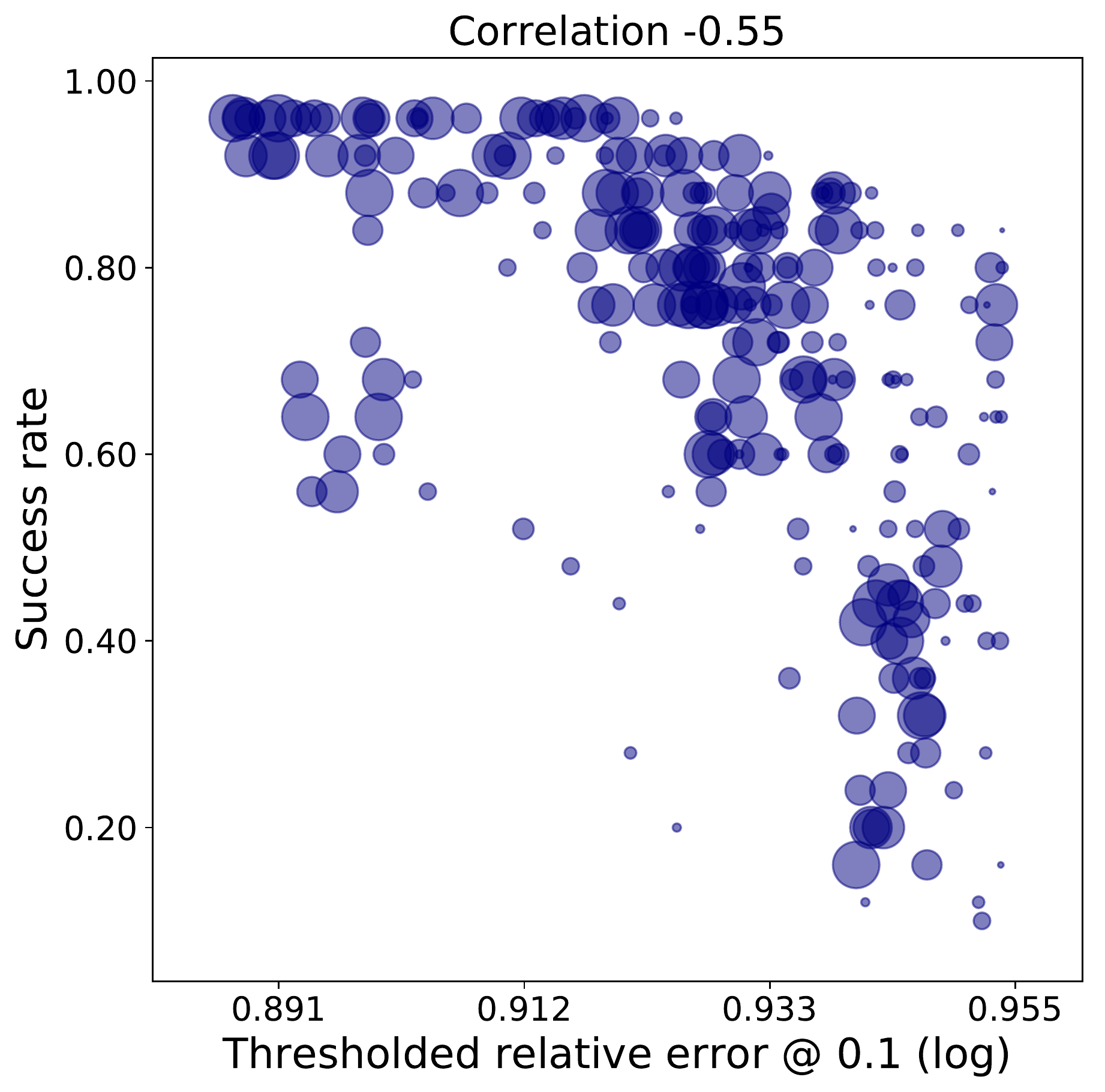}
  \end{tabular}
  }
  \caption{Scatter plots of goal-directed navigation success rate vs different offline metrics. Town 1 (training conditions), best 50\% of the models.}
  \label{fig:scatter_offline_metrics_town1best}
\end{figure}

\begin{figure}
  { \fontsize{7pt}{9pt}\selectfont
  \centering
  \begin{tabular}{ccc}
		\multicolumn{3}{c}{\normalsize Town 1 (training conditions), all models.}\\
    \quad Success rate vs Avg. completion & \quad Km per infraction vs Success rate & \quad Km per infraction vs Avg. completion \\
    \includegraphics[height=0.32\linewidth]{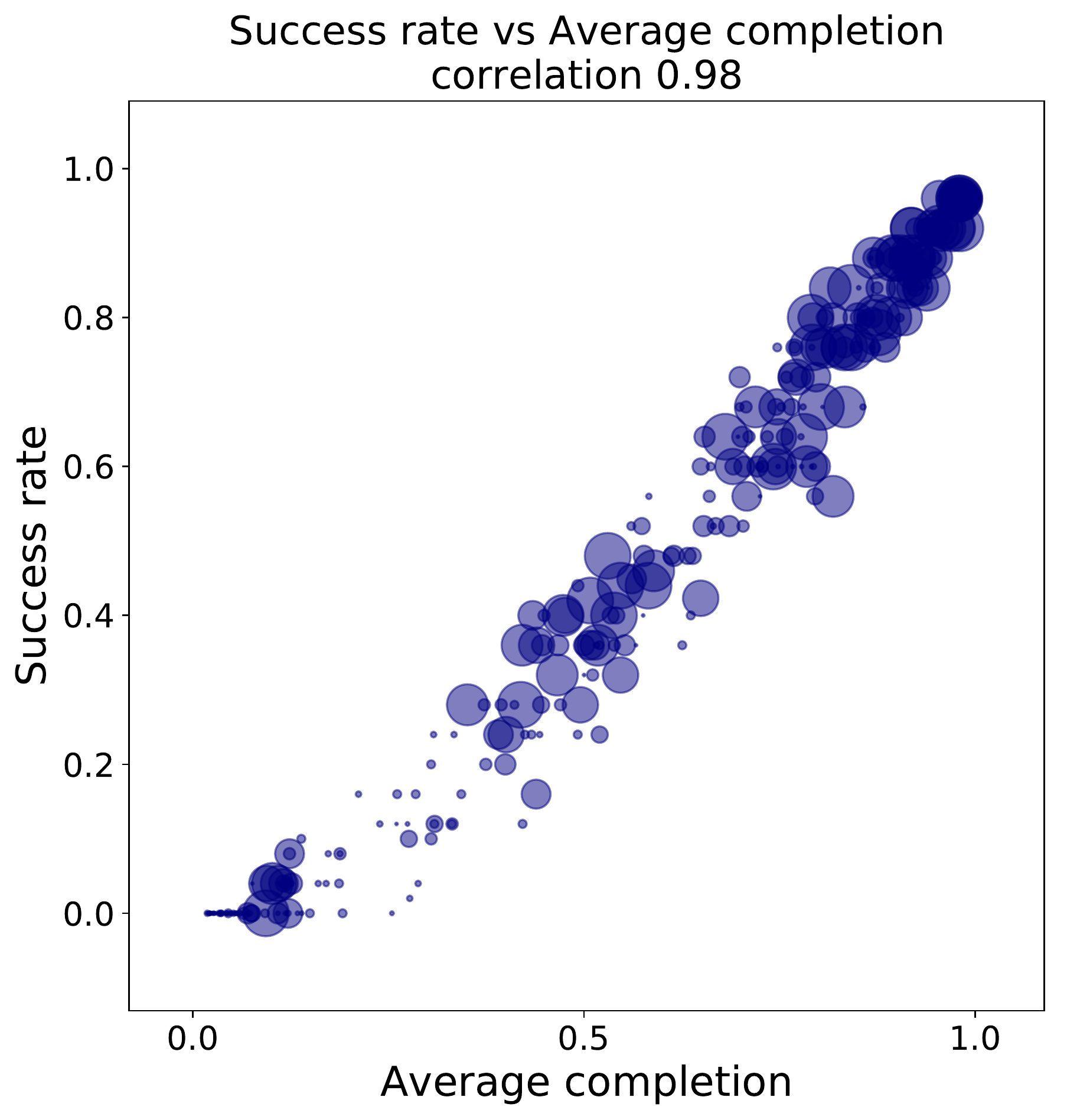} &
    \includegraphics[height=0.32\linewidth]{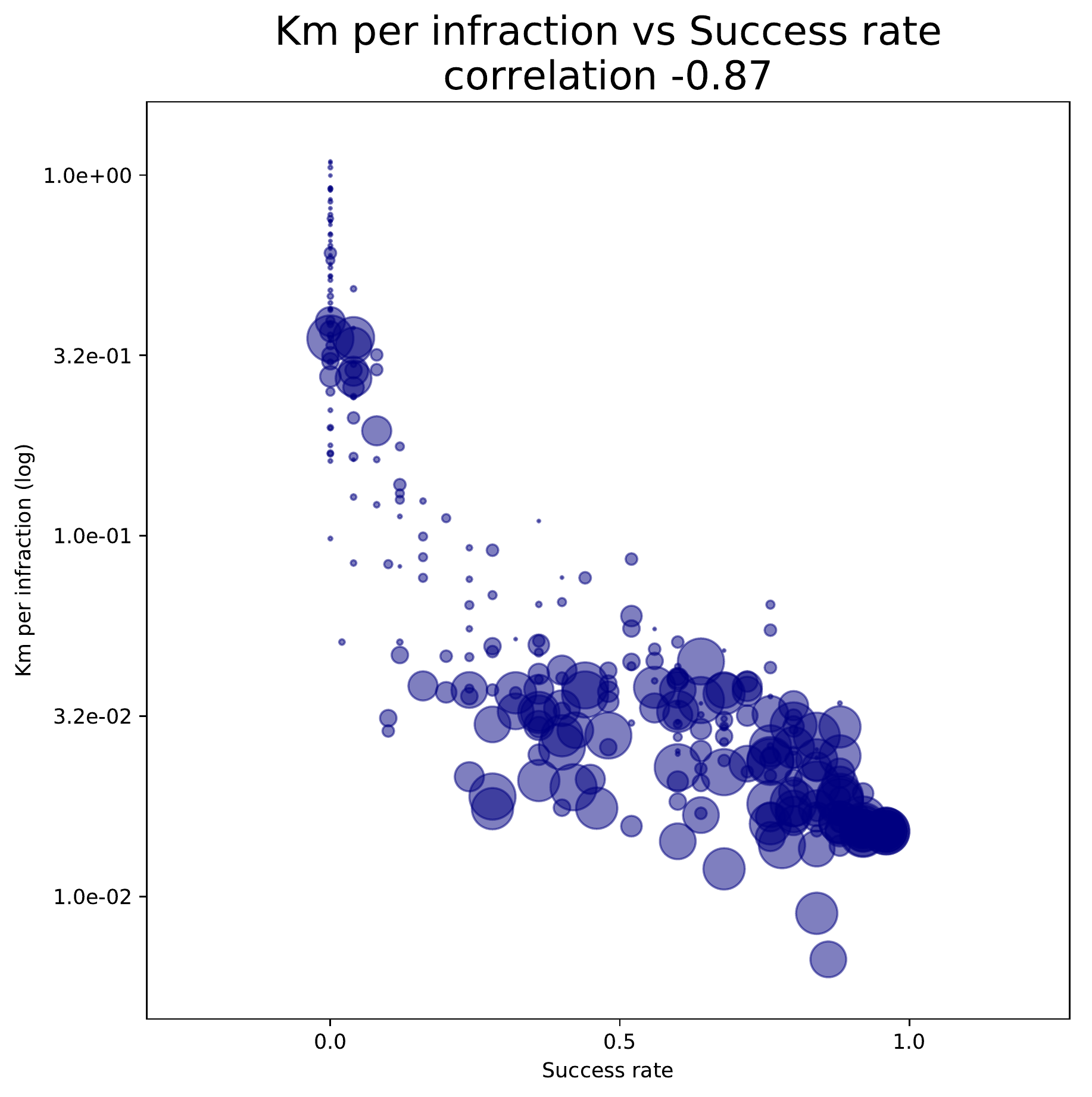} &
    \includegraphics[height=0.32\linewidth]{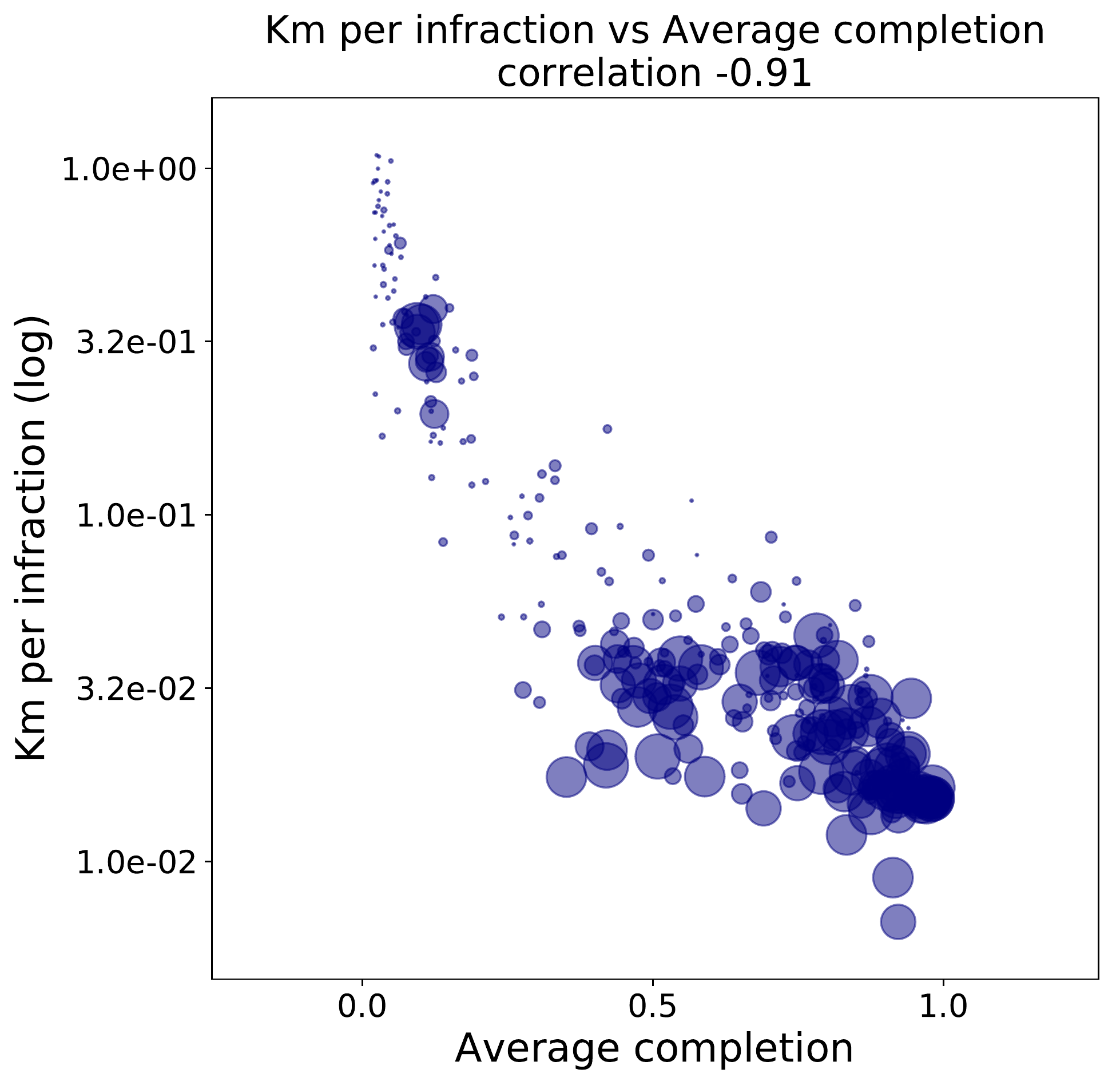}
  \end{tabular}
  }
  \caption{Scatter plots of online driving quality metrics versus each other. The metrics are: success rate, average fraction of distance to the goal covered (average completion), and average distance (in km) driven between two infractions. Town 1 (training conditions), all models.}
  \label{fig:scatter_online_metrics_town1all}
\end{figure}

\begin{figure}
  \centering
  { \fontsize{7pt}{9pt}\selectfont
  \begin{tabular}{ccc}
		\multicolumn{3}{c}{\normalsize Town 1 (training conditions), all models.}\\
    \quad Central camera, no noise & \quad Central camera, with noise & \quad Three cameras, no noise \\
    \includegraphics[height=0.32\linewidth]{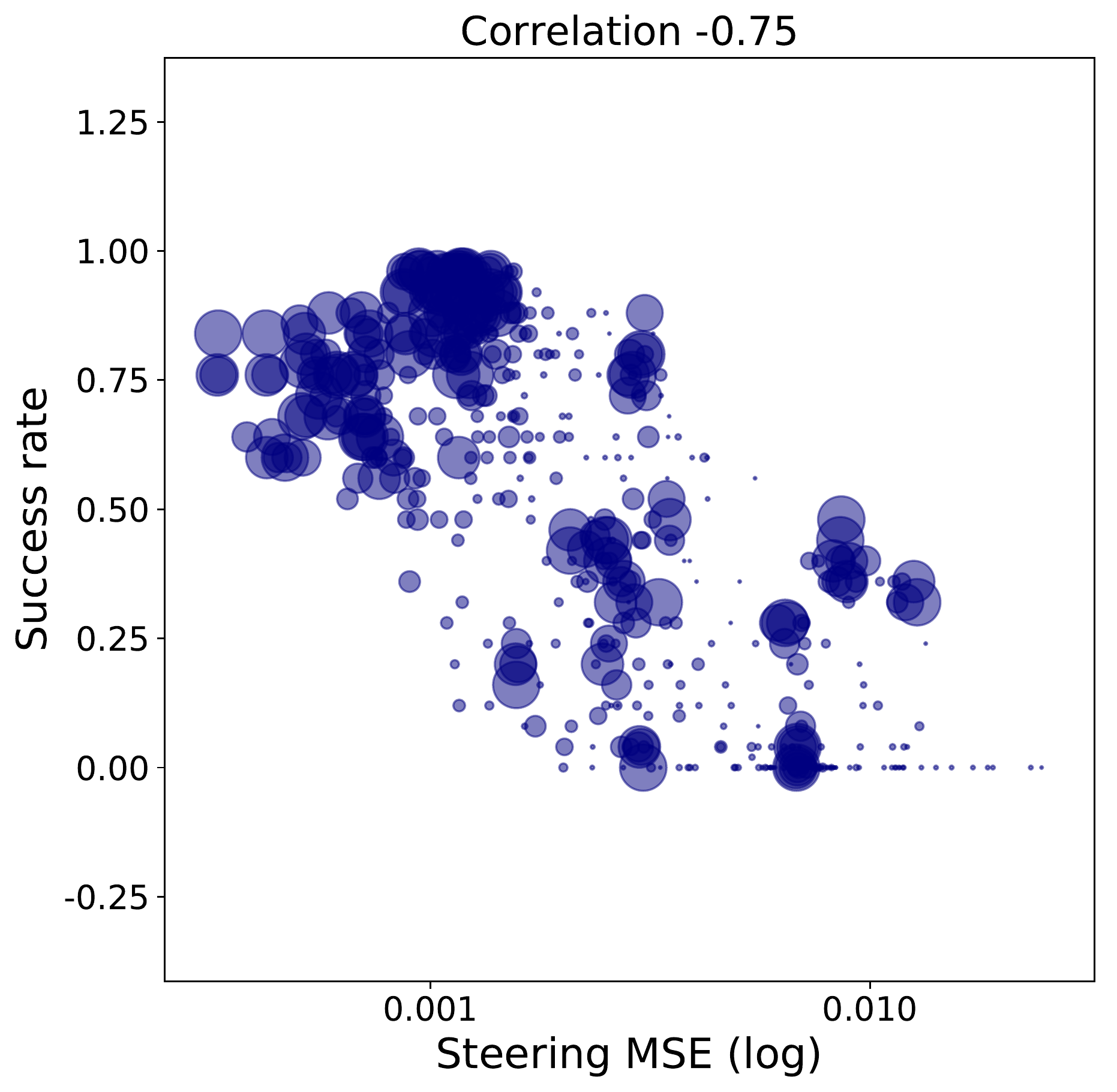} &
    \includegraphics[height=0.32\linewidth]{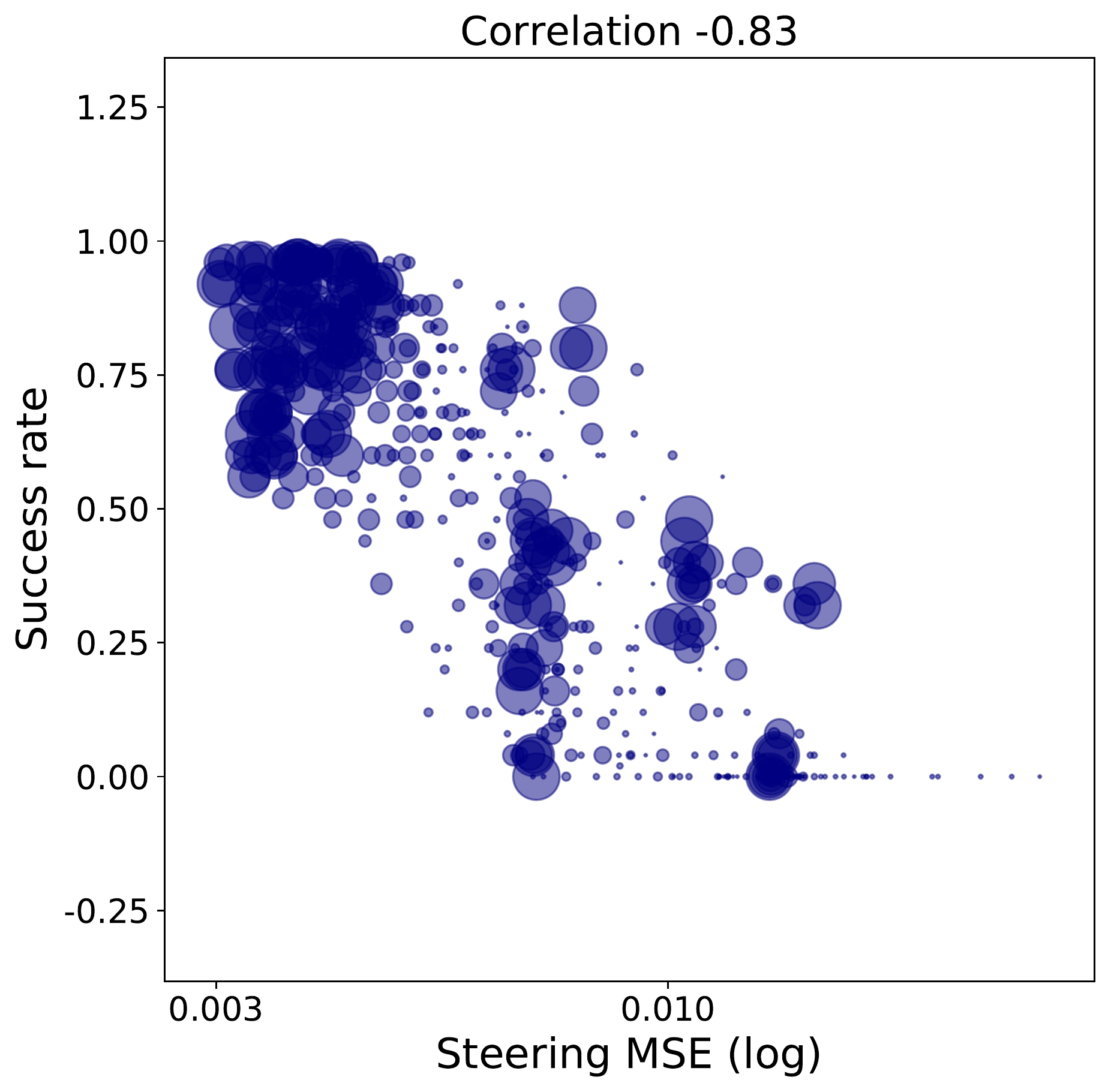} &
    \includegraphics[height=0.32\linewidth]{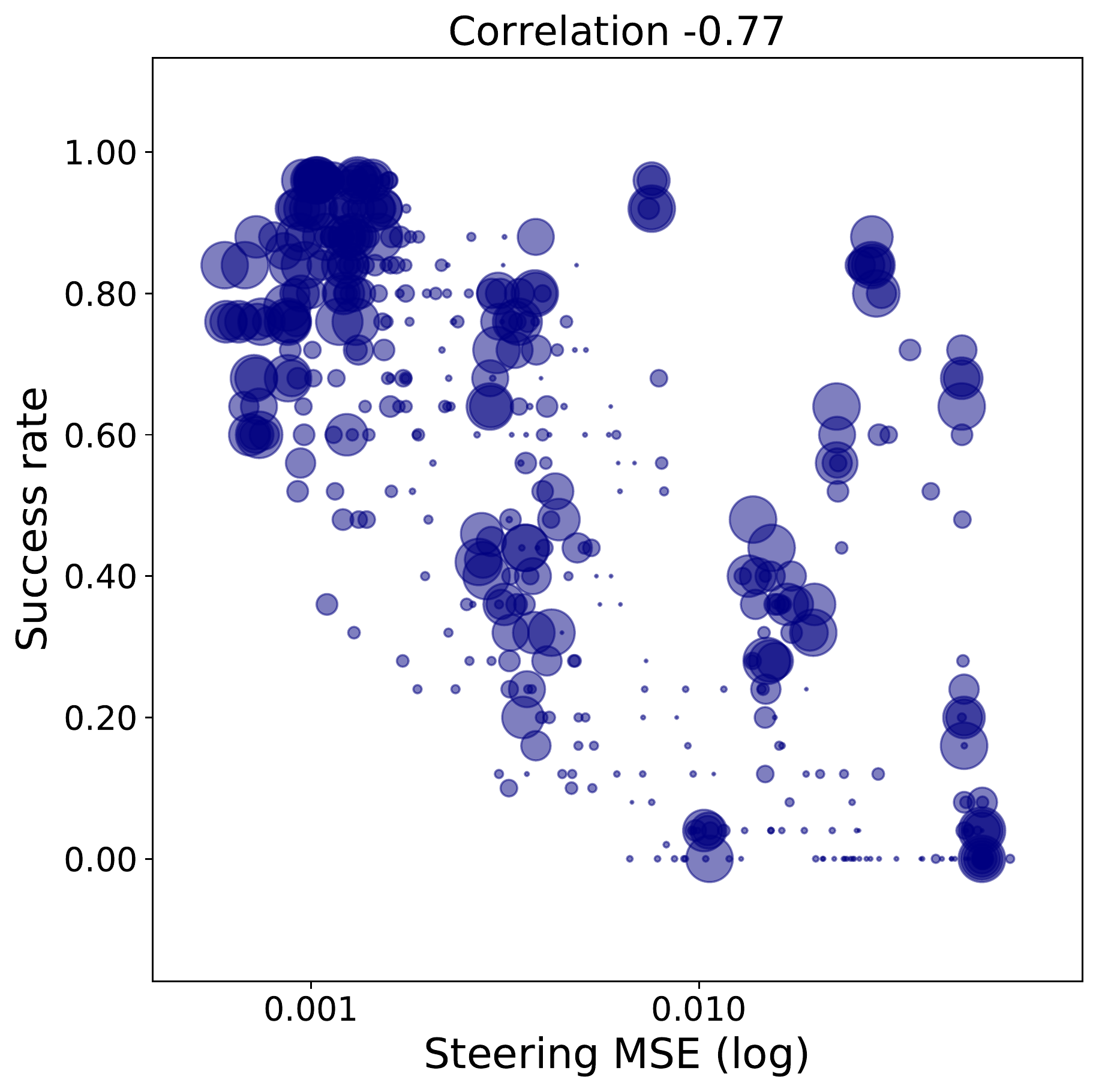}
  \end{tabular}
  }
  \caption{Scatter plots of goal-directed navigation success rate vs steering absolute error when evaluated on data from different distributions. Town 1 (training conditions), all models.}
  \label{fig:scatter_steer_vary_data_town1all}
\end{figure}

\begin{figure}
  \centering
  { \fontsize{7pt}{9pt}\selectfont
  \begin{tabular}{ccc}
		\multicolumn{3}{c}{\normalsize Town 1 (training conditions), all models.}\\
    \quad Steering MSE & \quad Steering absolute error & \quad Speed-weighted error \\
    \includegraphics[height=0.32\linewidth]{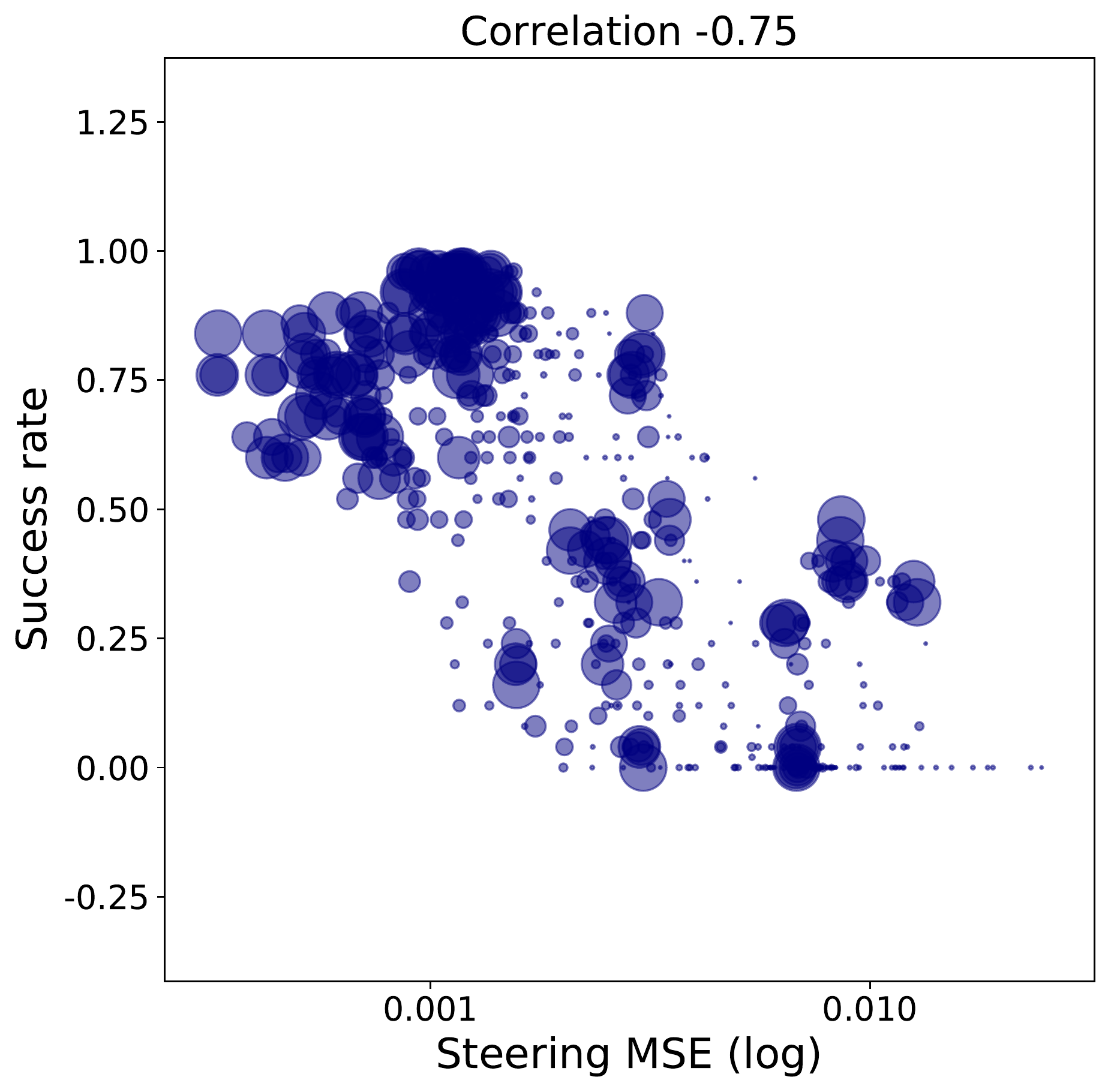} &
    \includegraphics[height=0.32\linewidth]{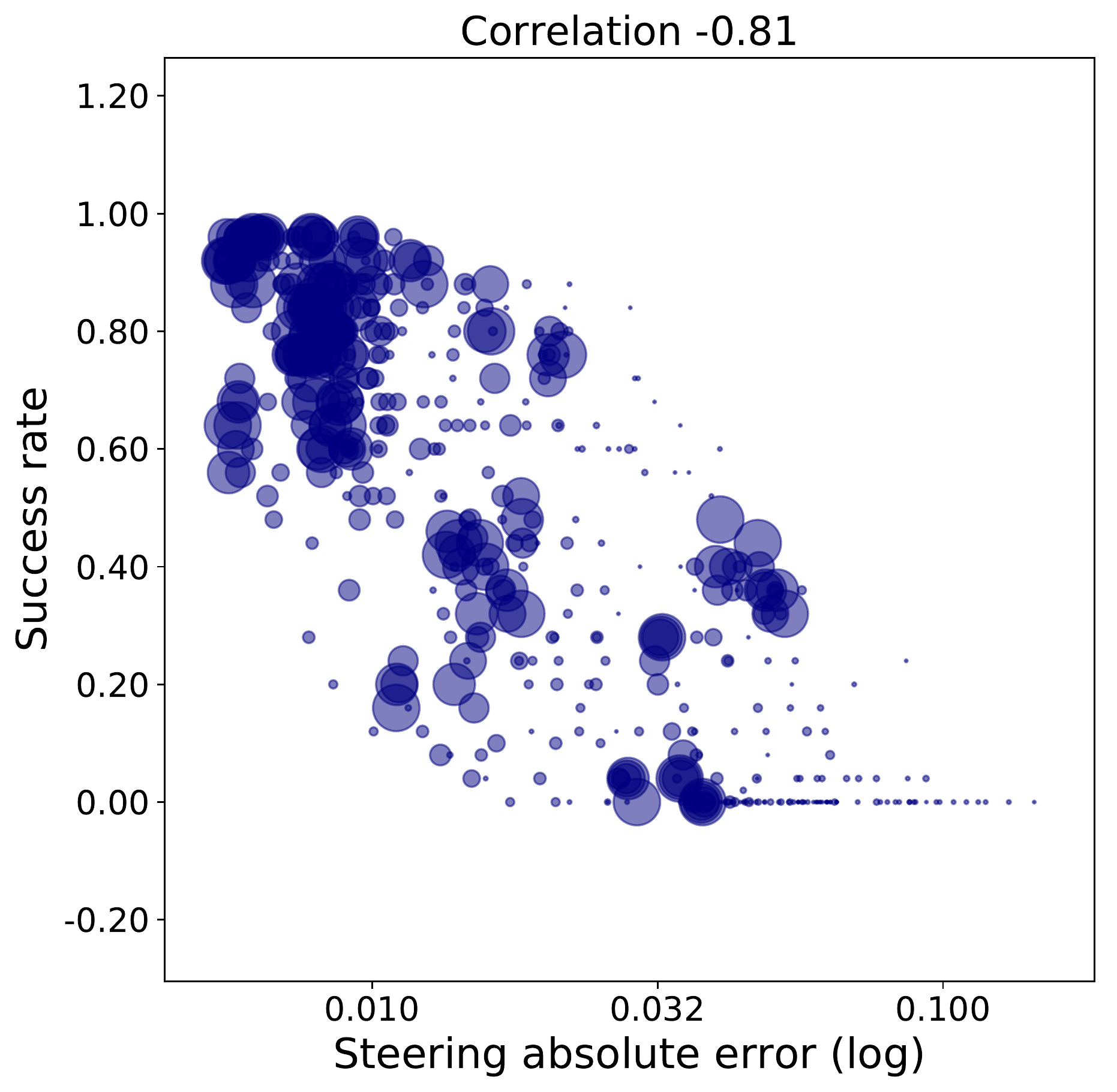} &
    \includegraphics[height=0.32\linewidth]{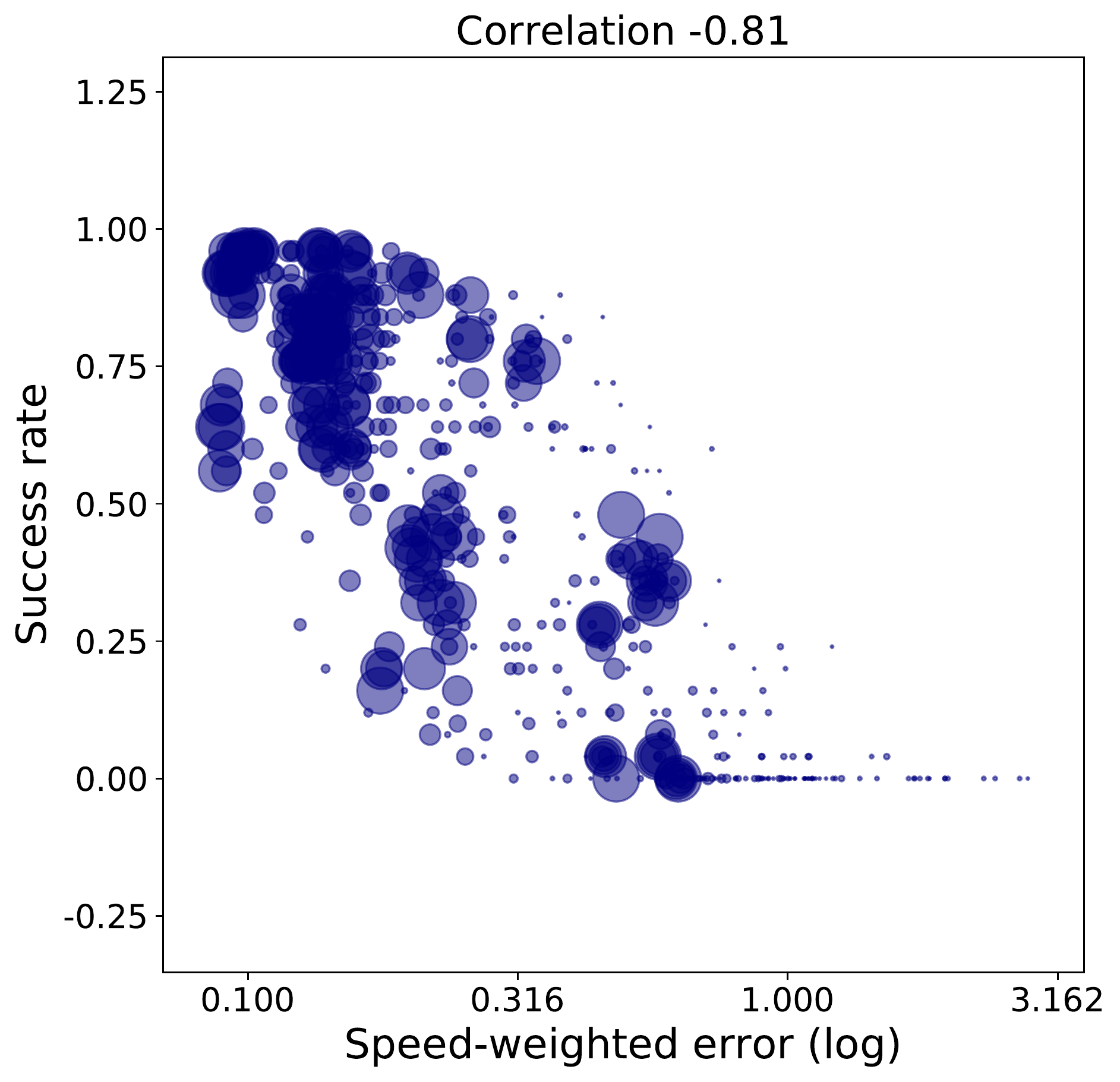} \\
    \quad Cumulative error & \quad Quantized classification & \quad Thresholded relative error \\
    \includegraphics[height=0.32\linewidth]{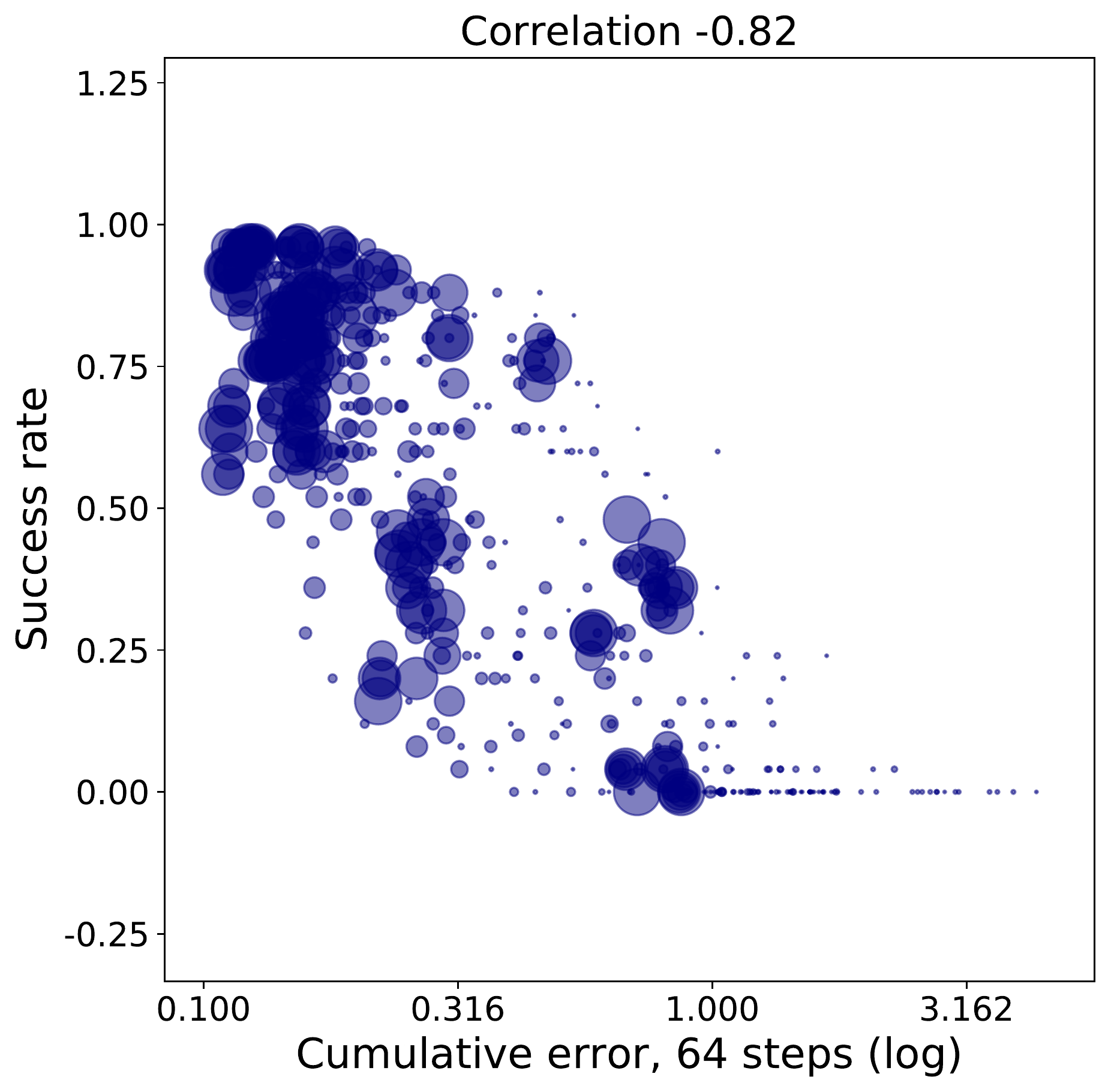} &
    \includegraphics[height=0.32\linewidth]{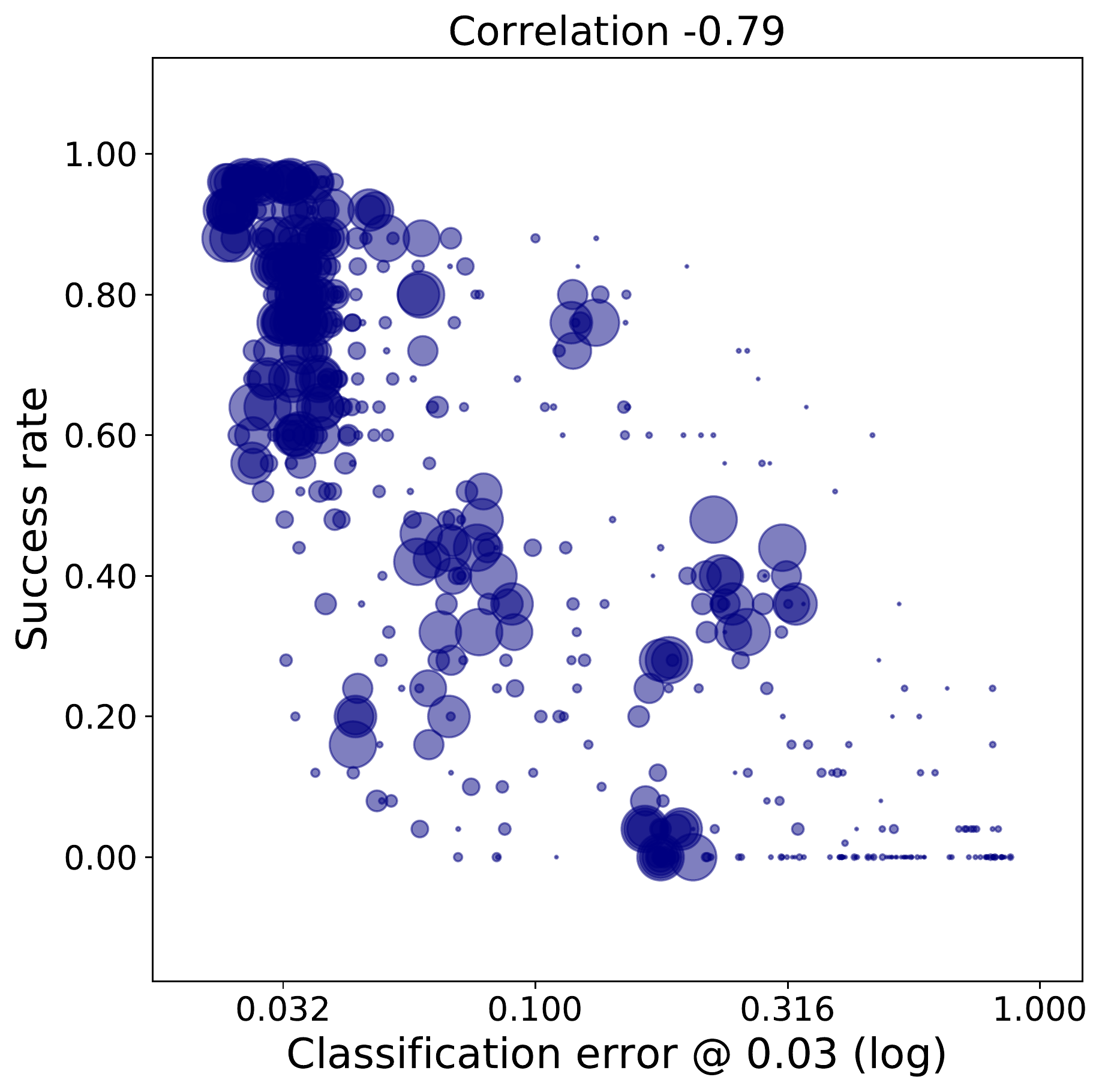} &
    \includegraphics[height=0.32\linewidth]{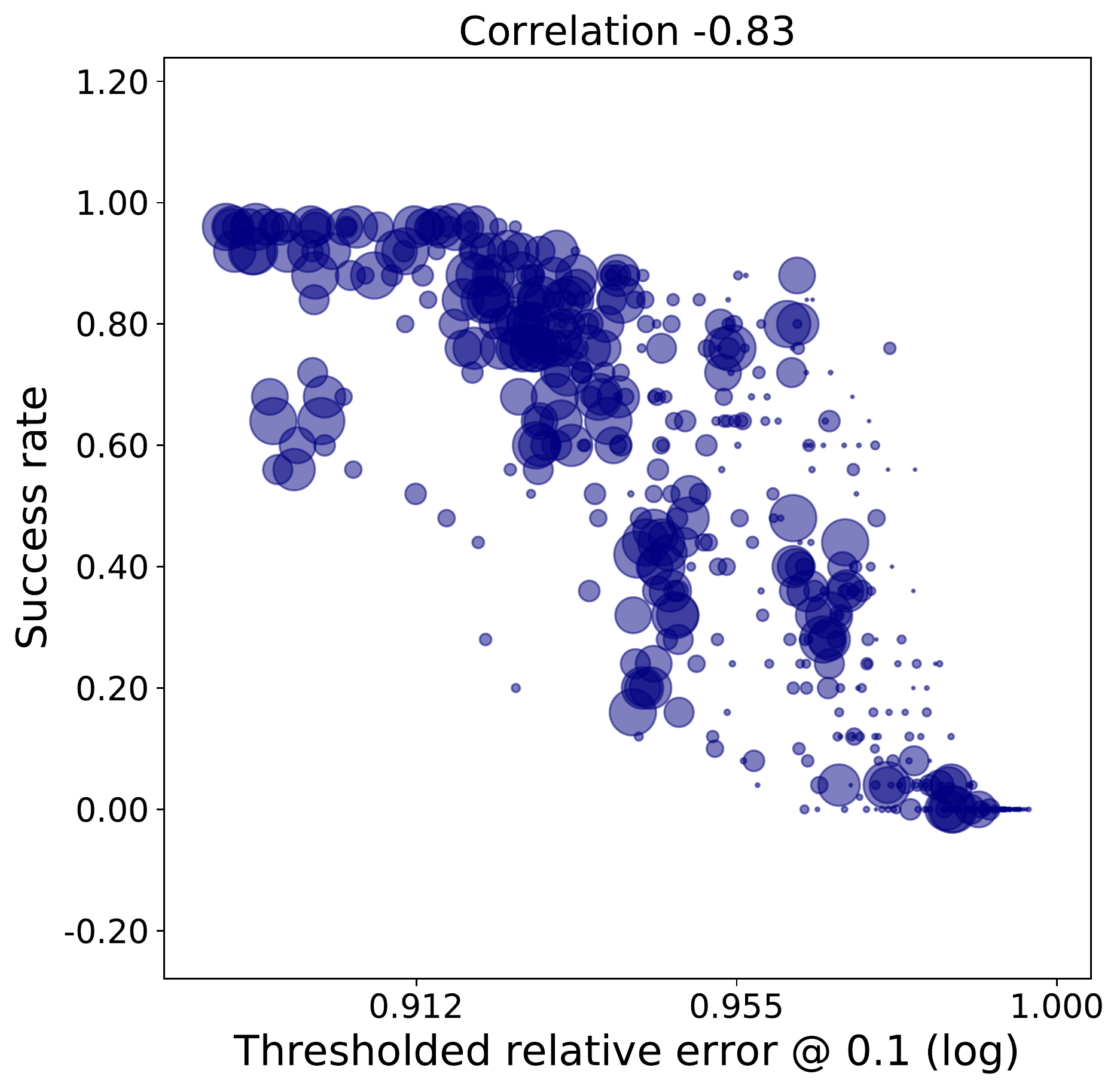}
  \end{tabular}
  }
  \caption{Scatter plots of goal-directed navigation success rate vs different offline metrics. Town 1 (training conditions), all models.}
  \label{fig:scatter_offline_metrics_town1all}
\end{figure}

\begin{figure}
  \centering
  { \fontsize{7pt}{9pt}\selectfont
  \begin{tabular}{ccc}
    \multicolumn{3}{c}{\normalsize Town 2 (generalization conditions), all models.}\\
    \quad Central camera, no noise & \quad Central camera, with noise & \quad Three cameras, no noise \\
    \includegraphics[height=0.32\linewidth]{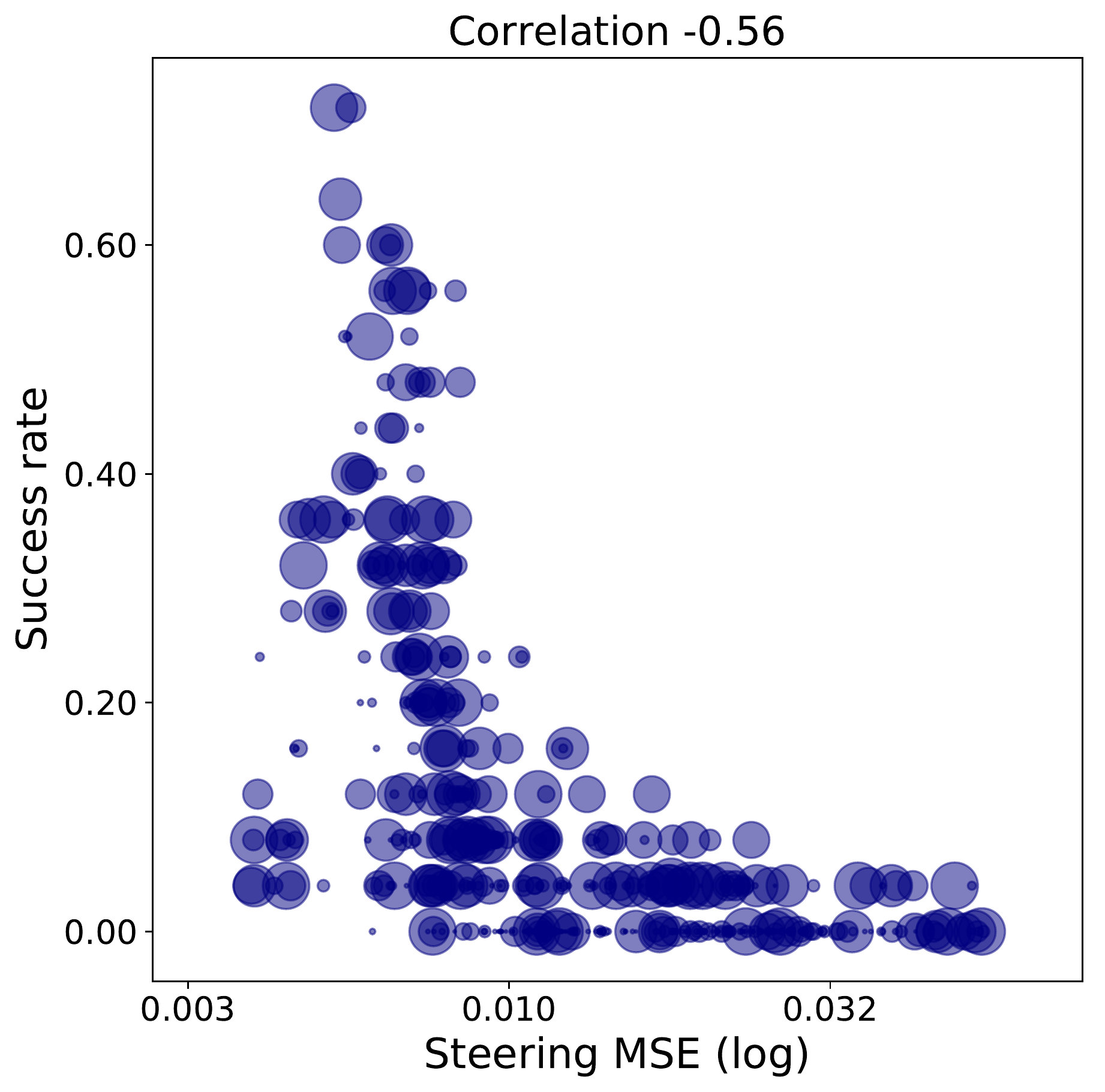} &
    \includegraphics[height=0.32\linewidth]{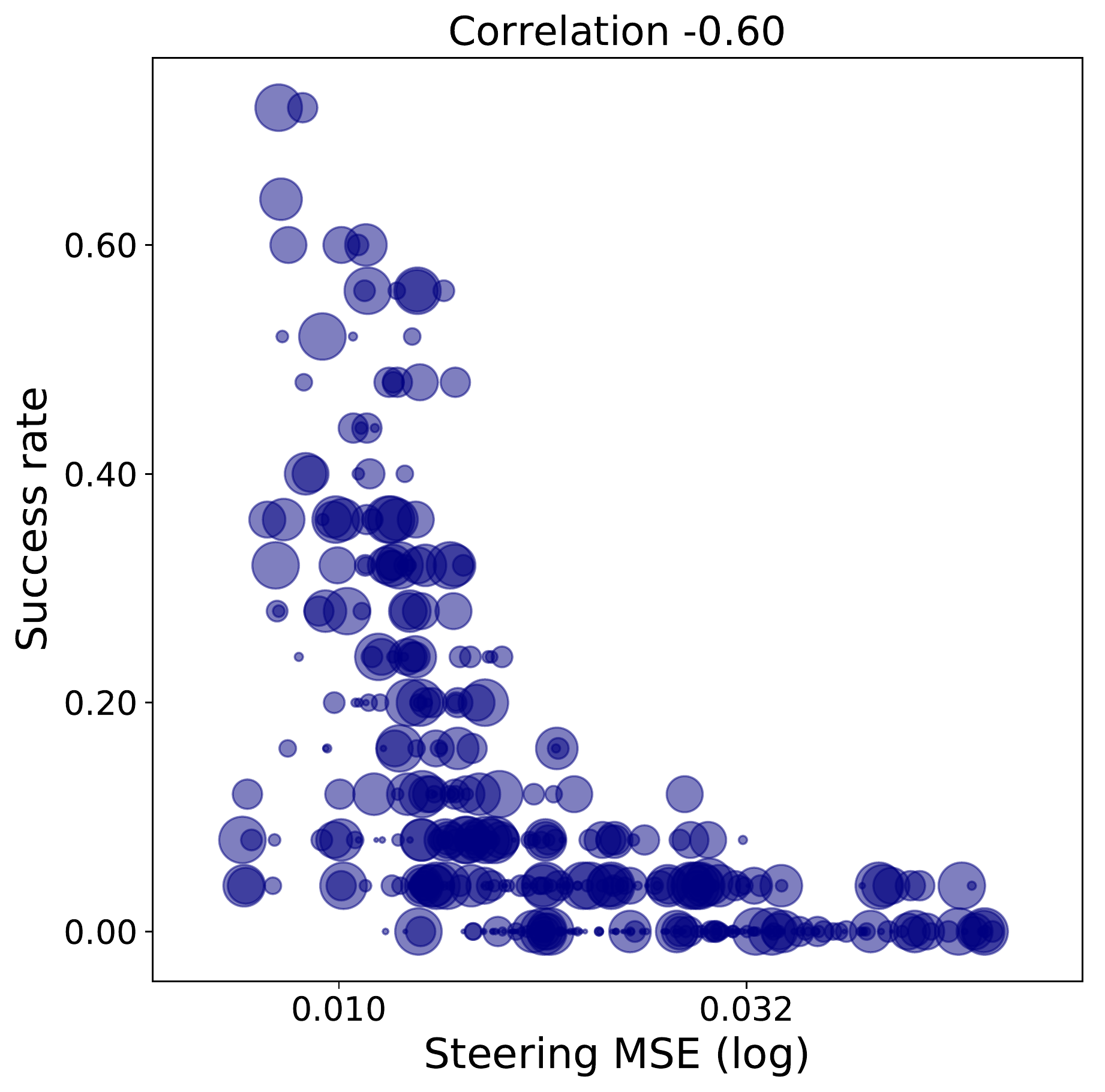} &
    \includegraphics[height=0.32\linewidth]{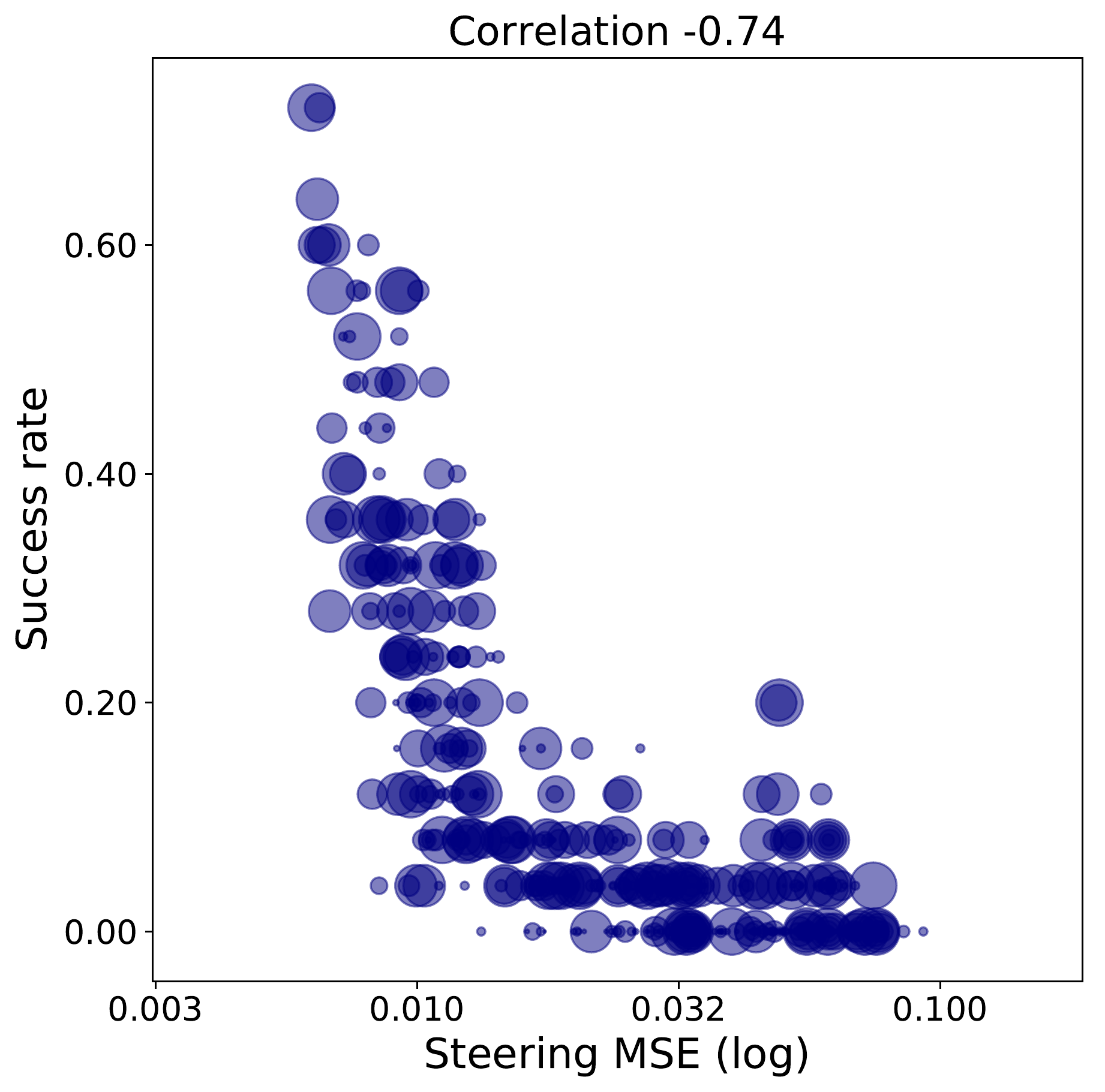}
  \end{tabular}
  }
  \caption{Scatter plots of goal-directed navigation success rate vs steering absolute error when evaluated on data from different distributions. Town 2 (generalization conditions), all models.}
  \label{fig:scatter_steer_vary_data_town2all}
\end{figure}

\begin{figure}
  \centering
  { \fontsize{7pt}{9pt}\selectfont
  \begin{tabular}{ccc}
		\multicolumn{3}{c}{\normalsize Town 2 (generalization conditions), all models.}\\
    \quad Steering MSE & \quad Steering absolute error & \quad Speed-weighted error \\
    \includegraphics[height=0.32\linewidth]{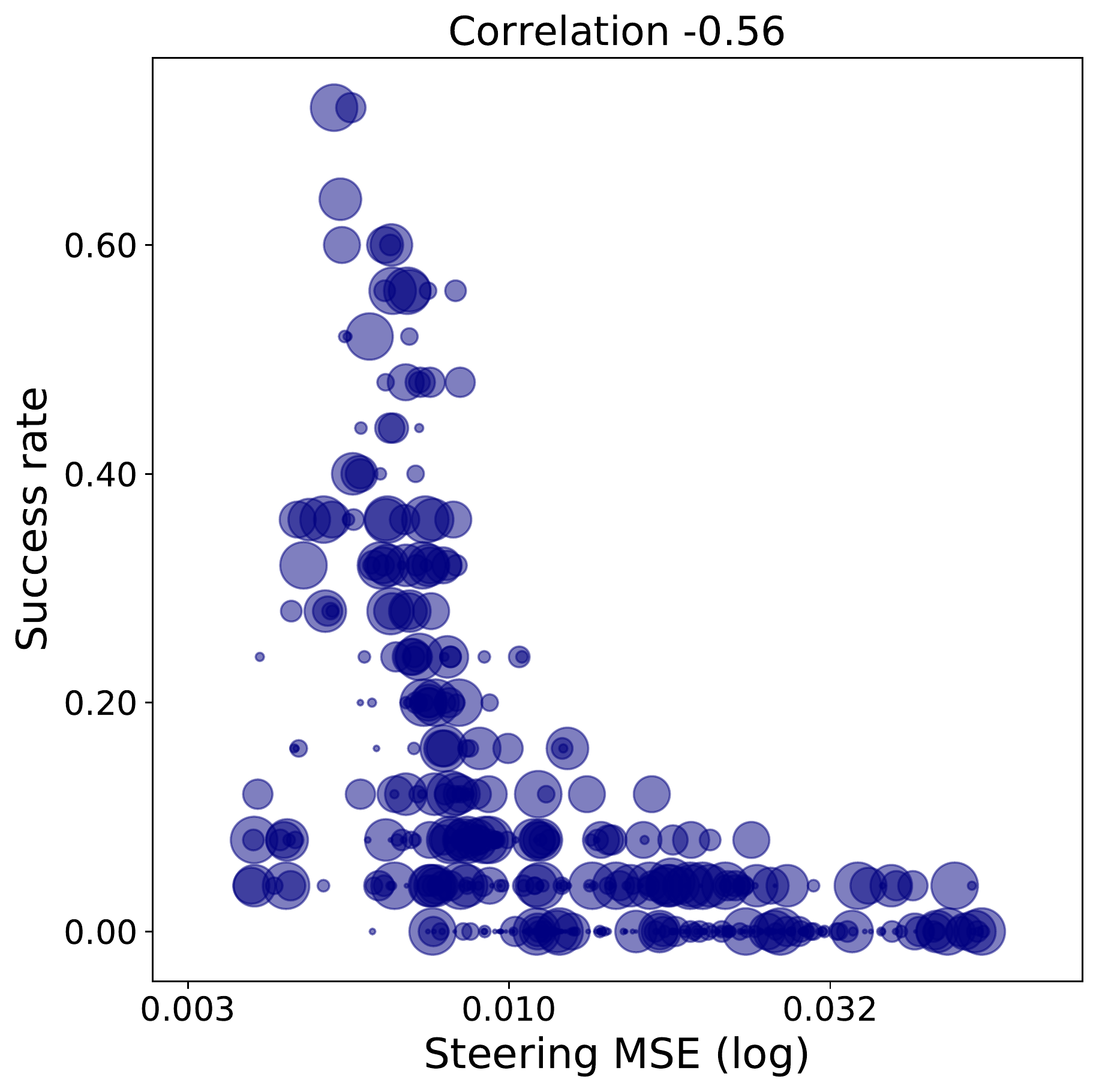} &
    \includegraphics[height=0.32\linewidth]{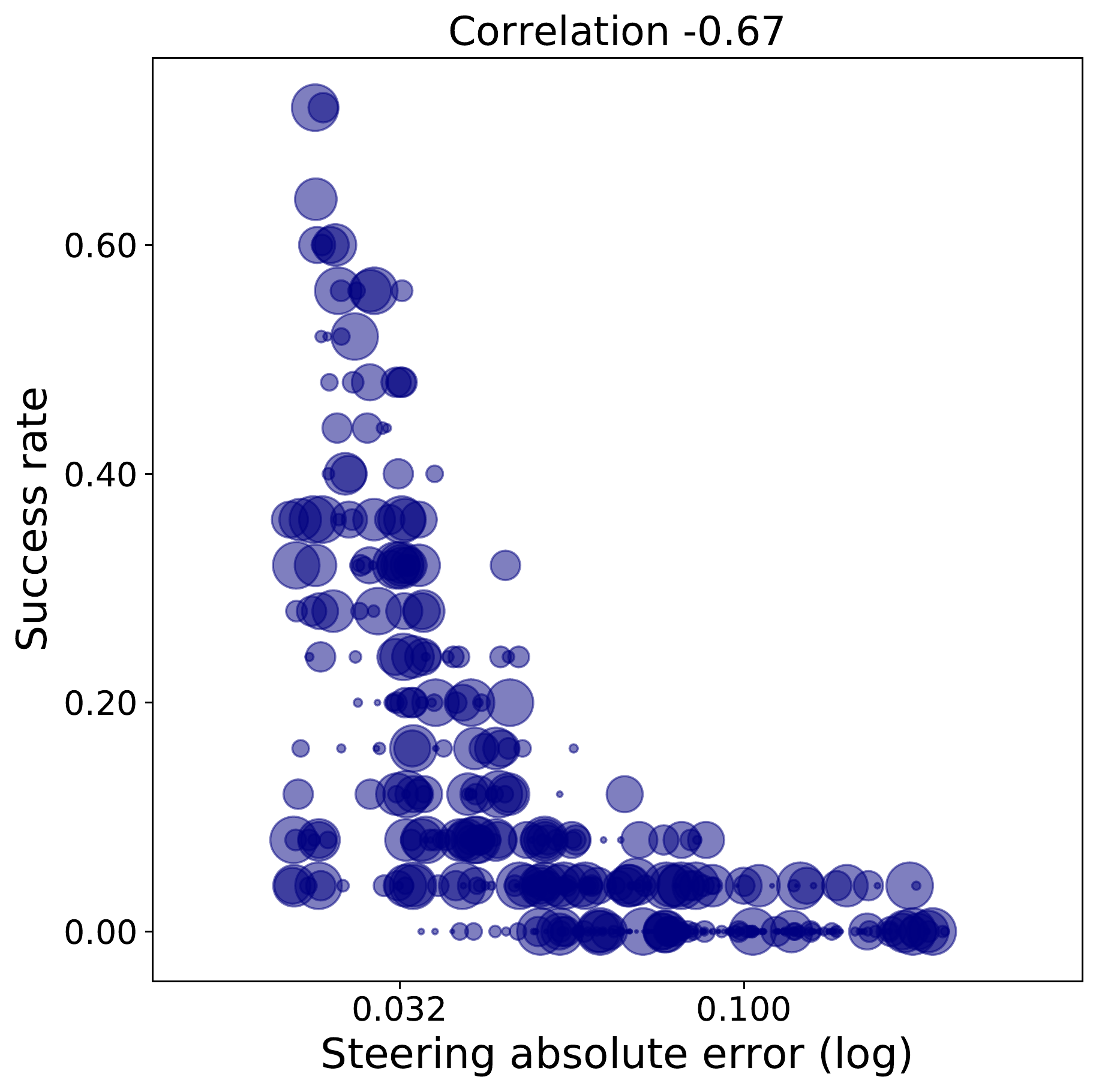} &
    \includegraphics[height=0.32\linewidth]{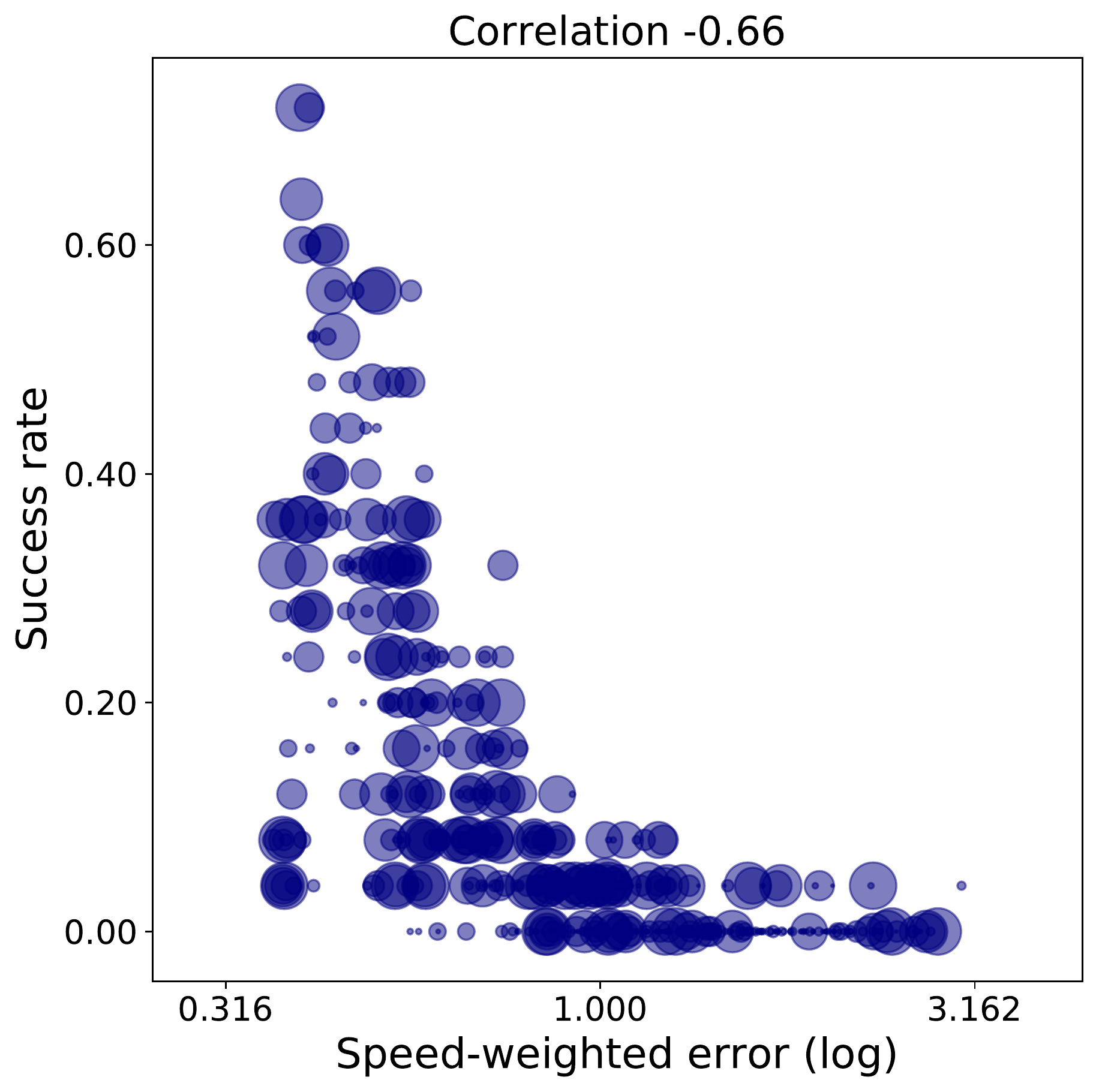} \\
    \quad Cumulative error & \quad Quantized classification & \quad Thresholded relative error \\
    \includegraphics[height=0.32\linewidth]{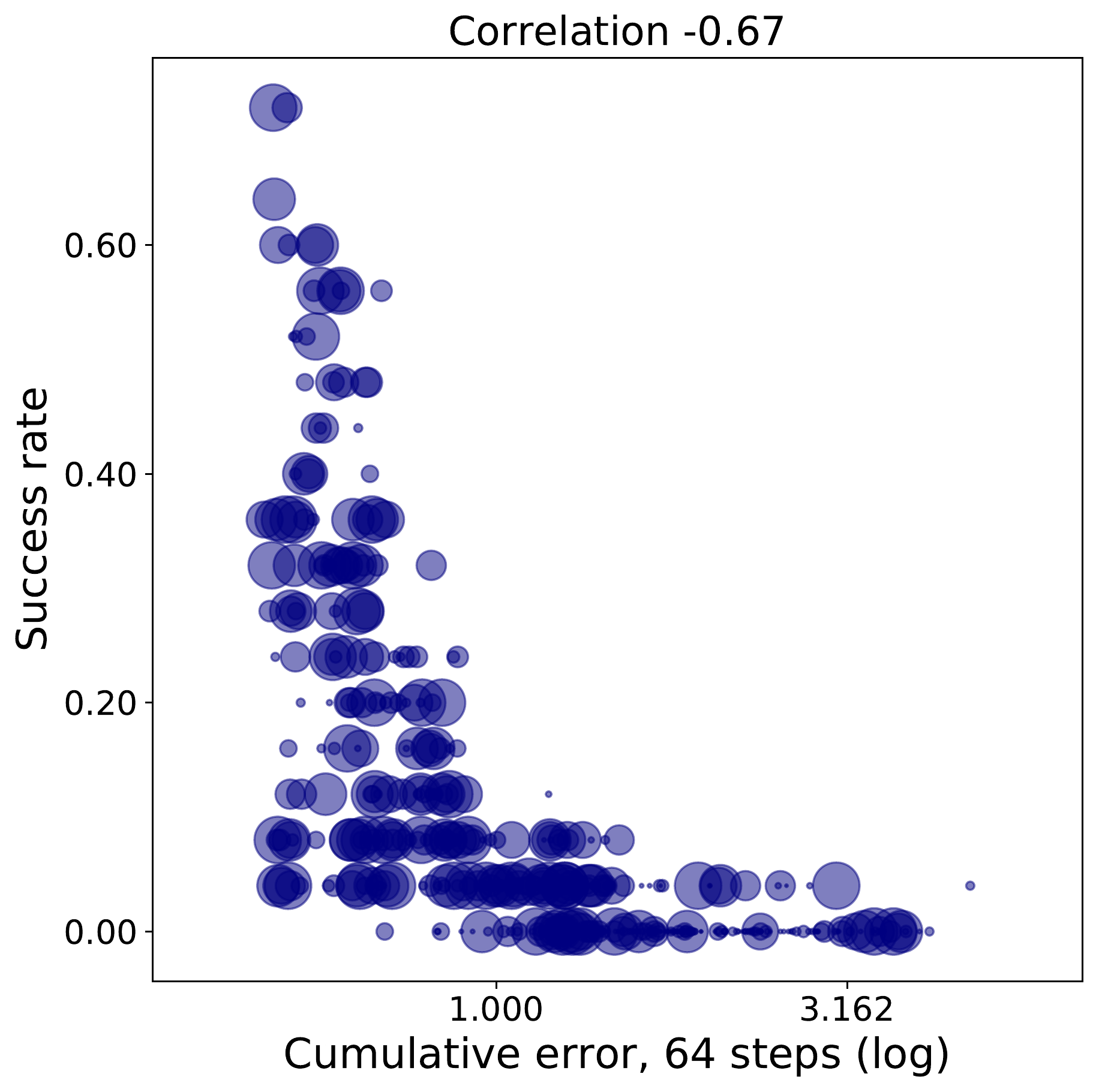} &
    \includegraphics[height=0.32\linewidth]{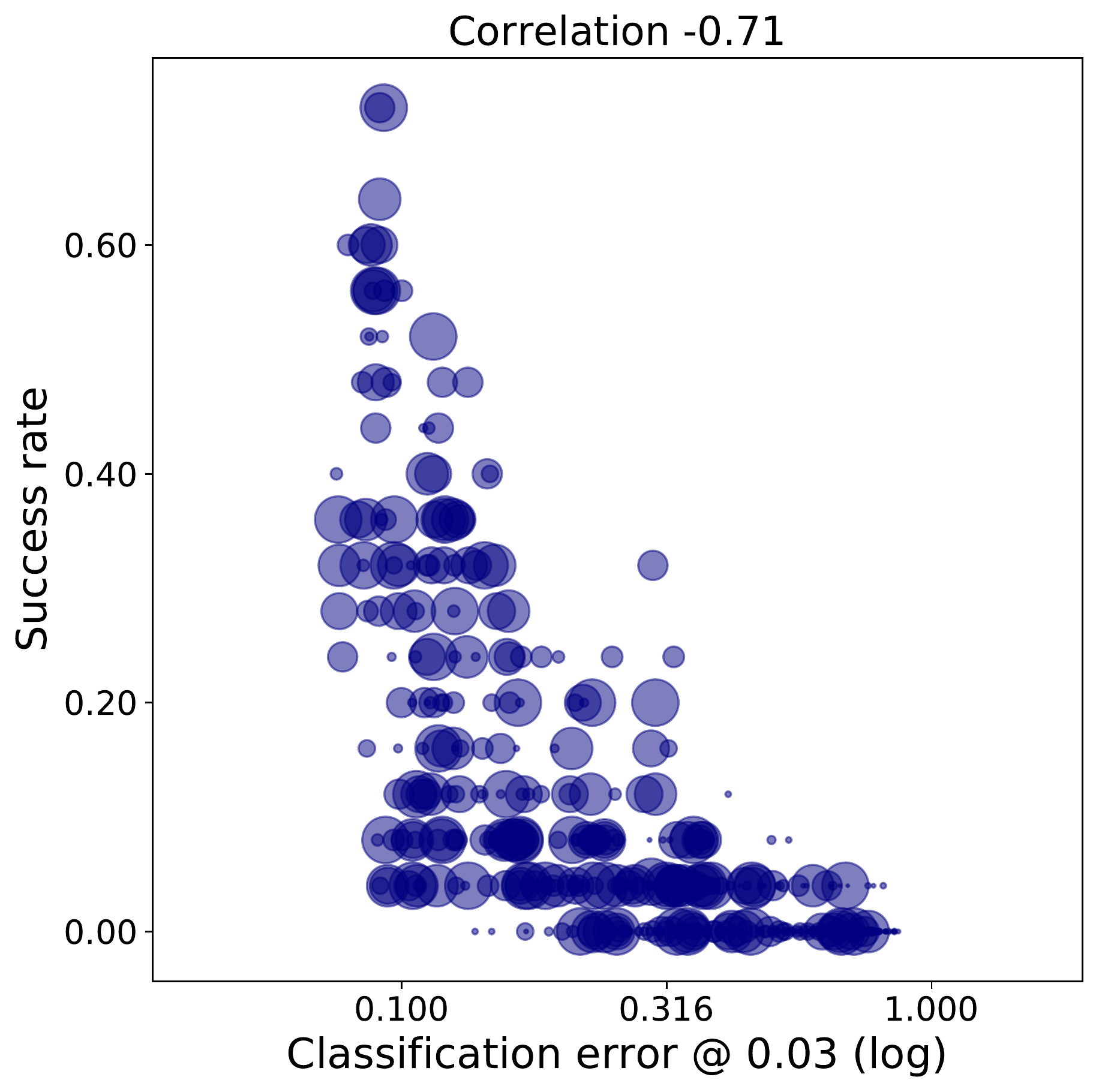} &
    \includegraphics[height=0.32\linewidth]{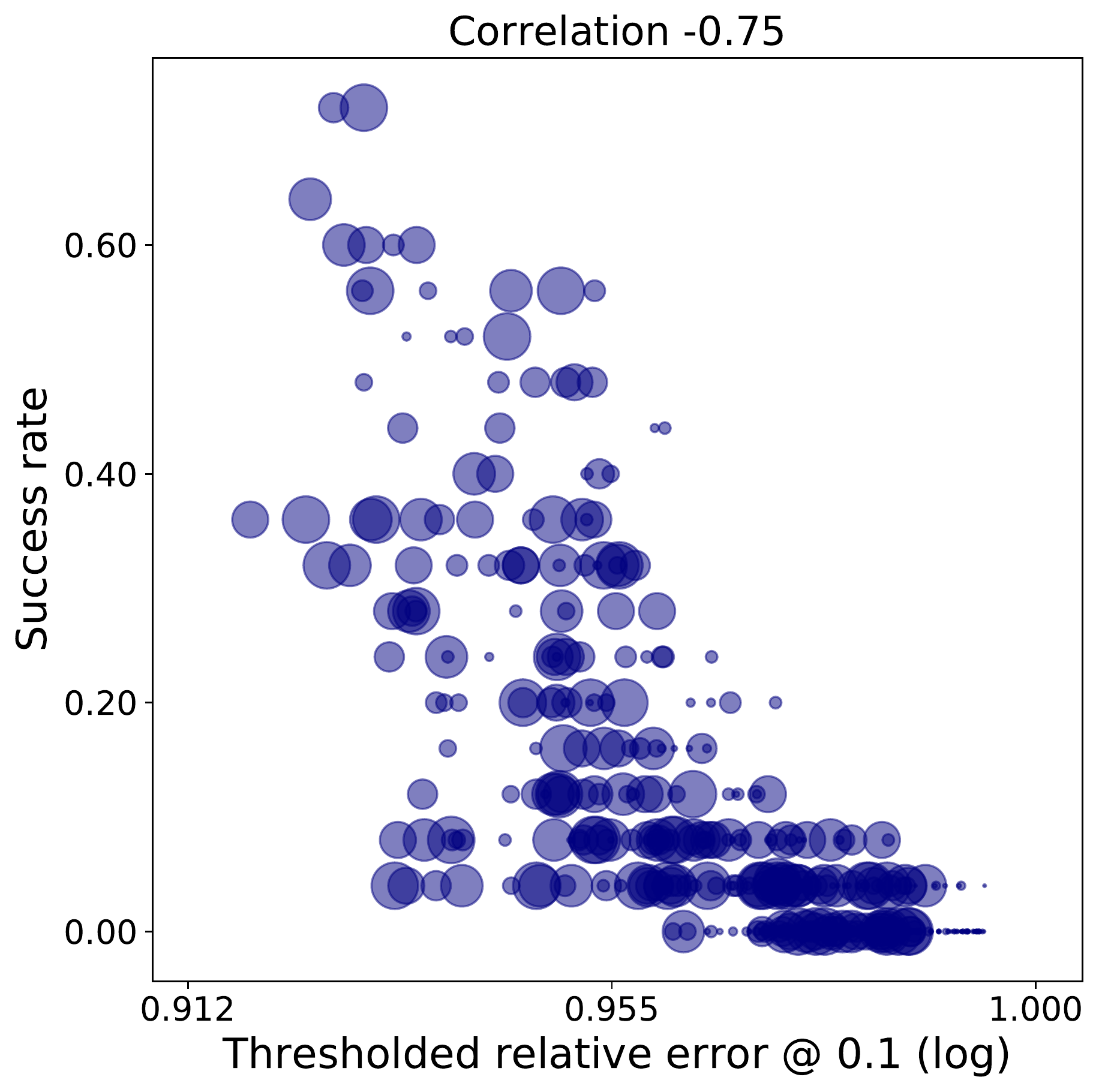}
  \end{tabular}
  }
  \caption{Scatter plots of goal-directed navigation success rate vs different offline metrics. Town 2 (generalization conditions), all models.}
  \label{fig:scatter_offline_metrics_town2all}
\end{figure}

Figures~\ref{fig:scatter_steer_vary_data_town1best} and \ref{fig:scatter_offline_metrics_town1best} show scatter plots of online vs offline metrics with 50\% best models, evaluated in Town 1.
Figure~\ref{fig:scatter_online_metrics_town1all} shows scatter plots of online driving quality metrics, evaluated in Town 1.
Figures~\ref{fig:scatter_steer_vary_data_town1all} and ~\ref{fig:scatter_offline_metrics_town1all} show scatter plots of online vs offline metrics with all models, evaluated in Town 1.
Figures~\ref{fig:scatter_steer_vary_data_town2all} and~\ref{fig:scatter_offline_metrics_town2all} show scatter plots of online vs offline metrics with all models, evaluated in Town 2.

\newpage

{\small
\bibliographystyle{splncs04}
\bibliography{paper}
}